\documentclass{article} 
\usepackage{iclr2021_conference,times}

\usepackage{amsmath,amsfonts,bm}

\def\eqref#1{equation~\ref{#1}}

\def\1{\bm{1}}

\def\vtheta{{\bm{\theta}}}
\def\va{{\bm{a}}}

\def\vx{{\bm{x}}}

\def\eva{{a}}

\def\mA{{\bm{A}}}
\def\mB{{\bm{B}}}

\def\mI{{\bm{I}}}
\def\mJ{{\bm{J}}}

\def\mX{{\bm{X}}}

\DeclareMathAlphabet{\mathsfit}{\encodingdefault}{\sfdefault}{m}{sl}
\SetMathAlphabet{\mathsfit}{bold}{\encodingdefault}{\sfdefault}{bx}{n}
\newcommand{\tens}[1]{\bm{\mathsfit{#1}}}

\def\tC{{\tens{C}}}

\def\tX{{\tens{X}}}

\def\sA{{\mathbb{A}}}
\def\sB{{\mathbb{B}}}

\def\sV{{\mathbb{V}}}
\def\sW{{\mathbb{W}}}

\def\emA{{A}}

\newcommand{\etens}[1]{\mathsfit{#1}}

\def\etC{{\etens{C}}}

\def\etX{{\etens{X}}}

\newcommand{\R}{\mathbb{R}}

\newcommand{\normltwo}{L^2}
\newcommand{\normlp}{L^p}

\usepackage{makecell}
\usepackage{amsmath}
\usepackage{bm}

 \usepackage{tabu}

\usepackage{hyperref}
\usepackage{url}
\usepackage{enumitem}
\usepackage{amsmath}
\usepackage[bb=dsserif]{mathalpha}
\usepackage{bm}

\usepackage{algorithm} 
\usepackage{algpseudocode}  
\usepackage{graphicx}
\usepackage{multirow}
\usepackage{booktabs}
\usepackage{tabularray}
\usepackage{nicematrix}

\graphicspath{ {./figures/} }

\usepackage{tikz}
\usepackage{graphics}
\usepackage{color}
\usepackage{fullpage}
\usepackage{float}
\usepackage{caption}
\usepackage{subcaption}
\usepackage{mathtools, cases}

\usepackage{annotate-equations}

\usepackage{ifthen}

\usepackage{xcolor}

\newcommand{\tbf}[1]{\textcolor{red}{#1}}

\newlength{\maxwidth}
\newcommand{\algalign}[2]
{\makebox[\maxwidth][r]{$#1{}$}${}#2$}

\DeclareDocumentCommand{\jingAlgo}{ m O{=\ } }{%
	{\rlap{$#1$} \hphantom{text}$#2$}%
}

\newcommand{\bW}{\boldsymbol{W}}
\newcommand{\bw}{\boldsymbol{w}}
\newcommand{\bX}{\boldsymbol{X}}
\newcommand{\bY}{\boldsymbol{Y}}
\newcommand{\bS}{\boldsymbol{S}}

\newcommand{\bU}{\boldsymbol{U}}

\newcommand{\bI}{\boldsymbol{I}}
\newcommand{\bx}{\boldsymbol{x}}
\newcommand{\by}{\boldsymbol{y}}
\newcommand{\bz}{\boldsymbol{z}}
\newcommand{\br}{\boldsymbol{r}}

\newcommand{\bh}{\boldsymbol{h}}
\newcommand{\bo}{\boldsymbol{o}}
\newcommand{\bt}{\boldsymbol{t}}
\newcommand{\bv}{\boldsymbol{v}}
\newcommand{\bq}{\boldsymbol{q}}
\newcommand{\bk}{\boldsymbol{k}}

\newcommand{\bO}{\boldsymbol{O}}

\newcommand{\bg}{\boldsymbol{g}}
\newcommand{\bmm}{\boldsymbol{m}}

\newcommand{\ba}{\boldsymbol{a}}

\newcommand{\bb}{\boldsymbol{b}}
\newcommand{\bp}{\boldsymbol{p}}

\newcommand{\bQ}{\boldsymbol{Q}}
\newcommand{\bK}{\boldsymbol{K}}
\newcommand{\bV}{\boldsymbol{V}}

\let\tbf\textbf

\definecolor{commentcolor}{RGB}{110,154,155}   

\usepackage{amsmath}

\usepackage{array, caption, tabularx, makecell, booktabs}%

\usepackage[framemethod=TikZ]{mdframed}
\usepackage{amsthm}

\usepackage{multirow}

\newcounter{theo}[section] 
\renewcommand{\thetheo}{\arabic{section}.\arabic{theo}}
\newenvironment{theo}[2][]{%
\refstepcounter{theo}%
\ifstrempty{#1}%
{\mdfsetup{%
frametitle={%
\tikz[baseline=(current bounding box.east),outer sep=0pt]
\node[anchor=east,rectangle,fill=blue!20]
{\strut Remark~\thetheo};}}
}%
{\mdfsetup{%
frametitle={%
\tikz[baseline=(current bounding box.east),outer sep=0pt]
\node[anchor=east,rectangle,fill=blue!20]
{\strut Remark~\thetheo:~#1};}}%
}%
\mdfsetup{innertopmargin=10pt,linecolor=blue!20,%
linewidth=2pt,topline=true,%
frametitleaboveskip=\dimexpr-\ht\strutbox\relax
}
\begin{mdframed}[]\relax%
\label{#2}}{\end{mdframed}}
\newcounter{lem}[section] 
\renewcommand{\thelem}{\arabic{section}.\arabic{lem}}

\newcounter{prf}[section]
\renewcommand{\theprf}{\arabic{section}.\arabic{prf}}

\graphicspath{ {./figures/} }

\DeclareDocumentCommand{\jingAlgo}{ m O{=\ } }{%
	{\rlap{$#1$} \hphantom{text}$#2$}%
}

\title{Understanding Optimization of Deep Learning via \\Jacobian Matrix and Lipschitz Constant}

\author{
Xianbiao Qi, Jianan Wang, and Lei Zhang \\
International Digital Economy Academy (IDEA), Shenzhen, Guangdong, China.\\
\texttt{\{qixianbiao,wangjianan,leizhang\}@idea.edu.cn} 
}

\newtheorem{principle}{Principle}
\newtheorem{definition}{Definition}

\newtheorem{lemma}{Lemma}

\newtheorem{example}{Example}

\iclrfinalcopy 

\begin{document}

\maketitle

\begin{abstract}

This article provides a comprehensive understanding of optimization in deep learning, with a primary focus on the challenges of gradient vanishing and gradient exploding, which normally lead to diminished model representational ability and training instability, respectively. We analyze these two challenges through several strategic measures, including the improvement of gradient flow and the imposition of constraints on a network's Lipschitz constant. To help understand the current optimization methodologies, we categorize them into two classes: explicit optimization and implicit optimization. Explicit optimization methods involve direct manipulation of optimizer parameters, including weight, gradient, learning rate, and weight decay. Implicit optimization methods, by contrast, focus on improving the overall landscape of a network by enhancing its modules, such as residual shortcuts, normalization methods, attention mechanisms, and activations. In this article, we provide an in-depth analysis of these two optimization classes and undertake a thorough examination of the Jacobian matrices and the Lipschitz constants of many widely used deep learning modules, highlighting existing issues as well as potential improvements. Moreover, we also conduct a series of analytical experiments to substantiate our theoretical discussions. This article does not aim to propose a new optimizer or network. Rather, our intention is to present a comprehensive understanding of optimization in deep learning. We hope that this article will assist readers in gaining a deeper insight in this field and encourages the development of more robust, efficient, and high-performing models.
\end{abstract}

\tableofcontents

\section{Introduction}
\label{sec:optimization_principles}

\begin{figure}[thb] \centering
    \includegraphics[width=0.5\textwidth]{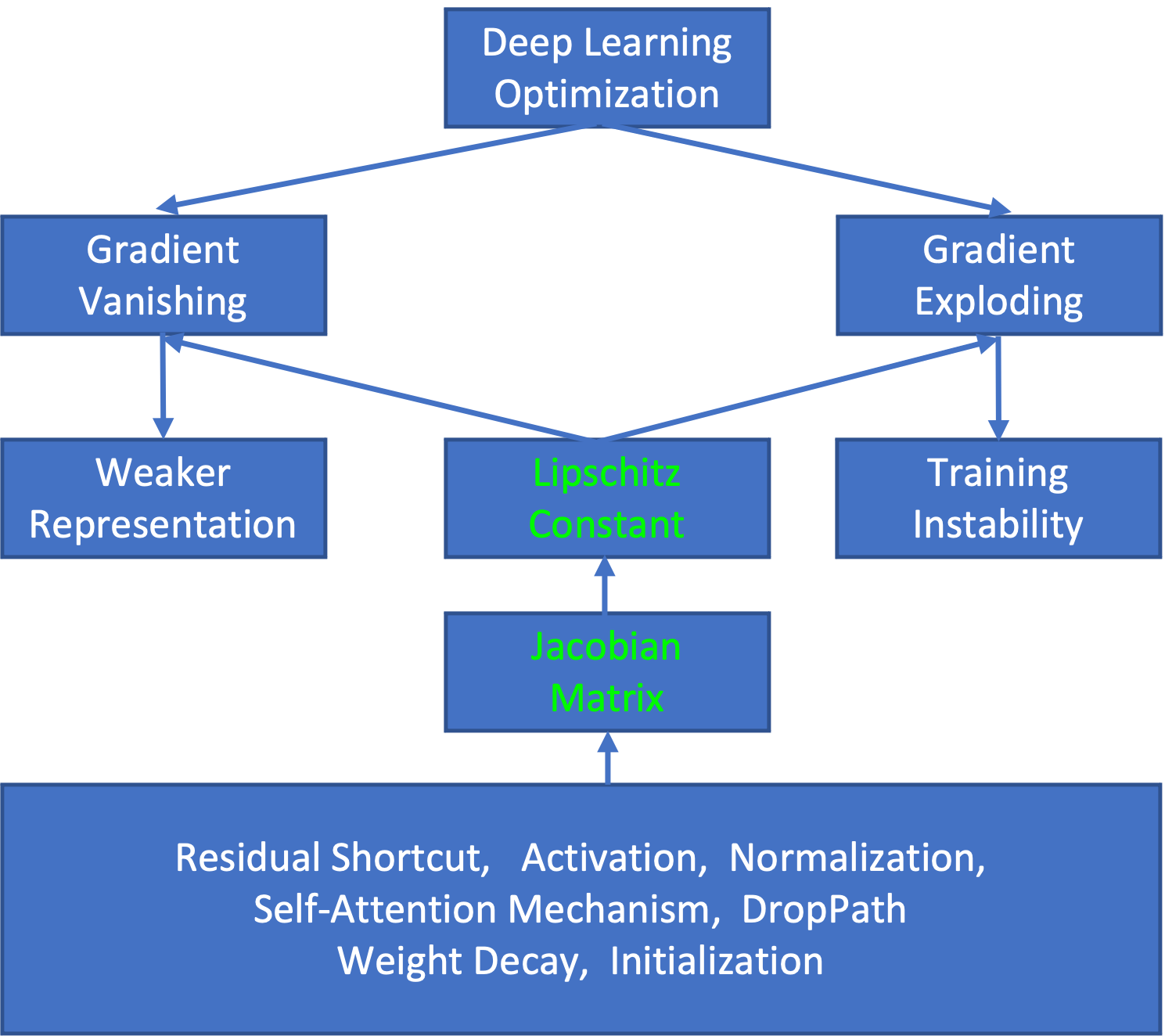}
    \caption{An optimization overview.} 
    \label{fig:optimization_overview}
\end{figure}

Deep learning has revolutionized a myriad of industries and disciplines, extending the boundaries of machine learning capabilities. The sectors transformed by this technology include computer vision (CV)~\citep{resnet_he2016deep, vit_dosovitskiy2020image, detr_carion2020end, swintransformer_liu2021swin, swin_v2_liu2021swin2, convnext_liu2022convnet}, natural language processing (NLP)~\citep{gpt_radford2018improving, gpt2_radford2019language, gpt3_brown2020language, palm_chowdhery2022palm, opt_zhang2022opt, chinchilla_hoffmann2022training, Emergent_ability_wei2022emergent, llama_touvron2023llama}, multi-modal understanding and generation~\citep{clip_radford2021learning, dalle_ramesh2021zero, dalle2_ramesh2022hierarchical, imagen_saharia2022photorealistic, stable_diffusion_rombach2022high}, and others~\citep{alphago_silver2016mastering, alphafold_jumper2020alphafold}.
Despite these remarkable achievements, the mastery of deep learning~\citep{lecture_about_Lipschitz_heinonen2005lectures, deeplearning_for_ai_bengio2021deep} still presents a series of unique challenges. This article aims to shed light on one such critical and intricate aspect: the optimization of deep learning models.

Optimization in deep learning~\citep{book_leon_bottou2018optimization, book_goodfellow2016deep, sun2019optimization} is a multifaceted endeavor. It involves tuning the parameters of a model through back-propagation~\citep{backpropagation_rumelhart1986learning, lecun1989backpropagation, lenet_lecun1998gradient} in an effort to minimize the discrepancy between the model's predictions and the actual data. However, the process of optimization is not straightforward. It constitutes a journey through a high-dimensional and often non-convex landscape, filled with numerous local minima and saddle points. Navigating this landscape introduces its own set of challenges. The two most notable challenges are gradient vanishing and gradient exploding.

The gradient vanishing problem~\citep{xavier_init_glorot2010understanding, resnet_he2016deep, identity_mapping_he2016identity} refers to a phenomenon where gradients shrink exponentially as they are propagated backwards through the layers of the network during training. This issue leads to the early layers of the network being updated slowly, resulting in a network with diminished representational ability as the early layers are unable to learn complex, meaningful representations of the input data.

On the other hand, the gradient exploding problem~\citep{difficulty_pascanu2013difficulty, admin_liu2020understanding, deepnet_wang2022deepnet} is characterized by the exponential growth of gradients during back-propagation. This issue often leads to unstable training as the model's parameters undergo large, volatile updates. Such instability can prompt a variety of issues, ranging from wildly oscillating loss values to, in extreme cases, the model's failure to converge.

Despite these challenges, numerous strategies and techniques exist to tackle both issues. To alleviate the gradient vanishing problem, the emphasis is typically on promoting improved gradient flow through the network. This can be achieved through various means, such as an implementation of skip or residual connections, a careful initialization of weights, or an use of non-saturating activation functions.
On the other hand, to counteract the gradient exploding problem, a common approach is to constrain the Lipschitz constant of the network. The Lipschitz constant serves as a measure of the network's sensitivity to changes in its inputs. By controlling this constant, we can constrain the growth of the gradients, thereby stabilizing the training process.

However, there remains a significant gap in the theoretical understanding of these methods. This article aims to bridge this gap. We categorize existing optimization methods into two primary facets: explicit and implicit optimization. Explicit optimization methods directly act upon optimizer parameters, which include weight, gradient, learning rate, and weight decay. Implicit optimization methods, on the other hand, focus on refining network modules to enhance the network's optimization landscape. These methods encompass techniques such as residual shortcuts, normalization methods, activations and attention mechanisms. In this article, we provide an in-depth analysis of these two classes of optimization. Specifically, we conduct a detailed examination of the gradient or Jacobian and the Lipschitz constant of the widely-used deep learning modules, pinpoint potential issues, and identify existing and prospective improvements. Figure~\ref{fig:optimization_overview} illustrates a general overview of our understanding of optimization in deep learning. We would like to highlight that the problems of gradient vanishing and gradient exploding both can be attributed to the Jacobian matrix of each module. One conclusion from Figure~\ref{fig:optimization_overview} is that \textit{Jacobian matrices determine the back-propagation process, and Lipschitz constant, that can be calculated according to the Jacobian matrices,
 affects representation ability and training stability of a network. Therefore, to understand the optimization of deep learning in depth, we need to analyze Jacobian matrix and Lipschitz constant of each module in detail.} 
In this paper, we will also provide theoretical analysis of some existing skills.
In addition to the theoretical analysis, we perform analytical experiments to verify our theoretical assertions.

Below, we briefly summarize some of our analyses and observations:

\begin{itemize}[leftmargin=*]
\item A convolutional network comprises homogeneous blocks~\footnote{Homogeneous operator denotes these modules have similar form of Jacobian matrices, such as linear layer and convolution, both are first-order linear operators. Heterogeneous operators mean these modules have very different Jacobian properties, such as linear layer and self-attention, the former is a linear operator but the latter is high-order nonlinear operator.}, such as Convolutions and Linear Layers. In contrast, the Transformer network includes heterogeneous blocks like Multi-head self-attention and Feed-forward Networks (FFN). These heterogeneous blocks have distinct Jacobian matrices and differing Lipschitz constants, adding complexity to the optimization of Transformer models.
\item The Adam optimizer demonstrates robustness to variations of Lipschitz constants during the training process, as it employs a normalized update value (i.e., the element-wise division between the first-order momentum and the square root of the second-order momentum). Conversely, the Stochastic Gradient Descent (SGD) optimizer is highly sensitive to changes in the Lipschitz constant of the network. The AdamW optimizer rectifies incorrect weight decay, thereby improving its performance.
\item The initialization of a network should be mindful of Lipschitz constant, particularly for larger models. To achieve this, we recommend the use of Lipschitz-aware initialization.
\item Residual shortcut, despite its advantage in mitigating the gradient vanishing problem in the backward process,  smooths the landscape of the network.
\item Normalization is a useful method to ensure that a network adheres to the forward optimization principle, also contributing to a smoother network landscape.
\item Weight decay and DropPath can reduce the Lipschitz constant of the network, thereby decreasing the likelihood of unstable training. Essentially, they function as contraction mappings.
\item Dot-Product attention and normalization techniques, despite their strong representation capabilities, exhibit large Lipschitz constants. Consequently, these methods are more likely to trigger unstable training during the backward process compared to convolution, fully connected (FC) layers, and activation functions.
\item Instances of unstable training often coincide with rapid increases in  Lipschitz constant of a network. This phenomenon is typically indicated by a swift increase in the top eigenvalues of the weight matrices.
\end{itemize}
Even though the research community has gained a deeper understanding of optimization in deep learning, numerous open questions still remain, such as: 
\begin{itemize}[leftmargin=*]
    \item What are the properties of weight updates in optimizers? Is the function for updating contraction mapping or expansion mapping? If it is expansion mapping, what is the expansion factor? 
    \item Is it possible to discover an automatic setup and adjustment strategy for the learning rate and weight decay according to a simulated Lipschitz constant of the network?
    \item What is the value and necessity of warmup? Why is  it so important especially in large model?
    \item What are the implications of constrained optimization methods in deep learning? 
    \item What is the relationship between representation ability and training stability?
\end{itemize}

There are many additional open problems that warrant in-depth exploration, including:
\begin{itemize}[leftmargin=*]
    \item  Is a second-order optimization method necessary and more powerful?
    \item How is Lipschitz smoothness considered in deep learning? Smoothness is usually the fundamental assumption when in numerical optimization.
    \item What is the comparison of generalization ability between non-smooth and smooth functions?
\end{itemize}
We will not delve into these open questions in this article due to limit space and our unclear understanding of these questions, but they certainly deserve serious consideration in future studies.
This article does not aim to provide a survey of optimization methods. Instead, our objective is to develop a simple, thorough, and comprehensive understanding of optimization in deep learning. For a survey of optimization methods, readers are referred to~\cite{sun2019optimization, sun2019survey, survey_fang_li2020accelerated}.

\subsection{Outline}
The structure of this article is as follows: In Section~\ref{sec:lipschitz}, we introduce fundamental optimization concepts, including Lipschitz continuity, contraction mapping, Lipschitz gradient and 
 Hessian continuity. In Section~\ref{sec:deeplearning}, we review the essential modules in deep learning, which include linear layer, convolutions, normalization, residual shortcut, self-attention, activation, and feed-forward network. Section~\ref{sec:optimization_principles_deeplearning} provides an overview of deep learning optimization, covering both forward and backward optimization perspectives and introducing our general optimization principles for deep learning. Section~\ref{sec:implicit_optimization} discusses methods for implicitly optimizing the network, while Section~\ref{sec:explicit_optimization} addresses practical considerations from an optimizer's perspective. Here, we also provide both theoretical and practical remarks about each factor. In Section~\ref{sec:guideline}, based on our analysis and discussions from previous sections, we compile guidelines for deep learning optimization. We then conduct experiments in Section~\ref{sec:experiments} to validate our theoretical analysis. In Section~\ref{sec:discussion}, we discuss the difficulties of optimizing large models and other existing problems in deep learning optimization. Finally, in Section~\ref{sec:Conclusion}, we draw a conclusion for this article.

\subsection{Notation}
Before we delve into specific algorithms, let us provide a brief introduction to the notation system utilized throughout this article. We primarily follow the notation system of the renowned deep learning book~\footnote{\url{https://github.com/goodfeli/dlbook_notation}}~\citep{book_goodfellow2016deep}. We use $\bx$ to denote a column vector with $\bx \in \mathbb{R}^{D}$, and $\bX$ to represent a set with $N$ points with $\bX \in \mathbb{R}^{D\times N}$. $\bW$ is a weight matrix with $\bW \in \mathbb{R}^{D_1 \times D}$, and $\by = \bW \bx$ with $\by$ being a  column vector in $\mathbb{R}^{D_1}$. It should be noted that our notation of self-attention in this article differs from that in~\citep{lipsformer_qi2023lipsformer}, for which we apologize for any inconsistency. As an example, here, $\bY = {(\bW_1 \bX)}^\top (\bW_2 \bX)$, where both $\bW_1$ and $\bW_2 \in \mathbb{R}^{D_1 \times D}$, $\bX \in \mathbb{R}^{D \times N}$, and $\bY$ is a tensor in $\mathbb{R}^{N \times N}$.

In terms of matrix calculus, we use the denominator layout~\footnote{\url{https://en.wikipedia.org/wiki/Matrix_calculus}}. Therefore, the Jacobian matrix of $\by$ with respect to $\bx$ is represented as $\boldsymbol{J}_{\by}(\bx) = \bW^{\top}$. Consequently, we have the following equations:
\begin{equation}
\begin{array}{cc}
\frac{\partial \bw^{\top} \bx}{\partial \bx}=\bw, \ \ 
\frac{\partial \bW \bx}{\partial \bx}=\bW^{\top},
\frac{\partial \bx^{\top} \bW \bx}{\partial \bx}=\left(\bW+\bW^{\top}\right) \bx.
\end{array}
\end{equation}

Suppose we have a chain function $\bo = f(g(h(\bx)))$, where $\by = h(\bx)$, $\bz = g(\by)$, and $\bo = f(\bz)$. Then, utilizing the denominator layout, the Jacobian matrix of $\bo$ with respect to $\bx$ according to the chain rule is: 
\begin{equation}
    \boldsymbol{J}_{\bo}(\bx) = \frac{\partial \by}{\partial \bx} \frac{\partial \bz}{\partial \by} \frac{\partial \bo}{\partial \bz}.
\end{equation}

More knowledge about calculus can be found in~\cite{matrix_cookbook_petersen2008matrix}. A checklist of notations can be found in the Appendix~\ref{sec:appendix_notation}.

\section{Foundations on Optimization}
\label{sec:lipschitz}
Lipschitz continuity\footnote{\url{https://en.wikipedia.org/wiki/Lipschitz_continuity}}, in mathematical analysis, represents a strong form of uniform continuity for functions. To probe the characteristics of functions, it is beneficial to understand their Lipschitz properties, along with those of their derivatives. Considering that a neural network is a specific function composed of multiple layers of simple functions, it's critical to comprehend the basic concepts of Lipschitz continuity to grasp deep neural networks (also referred as deep learning).

\subsection{Lipschitz Continuity}
\begin{definition}[Lipschitz Continuity]
\label{def-1}
A function \(f(\bx)\) : $\mathbb{R}^{D} \rightarrow \mathbb{R}^{D_1}$ is said to be Lipschitz continuous (or $K_0$-Lipschitz) under a chosen p-norm $\| \cdot \|_p$ in the variable \(\bx\) if there exists a constant $K_0$ such that for all \(\bx_1\) and \(\bx_2\) in the domain of \(f\), the following inequality is always satisfied,
\begin{equation*}
  \|f(\bx_1)-f(\bx_2)\| \leq K_0 \|\bx_1-\bx_2\|.
\end{equation*}
\end{definition}

For ease of understanding, we can default to considering the norm $\| \cdot \|$ as the Euclidean norm. We will specify if we use different norms.

Lipschitz continuity provides a bound on the rate at which a function can change and ensures that the function does not exhibit any extreme variations in value.

\begin{definition}[Local Lipschitz Continuity]
\label{def-1}
Given a point $\bx$,  and a function \(f(\bx)\) : $\mathbb{R}^{D} \rightarrow \mathbb{R}^{D_1}$. \(f(\bx)\) is said to be local Lipschitz continuous at point $\bx$ if there exists a constant $K_0$ such that for all points \(\bx + \bm{\epsilon}\), the following inequality is always satisfied,
\begin{equation*}
  \|f(\bx + \bm{\epsilon})-f(\bx)\| \leq K_0 \|\bm{\epsilon}\|,
\end{equation*}
\end{definition}
where $\|\bm{\epsilon}\| \leq \tau$.

Lipschitz constant at a point $\bx$  characterizes the curvature of the network at current point $\bx$.  Lipschitz constant of the whole network depicts the optimization landscape of the network.

\begin{lemma}[First-order Condition for Lipschitz Continuity]
\label{lemma-1}
A continuous and differentiable function \(f\) is $K_0$-Lipschitz continuous if and only if the norm of its gradient is bounded by \(K_0\),
\begin{equation}
  \|\nabla f(\bx)\| \leq K_0.
\end{equation}
\end{lemma}

It should be noted that the categories of ``continuously differentiable'' and ``Lipschitz continuous'' have the following relationship:
\begin{equation}
\text{Continuously differentiable} \subset \text{Lipschitz continuous}.
\end{equation}

This relationship indicates that every continuously differentiable function is also Lipschitz continuous, but the reverse is not necessarily true. In other words, the set of continuously differentiable functions is a subset of Lipschitz continuous functions. For example, the ReLU function is not continuously differentiable but is Lipschitz continuous. The condition of being continuously differentiable is stricter than being Lipschitz continuous, as it requires the function to have a limited gradient or Jacobian.

\begin{example}
Consider that $f({\bx}) = c$, where $c$ is a constant, the Lipschitz constant of $f(\bx)$ is 0. If $f({\bx}) = \ba^{\top} \bx + 1$, its Lipschitz constant can be computed as $\|\ba\|$. Now, if $f(\bx) = \bx^{\top} \bx + 2$, where $\bx$ is a column vector and $\bx \in \mathbb{R}^D$, the Lipschitz constant of the function $f(\bx)$ becomes $\infty$ and thus, $f(\bx)$ is not Lipschitz continuous.
\end{example}

\begin{definition}[Contraction Mapping]
\label{def-contractionmapping}
Let \((\bX, {f})\) be a metric space. A mapping \(\mathcal{M}: \bX \rightarrow \bX\) is called a contraction mapping if there exists a constant $K_0$, with $0 \leq K_0 <1$, such
that
\begin{equation}
 {f}(\mathcal{M}(\bx), \mathcal{M}(\by)) \leq K_0 {f}(\bx, \by),
\end{equation}
for all \(\bx, \by \in \bX\).
\end{definition}

\subsection{Lipschitz Gradient Continuity}

\begin{definition}[Lipschitz Gradient Continuity]
\label{def-2}
A function \(f(\bx)\) : $\mathbb{R}^{D} \rightarrow \mathbb{R}^{D_1}$ is said to have a Lipschitz continuous gradient (or $K_1$-Lipschitz) under a choice of p-norm $\| \cdot \|_p$ in the variable \(\bx\) if there exists a constant $K_1$ such that for all \(\bx_1\) and \(\bx_2\) in the domain of \(f\), the following inequality is always satisfied,
\begin{equation}
  \|\nabla f(\bx_1)-\nabla f(\bx_2)\| \leq K_1 \|\bx_1-\bx_2\|.
\end{equation}
\end{definition}

Lipschitz gradient continuity provides a bound on the rate at which the gradient of the function can change, ensuring that the function's slope does not change too abruptly.

For Lipschitz gradient continuity, we have the following lemma,
\begin{lemma}[(Smoothness Lemma]
\label{lemma-2}
A continuous and twice differentiable function \(f\) is $K_1$-smoothness if and only if
\begin{equation}
  \|\nabla^2 f(\bx)\|  < K_1.
\end{equation}
\end{lemma}

\begin{theo}[Lipschitz Continuity and Lipschitz Constant]{thm:lips_continuity}
\label{remark:lips_continuity}
\begin{enumerate}[leftmargin=*]
\item Lipschitz continuity is more general than continuously differentiable. \\
\item Local Lipschitz continuity and its Lipchitz constant at a point $\bx$ characterize the curvature of the network at the current point.  Lipschitz constant of the whole network depicts the optimization landscape of the network. \\
\item Analyzing Lipschitz constant of each module and even the whole network is an important and effective way to understand the properties of the network. \\
\end{enumerate}
\end{theo}

\subsection{Lipschitz Hessian Continuity}
Furthermore, we can define Lipschitz Hessian continuity as:
\begin{definition}[Lipschitz Hessian Continuity]
\label{def-lipshessian}
A function \(f(\bx)\) : $\mathbb{R}^{D} \rightarrow \mathbb{R}^{D_1}$ is said to have Lipschitz Hessian continuity (or $K_2$-Lipschitz) under a chosen p-norm $\| \cdot \|_p$ in the variable \(\bx\) if there exists a constant $K_2$ such that for all \(\bx_1\) and \(\bx_2\) in the domain of \(f\), the following inequality is always satisfied:
\begin{equation}
  \|\nabla^2 f(\bx_1)-\nabla^2 f(\bx_2)\| \leq K_2 \|\bx_1-\bx_2\|.
\end{equation}
\end{definition}

Lipschitz Hessian continuity provides a bound on the rate at which the curvature of the function can change, ensuring that a function's second-order derivatives do not change too abruptly.

In summary, Lipschitz continuity, Lipschitz gradient continuity, and Lipschitz Hessian continuity provide bounds on the rates of change of a function, its first-order derivatives, and its second-order derivatives, respectively. These properties help  understand the behavior of a function and are useful in optimization and numerical analysis problems. In Remark~\ref{remark:lips_continuity}, we have built several remarks about Lipschitz continuity and Lipschitz constant.

Interested audiences can refer to~\citep{book_nesterov_nesterov2003introductory, book_leon_bottou2018optimization, bubeck_bubeck2015convex, book_high_dimensional_data_wright2022high} for a more detailed introduction. For a deeper understanding, a Lipschitz monograph~~\citep{lecture_about_Lipschitz_heinonen2005lectures} is recommended.

\section{Foundations on Deep Learning}
\label{sec:deeplearning}
In this section, we will briefly introduce the mathematical definitions of some popular modules (also called layers) in deep learning. We will cover more discussions about certain improvements and their underlying mathematical principles in Section~\ref{sec:implicit_optimization}.

\subsection{Basic Modules in Deep Learning}

\subsubsection{Linear Layer}
\label{sec:intro_linear}
Linear projection (also called a linear layer in deep learning) is the most fundamental module in deep learning. Its definition is as follows:
\begin{equation}
    \by = f(\bx; \bW, \bb) = \bW \bx + \bb.
\end{equation}

The nature of linear projection is a linear feature transformation, which mathematically corresponds to a coordinate system transformation.

The Jacobian matrix~\footnote{\url{https://en.wikipedia.org/wiki/Jacobian_matrix_and_determinant}} of $\by$ with respect to $\bx$ can be calculated as:
\begin{equation}
    \frac{\partial \by}{\partial \bx} = {\bW}^{\top}.
\end{equation}

For an affine transformation $f\left(\bx; \bW, \bb \right) = \bW \bx + \bb$, its Lipschitz constant is,
\begin{equation} 
\label{eq:affine}
\operatorname{Lip}_p({f({\bx}}; \bW, \bb)) =\sup _{\|\boldsymbol{x}\|_{p}=1}\|\bW \boldsymbol{x} + \bb \|_{p} \\ =\left\{\begin{array}{ll}\sigma_{\max }(\bW), & \text { if } p=2 \\ \max _{i} \sum_{j}\left|W_{i j}\right| & \text { if } p=\infty\end{array}\right.
\end{equation}
where $\sigma_{\max}(\bW)$ is the largest absolute eigenvalue of $\bW$. 

Let $\| \bx \|_2 = 1$, if $\bx$ lies in the same direction as the maximum eigenvector of the matrix $\bW$, then $\| \bW \bx \| = \sigma_{\max }(\bW)$. On the other hand, if $\bx$ lies in the same direction as the minimum eigenvector of the matrix $\bW$, then $\| \bW \bx \| = \sigma_{\min }(\bW)$.

The forward process of a typical neural network propagates computation as $\by^{l+1} = \bW^{l+1} \bx^l + \bb^{l+1}$, where $\bx^l$ and $\bW^{l+1}$ are the input and the weight matrix of layer $l+1$. To back-propagate the network loss $\mathcal{L}$, we have
$$
\frac{\partial \mathcal{L}}{\partial \bx^{l}} = {(\bW^{l+1})}^{\top}
\frac{\partial \mathcal{L}}{\partial \by^{l+1}}, \ \ \ \ 
\frac{\partial \mathcal{L}}{\partial \bW^{l+1}} = 
{\frac{\partial \mathcal{L}}{\partial \by^{l+1}}} {(\bx^{l})}^{\top}, \ \ \ \ \frac{\partial \mathcal{L}}{\partial \bb^{l+1}} = 
{\frac{\partial \mathcal{L}}{\partial \by^{l+1}}}.
$$


Since deep learning is optimized using a stochastic optimization mechanism, the updated value of $\bW$ will affect the back-propagation process of $\bx$ in the next training step. Similarly, the value of $\bx$ will influence the update of $\bW$.

\subsubsection{Convolution}
\label{sec:intro_conv}
Convolution~\citep{lenet_lecun1998gradient, alexnet_krizhevsky2012imagenet, vgg_simonyan2014very, resnet_he2016deep} is a widely used and effective method in computer vision, with the concept of local receptive fields (LRF) being central to its effectiveness.

In convolutional neural networks (CNN), a LRF refers to a region in the input data (such as a small region in an image) that is connected to a neuron in a convolutional layer. This approach allows the network to focus on local features of the input data, reducing computational complexity and making the network more robust to variations in the input.

One advantage of using LRF is that it significantly reduces the number of parameters in the model. Instead of connecting each neuron to every pixel in the input image, each neuron is only connected to a small region of an image, resulting in a more manageable number of weights to learn. Another advantage of LRF is its ability to learn features in a hierarchical manner. When applied to image data, convolutional layers with local receptive fields can learn to recognize local features like edges and corners in early layers, which can then be combined in later layers to recognize higher-level features such as shapes and objects.

Suppose we have an input tensor $\bX \in \mathbb{R}^{\bar{W}\times \bar{H} \times C}$, with width $\bar{W}$, height $\bar{H}$, and channel $C$, and a kernel size of $K \times K$. The 2D convolution operation with stride 1 is defined as:
\begin{equation}
Y_{i,j,o} = \sum_{m=-\frac{K+1}{2}}^{\frac{K+1}{2}} \sum_{n=-\frac{K+1}{2}}^{\frac{K+1}{2}} \sum_{c=0}^{C-1} X_{i+m,j+n, c} \cdot W_{m,n,c,o}, 
\end{equation}

where $Y$ is the output tensor, and $i$, $j$, and $o$ are the row, column, and output channel indices of the output tensor $Y$. $\bW \in \mathbb{R}^{K\times K \times C \times O}$ is a 4d tensor, where $O$ is the number of output channels. This operation is carried out for $i = 0, 1, 2, ..., \bar{W}-K$, $j = 0, 1, 2, ..., \bar{H}-K$, and $c = 0, 1, 2, ..., C-1$, where the kernel $\bW$ can fit into the input tensor $\bX$.

To simplify the representation, we can use the Einstein notation~\footnote{\url{https://en.wikipedia.org/wiki/Einstein_notation}}. Using this notation, we can rewrite the equation as:
\begin{equation}
  \bY^O = \bW_{K,K,C}^{O} \bX^{K,K,C}.
  \label{eq:einsum_conv}
\end{equation}

In essence, for each location in the convolution, it corresponds to a linear projection where the parameter weights are shared among all locations.


Let us discuss the gradients of $\frac{\partial \mathcal{L}}{\partial \bW}$ and $\frac{\partial \mathcal{L}}{\partial \bX}$ respectively. Here, we use $\bY'$ to represent $\frac{\partial \mathcal{L}}{\partial \bY}$.
During the back-propagation process, given $\bY'$, we can calculate the gradients $\bW'$ and $\bX'$ as follows:
\begin{equation}
\begin{aligned}
     \frac{\partial \mathcal{L}}{\partial \bW} &= \bY' {(\bX^{K,K,C})}^{\top}, \\
  \frac{\partial \mathcal{L}}{\partial \bX} &=  {(\bW_{K,K,C}^{O})}^{\top} \bY' .
  \label{eq:einsum_conv_bp} 
\end{aligned}
\end{equation}

Convolution, in essence, is a linear operator that can be applied to multi-dimensional tensors. Hence, we can consider Convolution is a homogeneous operator as a linear layer. Here, homogeneous operator means the Convolution and the linear layer are both first-order linear operator.

In fact, Conv1D can be seen as an equivalence of a linear layer. Additionally, a Conv2D operator can be converted to a matrix multiplication using the im2col~\footnote{\url{https://caffe.berkeleyvision.org/tutorial/layers/im2col.html}} operator. In conclusion, Convolution and Linear Layer (also known as Fully-Connected or FC) are homogeneous operators.

\subsubsection{Normalization}
\label{sec:intro_normalization}
Batch Normalization~\citep{batchnorm_ioffe2015batch} and Layer Normalization~\citep{layernorm_ba2016layer} are widely used techniques in deep learning to improve the training of neural networks.

\textbf{Batch Normalization} (BN)~\citep{batchnorm_ioffe2015batch} is primarily employed in CNNs~\citep{batchnorm_ioffe2015batch, resnet_he2016deep}.

Let us consider a mini-batch $\bX$ with a shape of $D \times N$, where $D$ represents the feature dimension and $N$ is the batch size. The definition of BN is as follows:
\begin{equation}
\begin{aligned} 
\bm{\mu} & = \frac{1}{N} \sum_{i=1}^{N} \bX_{:,i} \\ 
\bm{\sigma}^{2} & = \frac{1}{N} \sum_{i=1}^{N}\left(\bX_{:,i}-\bm{\mu}\right) \odot \left(\bX_{:,i}-\bm{\mu}\right) \\ 
\widehat{\bX}_{:,i} & = \left(\bX_{:,i}-\bm{\mu}\right) \oslash {\sqrt{\bm{\sigma}^{2}+\epsilon}} \\ 
\mathrm{BN}\left(\bX_{:,i} \right) & =  \bm{\gamma} \odot \widehat{\bX}_{:,i} + \bm{\beta},
\end{aligned}
\end{equation}
where $\odot$ and $\oslash$ represent element-wise multiplication and division respectively, $\bm{\mu} \in  \mathbb{R}^{D}$ and $\bm{\sigma} \in  \mathbb{R}^{D}$, $\epsilon$ is a smoothing factor. It's worth noting that we discuss a two-dimensional matrix here, but this can easily be extended to a four-dimensional tensor by reshaping the 4D tensor into a 2D tensor.
In the training process, $\bm{\mu}$ and $\bm{\sigma}$ are updated by a moving average. $\bm{\gamma}$ and $\bm{\beta}$ are optimized by a stochastic gradient descent.

To derive the Jacobian matrix, let us consider 
$\frac{\partial  \widehat{X}_{j,i}}{\partial X_{k,l}}$. When $k \neq j$, its value is 0. 

When $j=k$, let us consider $\frac{\partial \widehat{X}_{j,i}}{\partial X_{j,i}}$,
\begin{equation}
\label{eq:bn_deriv}
\centering
\frac{\partial \widehat{X}_{j,i}}{\partial X_{j,i}}= \frac{(1 - \frac{1}{N}) \sqrt{\sigma_j^2 + \epsilon} - \frac{(X_{j,i} - \mu_j)^2}{N \sqrt{\sigma_j^2 + \epsilon}}}{\sigma_j^2 + \epsilon}
\end{equation}
where $\sigma_j$ is the $j$-th dimension of $\bm{\sigma}$.

Further, where $j=k$ and $l \neq i$, let us consider  $\frac{\partial \widehat{X}_{j,i}}{\partial X_{j,l}}$,

\begin{equation}
\label{eq:bn_deriv2}
\centering
\frac{\partial \widehat{X}_{j,i}}{\partial X_{j,l}}= \frac{(0 - \frac{1}{N}) \sqrt{\sigma_j^2 + \epsilon} - \frac{ (X_{j,i} - \mu_j)(X_{j,l} - \mu_j)  }{N \sqrt{\sigma_j^2 + \epsilon}}}{\sigma_j^2 + \epsilon}
\end{equation}

\textbf{Layer Normalization} (LN)~\citep{layernorm_ba2016layer} has broader applications compared to BN. It is widely utilized in various domains such as Transformers (including Vision Transformers~\citep{wu2021cvt, swintransformer_liu2021swin, vit_dosovitskiy2020image}) and Language Models~\citep{llama_touvron2023llama,opt_zhang2022opt,palm_chowdhery2022palm,gpt_radford2018improving, gpt2_radford2019language, gpt3_brown2020language}), as well as ConvNeXt~\citep{convnext_liu2022convnet}.

To simplify notation, here, let us define $\bx = {\bX_{:, i}}$, where $\bx$ is a column vector.
The forward process of LN with a smoothing factor $\epsilon$ is defined as follows:
\begin{equation}
\begin{aligned} 
\operatorname{LN}(\bx) &= \boldsymbol{\gamma} \odot \bz + \boldsymbol{\beta},
\ \text{where}\ \ \bz = \sqrt{D} \frac{\by}{ \sqrt{{\| \by \|}_2^2 + \epsilon}} \ \ \text{and}\ \  \by = \left(\boldsymbol{I}-\frac{1}{D}  \boldsymbol{1} \boldsymbol{1}^{\top}\right) \bx,
\end{aligned}
\label{equ:ln}
\end{equation}
Where $D$ is the dimension of the input, $\bm{\gamma}$ and $\bm{\beta}$ are learned parameters, similar to BN, obtained through gradient descent. $\epsilon$ is a smoothing factor. It is important to note that in LN, there is no need to maintain a moving average of $\bm{\mu}$ and $\bm{\sigma}$.

The Jacobian matrix of the variable $\bz$ with respect to $\bx$ can be calculated as:
\begin{equation}
\begin{aligned} 
\boldsymbol{J}_{\bz}(\bx) &=\frac{\partial \bz}{\partial \bx} =  \frac{\partial \by}{\partial \bx}  \frac{\partial \bz}{\partial \by} = \frac{\sqrt{D}}{{ \sqrt{{\| \by \|}_2^2 + \epsilon}}} \left(\boldsymbol{I}-\frac{1}{D} \boldsymbol{1} \boldsymbol{1}^{\top}\right)  \left(\boldsymbol{I}-\frac{\by \by^{\top}}{\|\by\|_{2}^{2} + \epsilon }\right).
\end{aligned}
\label{equ:certernorm}
\end{equation}

It should be noted that in this article, we consider a version of LayerNorm that incorporates smoothing, whereas LipsFormer~\cite{lipsformer_qi2023lipsformer} discusses a non-smoothing LayerNorm. The non-smoothing LN is not Lipschitz continuous, while the smoothing LN is Lipschitz continuous but with a very large Lipschitz constant due to a typically small value of $\epsilon$.

LN can be applied to a wider range of deep learning problems than BN, regardless of whether they involve variable-length or fixed-length sequences. On the other hand, BN is more suitable for problems with constant sequence lengths or input sizes. For instance, BN cannot be applied to generative language models because the sequence length increases during the generation process, while LN is a viable choice in such scenarios.

\subsubsection{Self-attention}
\label{sec:intro_selfattention}
Attention mechanism is firstly introduced in~\cite{attention_bahdanau2014neural}, and then widely used in CV and NLP areas.
Dot-product attention (DPA)~\citep{transformer_vaswani2017attention} is a crucial component in Transformer, enabling the capture of long-range relationships within data. 
In practice, multi-head attention is employed to effectively capture such relationships in different contexts. The formulation of single-head attention is as follows:
\begin{equation}
\operatorname{Attn\_DP}(\boldsymbol{X}; \boldsymbol{W}^{Q}, \boldsymbol{W}^{K}, \boldsymbol{W}^{V}) = \boldsymbol{W}^{V} \boldsymbol{X} \cdot \mathcal{S}\left(\frac{{\left(\boldsymbol{W}^{Q} \boldsymbol{X}\right)}^{\top} \boldsymbol{W}^{K} \boldsymbol{X}}{\sqrt{D}}\right),
\end{equation}
where $\boldsymbol{W}^{Q}, \boldsymbol{W}^{K}, \boldsymbol{W}^{V}$ are the projection matrices used to transform $\boldsymbol{X}$ into query, key, and value matrices, respectively. $\mathcal{S}(\cdot)$ denotes the softmax function. Intuitively, each token aggregates information from all visible tokens by calculating a weighted sum of the values of visible tokens based on the similarity between its query and the key of each visible token.
The similarity between the $i$-th query $\bq_{i}$ and the $j$-th key $\bk_{j}$ is represented as $\boldsymbol{P}_{ij} \propto {\boldsymbol{x}_{i}}^{\top} (\boldsymbol{W}^{Q})^{\top} {\boldsymbol{W}^{K}} {\boldsymbol{x}_{j}}.$

Usually, multi-head self-attention is used in practice. For the $i$-th attention head, where $i \in \{1, ..., H\}$, we define it as follows:
\begin{equation*}
\boldsymbol{h}_i(\bx; \bm{\sW}_i) = {\operatorname{Attn\_DP}_i}(\boldsymbol{X}; \boldsymbol{W}_{i}^Q, \boldsymbol{W}_i^{K}, \boldsymbol{W}_i^{V}),
\end{equation*}
where $\bm{\sW}_i$ represents the set of projection weight matrices $(\boldsymbol{W}_i^{Q}, \boldsymbol{W}_i^{K}, \boldsymbol{W}_i^{V})$.

In multi-head attention, the different attention results are concatenated as follows:
\begin{equation}
   {\boldsymbol{h}(\bx; \bm{\sW})} = [\boldsymbol{h}_1(\bx; \bm{\sW}_1); \ \boldsymbol{h}_2(\bx; \bm{\sW}_2); \  ...; \  \boldsymbol{h}_H(\bx; \bm{\sW}_H)].
\end{equation}

Compared to linear layers and convolutions, self-attention is a high-order nonlinear function. It exhibits high-order nonlinearity due to the following aspects: 
1) It captures complex dependencies between elements in the input sequence.
2) It employs a nonlinear softmax activation function to compute attention scores, resulting in a nonlinear relationship between the input sequence and the output.
3) The query, key, and value projections create a high-dimensional space where interactions between elements become more complex, contributing to the high-order nonlinearity of the function.

This high-order nonlinearity enables self-attention to effectively model intricate relationships and dependencies within the input data. We will discuss more attention mechanisms in Section~\ref{sec:opt_selfattention}.

\subsubsection{Residual Shortcut}
\label{sec:intro_residual}
Residual shortcut~\citep{resnet_he2016deep, identity_mapping_he2016identity} is a breakthrough technology that effectively addresses the vanishing gradient problem. Prior to its introduction, deep neural networks faced significant challenges with vanishing gradients~\citep{vgg_simonyan2014very,googlenet_szegedy2015going}. Back to 2014, training a 16-layer VGG network was difficult. However, since 2015, ResNet50 has become a standard and fundamental configuration for deep learning researchers. 
Subsequent neural network architectures have widely adopted the residual shortcut.

The residual shortcut is defined as:
\begin{equation}
    \by = \bx + f(\bx; \bW).
\label{eq-resnetshortcut2}
\end{equation}

The Jacobian matrix of $\by$ with respect to $\bx$ is:
\begin{equation}
\begin{aligned} 
\boldsymbol{J}_{\by}(\bx) = \frac{\partial \by}{\partial \bx} = \frac{\partial \bx}{\partial \bx} + \frac{\partial f(\bx; \bW)}{\partial \bx} = \bI + \frac{\partial f(\bx; \bW)}{\partial \bx}.
\end{aligned}
\label{equ:residualshortcutjacobian}
\end{equation}

Given $\frac{\partial \mathcal{L}}{\partial \by}$, even when $\frac{\partial f(\bx; \bW)}{\partial \bx}$ is very small, $\frac{\partial \mathcal{L}}{\partial \bx} \approx \frac{\partial \mathcal{L}}{\partial \by}$, meaning the error information can still be propagated to shallow layers through $\frac{\partial \mathcal{L}}{\partial \bx}$. Without the residual shortcut, the gradient information would be blocked at this layer.

\subsubsection{Activation}
\label{sec:intro_activation}
Activation~\citep{relu_dahl2013improving, prelu_he2015delving, gelu_hendrycks2016gaussian, glu_shazeer2020glu} is an effective method for introducing nonlinearity to neural networks. Among various types of activation functions, ReLU is widely used. It is defined as:
\begin{equation}
\by = \max \left(\bold{0}, \bx \right).
\end{equation}

The Jacobian matrix of ReLU is:
\begin{equation}
\boldsymbol{J}_{\by}(\bx) = \frac{\partial \by}{\partial \bx} = \operatorname{diag} \left( \displaystyle \1\left({\bx > \bold{0}}\right)\right),
\end{equation}
where $\1(\cdot)$ is the indicator function.

Compared to the Sigmoid and Tanh activations, ReLU preserves the gradients better, and thus mitigate the gradient vanishing problem well. 
ReLU is a non-smooth function at the point of zero. The generalization ability of non-smooth functions requires further investigation. Meanwhile, according to the classical numerical optimization, non-smooth function will lead to a slower convergence rate.
After ReLU, several extensions have been proposed, including PReLU~\citep{prelu_he2015delving}, Swish~\citep{swish_ramachandran2017searching}, GeLU~\citep{gelu_hendrycks2016gaussian}, and GLU~\citep{glu_shazeer2020glu}. 

\subsubsection{Feed-Forward Network}
\label{sec:ffn_introductin}

In Transformer architecture~\citep{transformer_vaswani2017attention}, in addition to self-attention, Feed-Forward Network (FFN) is another key component. An FFN is defined as:
\begin{equation} 
\by = \operatorname{FFN}(\bx; \bW_1, \bW_2, \bb_1, \bb_2) = \bW_{2} \max \left(\bold{0}, \bW_{1} \bx + \bb_{1}\right) + \bb_{2},
\label{eq:ffn_equation}
\end{equation}

where $\bW_1, \bW_2, \bb_{1}, \bb_{2}$ are learned parameters. FFN is a composition of linear projections and the ReLU activation function.

In FFN, $\bW_1$ is typically a large weight matrix that projects $\bx$ (with dimension $D$) into a higher dimensional space (usually $4D$), and then $\bW_2$ projects the high-dimensional feature back to the same dimension as $\bx$.

The Jacobian matrix of an FFN can be computed as:
\begin{equation} 
\boldsymbol{J}_{\by}(\bx) = \frac{\partial \operatorname{FFN}(\bx; \bW_1, \bW_2, \bb_1, \bb_2)}{\partial \bx} = {\bW_1}^{\top} \operatorname{diag} { \left( \1\left({ \bW_1  \bx+ \bb_1>\bold{0}}\right) \right)} {\bW_2}^{\top}.
\end{equation}

Similar to self-attention, FFN is typically used in conjunction with a residual shortcut. With the residual shortcut, we have $\by = \bx + \operatorname{FFN}(\bx; \bW_1, \bW_2, \bb_1, \bb_2)$, and its Jacobian matrix is $\boldsymbol{J}_{\by}(\bx) = \bI +  {\bW_1}^{\top} \operatorname{diag} { \left( \1\left({ \bW_1  \bx+ \bb_1>\bold{0}}\right) \right)} {\bW_2}^{\top}$.

So far, we have briefly introduced some basic modules in deep learning and derived their Jacobian matrices, which will be used in back-propagation.

\subsection{ResNet and Transformer}
\label{sec:intro_resnetandtransformer}
Based on the above introduction, we are now ready to discuss the renowned ResNet~\citep{resnet_he2016deep} and Transformer~\citep{transformer_vaswani2017attention}. ResNet allows the training of extremely deep convolutional networks by addressing the vanishing gradient problem through the use of residual connections. This enables the networks to effectively learn complex representations.

Transformer was initially introduced in natural language processing and has since expanded its capabilities to other modalities, including images, audio, video, and 3D vision. The Transformer architecture has opened doors for models that can handle diverse data types within a unified framework. Recent advancements, such as CLIP~\citep{clip_radford2021learning} and DALL·E~\citep{dalle2_ramesh2022hierarchical}, further demonstrate the flexibility and potential of the Transformer architecture in handling multi-modal data. Interested readers can refer to~\cite{survey_tay2022efficient, survey_lin2022survey} for a overview survey of Transformer.

\begin{theo}[ResNet vs Transformer]{thm:resnetandtransformer}
\label{remark:resnetandtransformer}
\begin{enumerate}[leftmargin=*]
\item Linear projection and convolution, in nature, are two similar linear first-order operators. They are \textit{homogeneous}. On the other hand, self-attention is a nonlinear high-order operator. It is a \textit{heterogeneous} operator compared to linear projection and convolution.
\item ResNet is composed of homogeneous operators, while Transformer is composed of heterogeneous operators. ResNet focuses on describing local regions, while Transformer captures longer and larger contextual information.
\item The Jacobian matrices of homogeneous and heterogeneous networks have very different properties, which result in different optimization difficulties. The heterogeneous Transformer is much harder to optimize.
\end{enumerate}
\end{theo}

In Remark~\ref{remark:resnetandtransformer}, we have provided several remarks to discuss and compare ResNet and Transformer. Heterogeneous operators mean these modules have very different Jacobian properties. For example, self-attention and linear layer are heterogeneous operators because linear layer is a first-order linear operator but self-attention is a high-order nonlinear operator.

\subsection{Forward and Backward in Neural Network}
\label{sec:intro_forwardandbackwardinnn}
In this section, we provide a brief definition of the forward and backward processes in neural networks, which are fundamental concepts in deep learning.

Given an input $\bx$, it is passed through a series of hidden layers, with each layer performing a specific transformation. The output $\bx^{l+1}$ of layer $l+1$ is calculated based on the input $\bx^l$ from the previous layer, and $\bx^l$ is calculated based on its corresponding input $\bx^{l-1}$. This implies that $\bx^l$ is influenced by the weight matrices $\bW^1$ to $\bW^l$. We can define the forward process to obtain $\bx^l$ as follows:
\begin{equation}
\centering
\bx^l = \mathcal{F}\left(\bx; \bW^{1}, \ldots, \bW^{l}\right).   
\end{equation}
During the forward process, $\bx^l$ is not influenced by the weight matrices in the subsequent layers, such as $\bW^{l+1}$.

Through back-propagation and the chain rule, we can define the backward view of deep learning as follows:
\begin{equation}
\begin{aligned}
    \frac{\partial \mathcal{L}}{\partial \bW^{l}} & = \nabla_{\bW^l} \mathcal{L}\left(\bx, \by; \bW^{1}, \ldots, \bW^{L} \right), \\
    \frac{\partial \mathcal{L}}{\partial \bx^{l}} & = \nabla_{\bx^l} \mathcal{L}\left(\bx, \by; \bW^{1}, \ldots, \bW^{L}\right),
\end{aligned}    
\end{equation}
These equations will be further discussed in the following section. The forward and backward views form the basic principles of optimization in deep learning.

\subsubsection{An Example to Understand Jacobian Matrix}
\label{sec:intro_an_example}
\begin{figure}[thb] 
    \centering
    \includegraphics[width=0.6\textwidth]{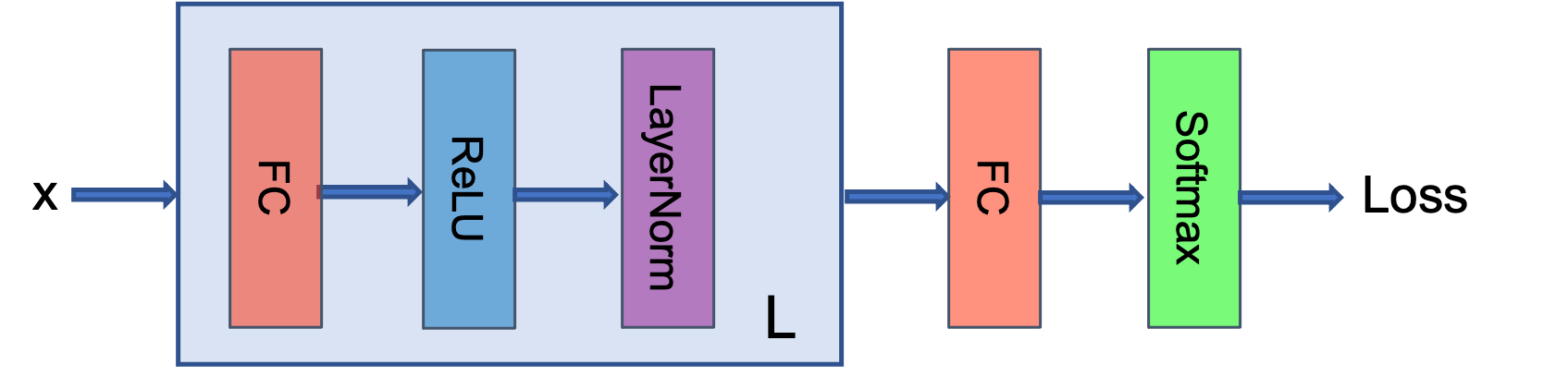}
    \caption{Visualization of a simple neural network with $L$ layers and a classification layer.} 
    \label{fig:simplenet}
\end{figure}

To gain a deeper understanding of the forward and backward processes, let us consider an example. The network structure is illustrated in Figure~\ref{fig:simplenet}.

In this example, $\bx$ represents the input data. The network consists of $L$ layers in the stem, where each layer contains three submodules: a fully connected (FC), a rectified linear unit (ReLU), and a layer normalization (LayerNorm). Finally, we have a classification layer,  a softmax, and a cross-entropy  to compute the final loss $\mathcal{L}$.

Mathematically, we can define the \textit{forward process} as follows:
\begin{equation}
\begin{aligned}
{\bx}^{1} &= \operatorname{LN}(\operatorname{ReLU}(\bW^1 {\bx})) \\
\vdots \\
{\bx}^{l+1} &= \operatorname{LN}(\operatorname{ReLU}(\bW^{l+1} {\bx}^{l})) \\
\vdots \\
{\bx}^{L} &= \operatorname{LN}(\operatorname{ReLU}(\bW^{L} {\bx}^{L-1})) \\
\bo &= \bO \bx^{L} \\
\bp &= \operatorname{softmax}({\bo})\\
\mathcal{L} &= \operatorname{cross\_entropy}(\bp, \bt),
\end{aligned}
\end{equation}

Here, $\bO$ represents the classification weight matrix, and $(\bx, \bt)$ denote the input and the label, respectively. We omit bias term for easier understanding.

To facilitate the subsequent derivations, let us define $\bh^{l+1} = \bW^{l+1} {\bx^{l}}$ and $\br^{l+1} = \operatorname{ReLU}(\bh^{l+1})$. For convenience, we rewrite LayerNorm equation as:
\begin{equation}
\bx^{l+1} = \operatorname{LN}(\br^l) = \boldsymbol{\gamma^{l+1}} \odot \bz^{l+1} + \boldsymbol{\beta^{l+1}},
\end{equation}
where $\bz^{l+1} = \sqrt{D} \frac{\by^{l+1}}{{\sqrt{{\| \by^{l+1} \|}_2^2 + \epsilon}}}$ and $\by^{l+1} = \left(\boldsymbol{I}-\frac{1}{D}  \boldsymbol{1} \boldsymbol{1}^{\top}\right) \br^{l+1}$.

In the \textit{backward process}, we calculate $\frac{\partial \mathcal{L}}{\partial \bx^l} $ and $\frac{\partial \mathcal{L}}{\partial \bW^l}$ in a reverse order, starting from the $L$-th layer and moving towards the $1$-st layer. We can compute $\frac{\partial \mathcal{L}}{\partial \bx^l} $ as follows:
\begin{equation}
\small
\label{eq:backward_example}
\begin{aligned} 
\frac{\partial \mathcal{L}}{\partial \bx^l} &= \frac{\partial \bx^{l+1}}{\partial \bx^{l}}  \frac{\partial \mathcal{L}}{\partial \bx^{l+1}}\\ 
& = \frac{\partial \bx^{l+1}}{\partial \bx^{l}} \ldots  \frac{\partial \bx^{L}}{\partial \bx^{L-1}}   \frac{\partial \mathcal{L}}{\partial \bx^{L}}\\
& =    \frac{\partial \bx^{l+1}}{\partial \bx^{^l}} \ldots \frac{\partial \bx^{L}}{\partial \bx^{L-1}}  \frac{\partial \bo}{\partial \bx^{L}}  \frac{\partial \mathcal{L}}{\partial \bo}\\
& =\left( \prod \limits_{k=l+1}^{L} \frac{\partial \bx^{k}}{\partial \bx^{k-1}}  \right)   \frac{\partial \bo}{\partial \bx^{L}} \frac{\partial \mathcal{L}}{\partial \bo} \\
& =\left( \prod \limits_{k=l+1}^{L}    \eqnmarkbox[red]{Psi2}{  \frac{\partial \bh^{k}}{\partial \bx^{k-1}} } 
\eqnmarkbox[blue]{Psi2}{\frac{\partial \br^{k}}{\partial \bh^{k}}}
 \eqnmarkbox[purple]{Psi2}{ \frac{\partial \bx^{k}}{\partial \br^{k}} }
\right) \eqnmarkbox[red]{Psi2}{\frac{\partial \bo}{\partial \bx^{L}} }  
\eqnmarkbox[green]{Psi2}{ \frac{\partial \mathcal{L}}{\partial \bo} }  
\\
& =\left( \prod \limits_{k=l+1}^{L}  {\eqnmarkbox[red]{Psi2}{{\left(\bW^k\right)}^{\top}}}
\eqnmarkbox[blue]{Psi2}{ \operatorname{diag} \left( { \1\left({\bW^k \bx^{k-1}>\bold{0}}\right)}\right)} 
  \eqnmarkbox[purple]{Psi2}{ \frac{\sqrt{D} \left(\boldsymbol{I}-\frac{1}{D} \boldsymbol{1} \boldsymbol{1}^{\top}\right)}{\sqrt{{\| \by^{k} \|}_2^2 + \epsilon}}  \left(\boldsymbol{I}-\frac{\by^{k} {\by^{k}}^{\top}}{\|\by^{k}\|_{2}^{2} + \epsilon}  \right) \operatorname{diag}\left( \bm{\gamma^k} \right)} 
\right)
\eqnmarkbox[red]{Psi2}{{\bO}^{\top}} 
 \eqnmarkbox[green]{Psi2}{(\bp - \bt)} .
\end{aligned}    
\end{equation}

$\frac{\partial \mathcal{L}}{\partial \bo}$ represents the gradient of the softmax and cross-entropy loss. Its value is $\bp - \bt$, which can be found in deep learning books~\cite{book_goodfellow2016deep}. We will not provide a detailed derivation here.

Once we have obtained $\frac{\partial \mathcal{L}}{\partial \bx^l}$, we can further calculate $\frac{\partial \mathcal{L}}{\partial \bW^{l+1}}$ as follows:
\begin{align} 
\frac{\partial \mathcal{L}}{\partial \bW^{l+1}} = \frac{\partial \mathcal{L}}{\partial \bh^{l+1}} {\bx^{l}}^{\top}.
\end{align}

Now, let us delve into Equation~\ref{eq:backward_example}. The range of $\frac{\partial \mathcal{L}}{\partial \bx^l}$ depends on five terms, represented by different colors: \textcolor{red}{red}, \textcolor{blue}{blue}, \textcolor{purple}{purple}, \textcolor{red}{red}, and \textcolor{green}{green}. All these five terms denote the Jacobian matrices of their corresponding modules.
Among these terms, three of them (\textcolor{red}{red}, \textcolor{blue}{blue}, and \textcolor{purple}{purple}) appear multiple times. The last term (\textcolor{green}{green}) is bounded by the range [-1.0, 1.0]. The second-to-last term in \textcolor{red}{red} represents the classification weight matrix and appears only once. Regarding the first three terms in the brackets, the \textcolor{blue}{blue} term, i.e., the ReLU term, corresponds to a contraction mapping. The range of the \textcolor{purple}{purple} term, i.e., the LayerNorm term, depends on the input data and activations, while the range of the \textcolor{red}{red} term, i.e., the FC term, depends on the weight matrix, particularly the eigenvalues of the weight matrix. 

\textit{In conclusion, Back-propagation is a composition function of Jacobian matrices of different modules. This example explains part of our analysis in Figure~\ref{fig:optimization_overview} that Jacobian matrix determines the training process.}

\subsection{Network Initialization}
\label{sec:intro_net_initialization}
Now that we have a network structure, it is crucial to properly initialize the network parameters. Initialization plays a crucial role in training neural networks. Xavier initialization~\citep{xavier_init_glorot2010understanding}, which emerged as a breakthrough, provides insights into the challenges of training deep networks.

Xavier initialization recommends two types of initialization. The first one is defined as follows:
\begin{equation}
W_{i, j} \sim \operatorname{U}\left(-\sqrt{\frac{6}{n_{in}+n_{out}}}, \sqrt{\frac{6}{n_{in}+n_{out}}}\right),
\end{equation}

Here, $\operatorname{U}(\cdot)$ represents a uniform distribution. The second type is defined as:
\begin{equation}
W_{i, j} \sim \operatorname{N}\left(0, \frac{2}{n_{in}+n_{out}}\right),
\end{equation}
where $\operatorname{N}(\mu, \sigma^2)$ denotes a Gaussian distribution with $\mu$ as the mean and $\sigma^2$ as the variance.
Here, $n_{in}$ represents the dimension of the input, and $n_{out}$ is the dimension of the output.

Xavier initialization has made training deeper neural networks possible. Fruthermore, the introduction of Batch Normalization (BN) has reduced the sensitivity of convolutional networks to weight initialization. BN enables the training of deeper networks. Additionally, the use of residual shortcuts has further made it feasible to train convolutional neural networks with thousands of layers.

\begin{figure}[h]
\centering
\begin{subfigure}{.33\textwidth}
  \centering
  \includegraphics[width=.95\linewidth]{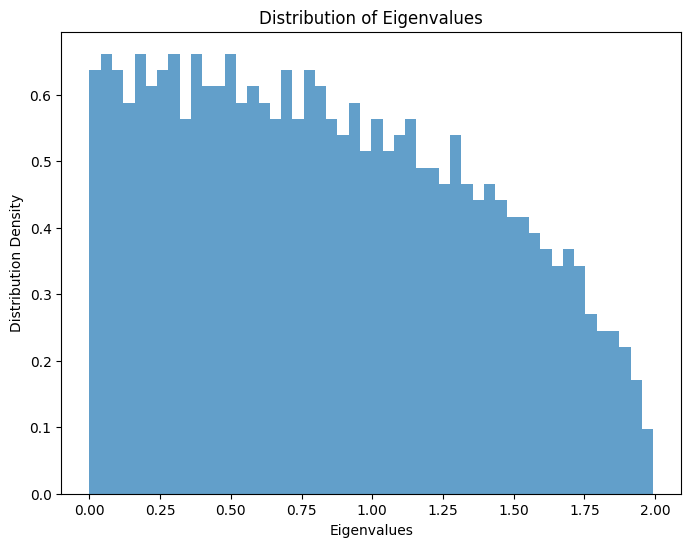}
  \caption{$n_{in}=1024$ and $n_{out} = 1024$.}
  \label{fig:eigenvalue_after_xavier_3}
\end{subfigure}%
\begin{subfigure}{.33\textwidth}
  \centering
  \includegraphics[width=.95\linewidth]{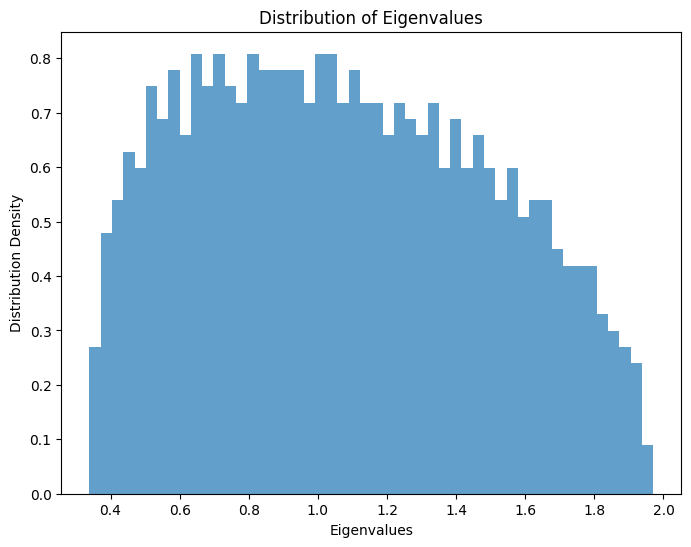}
  \caption{$n_{in}=1024$ and $n_{out} = 2048$.}
  \label{fig:eigenvalue_after_xavier_1}
\end{subfigure}%
\begin{subfigure}{.33\textwidth}
  \centering
  \includegraphics[width=.95\linewidth]{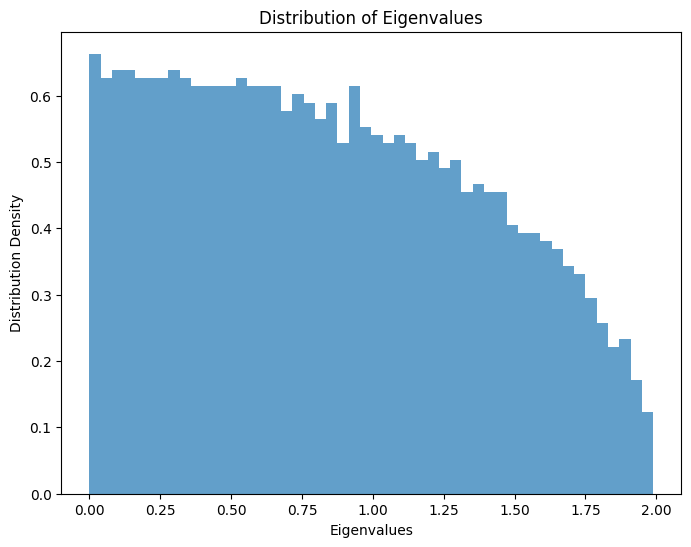}
  \caption{$n_{in}=2048$ and $n_{out} = 2048$.}
  \label{fig:eigenvalue_after_xavier_2}
\end{subfigure}
\caption{Distribution of eigenvalues after Xavier initialization. The left figure shows the distribution of $n_{in}=1024$ and $n_{out} = 1024$. The middle figure shows the distribution of $n_{in}=1024$ and $n_{out} = 2048$. The right shows the eigenvalue distribution of $n_{in}=2048$ and $n_{out} = 2048$.}
\label{fig:eigenvalue_after_xavier}
\end{figure}

Figure~\ref{fig:eigenvalue_after_xavier} shows three eigenvalue distributions of the weight matrix after Xavier initialization, one for $n_{in}=1024$ and $n_{out} = 1024$, one for $n_{in}=1024$ and $n_{out} = 2048$, and one for $n_{in}=2048$ and $n_{out} = 2048$. We have two observations. First, if $\bW$ is a square matrix, there will be many eigenvalues close to zero. Second, after Xavier initialization, the maximum eigenvalue is always less than 2.0.

\subsection{Formulation of Optimization Problem in Deep Learning}
\label{sec:intro_formulation_of_opt}
Generally, a deep neural network can be seen as an unconstrained optimization problem, with the exception of certain special cases like  optimization on sphere. Typically, we formulate the problem as follows:
\begin{equation}
\min_{\theta} \frac{1}{N} \sum_{i=1}^{N} \mathcal{L} \left(\by_{i}, F\left(\bx_{i}; \bm{\sW}\right)\right) + \sum_{l=1}^{L} \lambda \| \bW^l \|_{2}^{2},
\end{equation}

where $N$ represents the batch size of inputs, $\bx_{i}$ denotes the $i$-th input, $L$ is the number of layers, $\bW^l$ represents the weights of the $l$-th layer, and $\bm{\sW}$ represents the set of all weights $\{\bW^1, \bW^2, \ldots, \bW^L\}$, $\mathcal{L}(\cdot,\cdot)$ is the loss function.

Gradient Descent-based methods such as SGD are the classical optimization methods used for training deep neural networks. The weight update rule for these methods can be expressed as:
\begin{equation}
\bW_{new}^l = \bW^l - \alpha \nabla_{\bW^l} \mathcal{L} - \alpha \lambda \bW^l,
\label{eq:sgd}
\end{equation}
here, $\alpha$ represents the learning rate, and $\lambda$ denotes the weight decay factor.  They can be set as functions of various factors to control the optimization process effectively.

According to whether the optimization operator is directly applied to Eq~\ref{eq:sgd}, we have the following definition:

\begin{center}
    \normalsize \textit{Explicit optimization refers to operations directly conducted on $\bW$, $\displaystyle \nabla_{\bW} \mathcal{L}$, $\alpha$, and $\lambda$, \\ while the opposite is implicit optimization.}
\end{center}

\begin{table}[h]
\centering
\begin{tabular}{||c | c | c | c||} 
 \hline
 Weight & Gradient & Learning rate & Weight decay factor \\[0.5ex] 
 \hline
 $\bW$ & $\displaystyle \nabla_{\bW} \mathcal{L}$ & $\mathbf{\alpha}$ & $\lambda$ \\ [0.5ex] 
 \hline
\end{tabular}
\caption{Explicit optimization variables.}
\label{table:explicit_optimization}
\end{table}

More precisely, we categorize the optimization of deep neural networks into two types: explicit optimization and implicit optimization. The optimization space of explicit optimization is summarized in Table~\ref{table:explicit_optimization}. On the other hand, implicit optimization encompasses a wide range of techniques. It includes achieving normalized gradients for weights through activation, normalization, preventing vanishing gradients using residual shortcuts, preserving Lipschitz smoothing through strong data augmentation, and more. In later sections, we will discuss implicit optimization techniques in deep learning, as well as some standard methods in explicit optimization.

\begin{algorithm}
	\caption{SGD with momentum and decoupled weight decay} 
    \hspace*{\algorithmicindent} \textbf{Input: learning rate schedular $\alpha_t$, weight decay $\lambda$, and momentum $\beta$} \\
    \hspace*{\algorithmicindent} \textbf{Output: updated weight $\bw_T$} 
	\begin{algorithmic}[1]
		\For {$t=1,\ldots, T$}
		    \State $\alpha_{t} \leftarrow \operatorname{SetScheduleMultiplier}(t)$
		    \State $\bg_{t}=\nabla_{\bw} \mathcal{L}\left(\mathbf{w}_{t}, \boldsymbol{\theta}, t\right)$

				\State ${\color{blue}{\bv_{t}}}=
				\beta {\color{blue}{\bv_{t-1}}} + (1-\beta) \bg_{t}$

       \If {Weight Decay is Yes} 
			    \State {\jingAlgo{\boldsymbol{w}_{t}} $\boldsymbol{w}_{t-1} - \alpha_{t}{\color{blue}{\bv_{t}}} - \alpha_{t} \lambda \boldsymbol{w}_{t-1}$}
           \Else 
                    \State {\jingAlgo{\boldsymbol{w}_{t}} $\boldsymbol{w}_{t-1} - \alpha_{t} \color{blue}{{\bv_{t}}}$}
		\EndIf		    
		\EndFor
        \end{algorithmic} 
\label{alg:sgd}
\end{algorithm}

\subsection{Popular Optimizers in Deep Learning}
\label{sec:intro_optimizer}
SGD (Stochastic Gradient Descent)~\citep{sgd_robbins1951stochastic} is a fundamental optimization algorithm widely used in deep learning. It updates a model's parameters based on the gradient of the loss function with respect to the parameters. While SGD is fast and simple, it can struggle to converge to the global minimum.
Momentum SGD (mSGD)~\citep{msgd_nesterov1983method} enhances SGD by incorporating a momentum term in the update rule. This term helps the optimizer move more swiftly through shallow areas of the loss function, preventing it from getting trapped in local minima. It proves useful when the loss function exhibits many plateaus or valleys. We have shown the SGD with momentum in Algorithm~\ref{alg:sgd}.

Currently, adaptive learning rate optimization algorithms such as Adagrad~\citep{adagrad_duchi2011adaptive}, RMSProp~\citep{rmsprop_hinton2012rmsprop}, Adam~\citep{adam_kingma2014adam}, and AdamW~\citep{adamw_loshchilov2017decoupled} dominate neural network training, particularly with the widespread use of Transformers across different modalities. Adagrad~\citep{adagrad_duchi2011adaptive} is the first adaptive learning rate optimization algorithm that adjusts the learning rate for each parameter based on its past gradients. This adaptivity proves beneficial for sparse datasets, where some parameters may have large gradients while others have small gradients.
RMSProp~\citep{rmsprop_hinton2012rmsprop} is another adaptive learning rate algorithm that adjusts the learning rate based on the root mean square of past gradients. It helps prevent the learning rate from becoming excessively large or small.
Adam combines the concepts of momentum and adaptive learning rates. It calculates exponentially decaying averages of past gradients and their squares to update the learning rate for each parameter. Adam demonstrates good performance across a wide range of problems and is currently one of the most popular optimization algorithms.

\begin{algorithm}
	\caption{AdamW} 
    \hspace*{\algorithmicindent} \textbf{Input: learning rate schedular $\alpha_t$, weight decay $\lambda$, and first-order and second-order mementums  $\beta_1$, $\beta_2$} \\
    \hspace*{\algorithmicindent} \textbf{Output: updated weight $\bw_T$} 
	\begin{algorithmic}[1]
		\For {$t=1,2,\ldots, T$}
		    \State $\alpha_{t} \leftarrow \operatorname{SetScheduleMultiplier}(t)$
		    \State $\boldsymbol{g}_{t}=\nabla_{\boldsymbol{w}} \mathcal{L}\left(\boldsymbol{w}_{t}, \boldsymbol{\theta}, t\right)$

		            \State ${{{{{\boldsymbol{m}}_{t}}}} = \beta_{1} \boldsymbol{m}_{t-1}+\left(1-\beta_{1}\right) \boldsymbol{g}_{t}}$ 
		            
		            \State ${{{{\boldsymbol{v}}_{t}}}} = \beta_{2} \boldsymbol{v}_{t-1}+\left(1-\beta_{2}\right) \boldsymbol{g}_{t}^{2}$ \Comment{ ${{{\boldsymbol{m}}_{t}}}$ and ${{{\boldsymbol{v}}_{t}}}$ will be cached.}
		            
		            \State ${{{{\hat{\boldsymbol{m}}_{t}}}} = \boldsymbol{m}_{t} /\left(1-\beta_{1}^{t}\right)}$
		            
		            \State ${{{{\hat{\boldsymbol{v}}_{t}}}} = \boldsymbol{v}_{t} /\left(1-\beta_{2}^t\right)}$
		            
		            \State ${{{\color{green}{\boldsymbol{\mu}_{t}}}}} = {\hat{\boldsymbol{m}}_{t}}/{\hat{\boldsymbol{v}}_{t}}$ \Comment{ ${\color{green}{\boldsymbol{\mu}_{t}}}$ is the final gradient used for update.}

			    \If {Weight Decay is Yes} \Comment{ Some variables are weight decay free.}
			        \State {\jingAlgo{\boldsymbol{w}_{t}} $\boldsymbol{w}_{t-1} - \alpha_{t}{\color{green}{\boldsymbol{\mu}_{t}}} - \alpha_{t} \lambda_{t} \boldsymbol{w}_{t-1}$}
           \Else 
            \State {\jingAlgo{\boldsymbol{w}_{t}} $\boldsymbol{w}_{t-1} - \alpha_{t} \color{green}{{\boldsymbol{\mu}_{t}}}$}
			    \EndIf
		            
		\EndFor
	\end{algorithmic} 
 \label{alg:adam}
\end{algorithm}

AdamW~\citep{adamw_loshchilov2017decoupled} is an extension of Adam that rectifies the flawed $L_2$ regularization technique by incorporating weight decay. In the original version, the $L_2$ regularization is applied to the gradient, whereas in the corrected weight decay version, it is directly applied to the weight matrix. This correction further improves the performance of Adam.

In Algorithm~\ref{alg:sgd} and Algorithm~\ref{alg:adam}, we have presented SGD and AdamW with momentum and decoupled weight decay. Now, let us discuss the advantages and disadvantages of these methods. 
In addition to SGD, there are several improved optimizers that build upon it. Examples include SGDR~\citep{sgdr_loshchilov2016sgdr}, SVRG~\citep{svrg_johnson2013accelerating}, and signSGD~\citep{signsgd_bernstein2018signsgd}. These optimizers aim to enhance the performance of SGD in various ways.
For adaptive learning rate optimization, there are multiple algorithms designed to improve upon it. AdaHessian~\citep{adahessian_yao2020adahessian}, Adabelief~\citep{adabelief_zhuang2020adabelief}, and Adafactor~\citep{adafactor_shazeer2018adafactor} are a few examples. These algorithms aim to refine the adaptive learning rate mechanisms to achieve better optimization results.
Furthermore, there are specific optimizers~\citep{you_you2019large, lars_you2017scaling} tailored for addressing large-scale optimization problems. These optimizers are designed to handle the challenges posed by datasets and models of significant size.
Each optimizer brings its own set of benefits and considerations, depending on the specific characteristics of the problem at hand. We will discuss the optimizers in Section~\ref{sec:explicit_optimization} in detail.

\subsection{Mixed Precision Training}
\label{sec:intro_mpt}
Mixed Precision Training (MPT)~\citep{mixedprecision_micikevicius2017mixed}\footnote{{\scriptsize \url{https://docs.nvidia.com/deeplearning/performance/mixed-precision-training/index.html}}} is a widely adopted strategy in deep learning that brings significant speedup and memory savings during training. This technique enables researchers and practitioners to train larger and more complex models effectively.

In computer memory, a single-precision floating point (also known as FP32 or float32) typically occupies 32 bits. It offers a wide range of numeric values by utilizing a floating radix point. The minimum positive normal value is \(2^{-126} \approx 1.18 \times 10^{-38}\), and the largest positive normalized FP32 number $2^{128}\left(1-2^{-24}\right) \approx 3.4 \times 10^{38}$.

On the other hand, half precision (also referred to as FP16 or float16) is a floating-point format that occupies 16 bits (two bytes) in computer memory. It provides a more compact representation. The minimum positive normal value is \(2^{-14} \approx 6.10 \times 10^{-5}\). The maximum representable value is \(\left(2-2^{-10}\right) \times 2^{15} = 65504\). The calculation of FP16 values follows the equation:
\begin{equation}
(-1)^{\text{Sign}} \times 2^{\text{Exponent}-15} \times \left(1 + \frac{\text{Fraction}}{1024}\right).
\end{equation}

\begin{figure}[thb] \centering
    \includegraphics[width=0.65\textwidth]{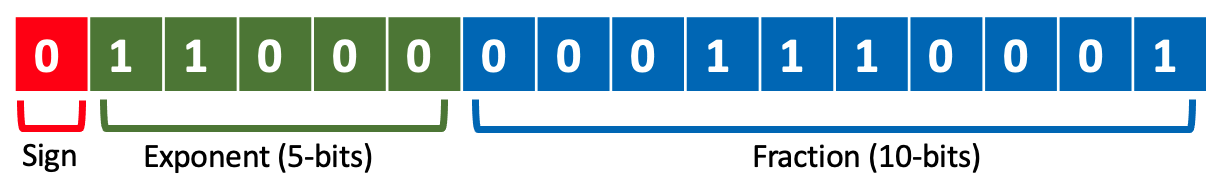}
    \caption{Visualization of Half-precision floating-point format (FP16).} 
    \label{fig:fp16visual}
\end{figure}

To visualize the format of half precision, refer to Figure~\ref{fig:fp16visual}.

In MPT, half-precision (FP16) is utilized for both storage and arithmetic operations, with weights, activations, and gradients being stored in FP16. However, a single-precision (FP32) master copy of weights is employed for updates. There are several advantages to use numerical formats with FP16 over FP32. Firstly, lower precision formats like FP16 consume less memory, which enables the training and deployment of larger neural networks. The reduced memory footprint is particularly beneficial when dealing with models that have a large number of parameters.
Secondly, FP16 formats demand less memory bandwidth, resulting in accelerated data transfer operations. This efficiency in memory usage improves the overall performance and speed of the training process.
Thirdly, mathematical operations execute much faster in reduced precision, especially on GPUs equipped with Tensor Core support specifically designed for the given precision. The utilization of FP16 allows for quicker computations, leading to enhanced training speed and efficiency.
In practical applications, the adoption of mixed precision training can yield substantial speedups, with reported gains of up to 3x on architectures like Volta and Turing. By leveraging FP16 for storage and arithmetic operations, combined with an FP32 master copy for weight updates, the training process is optimized, resulting in improved overall performance.

Regarding the question of why one needs to make an FP32 copy for weight matrices in MPT, let us consider the following example. Suppose the learning rate is 1e-5 and the weight value is 1e-4. The multiplication of these values is 1e-9. However, in the FP16 representation, 1e-9 is rounded to zero. Therefore, using weight matrices in FP16 alone is insufficient to represent such small variations accurately. To ensure the preservation of fine-grained details, an FP32 copy of the weight matrix is necessary.

\section{Optimization Principles of Deep Learning}
\label{sec:optimization_principles_deeplearning}
In this section, we will define two fundamental principles of optimization in deep learning. Before that, let us further review several definitions and lemmas that will be used in the optimization principles.

\subsection{Lipschitz Constant of Deep Neural Network}
\begin{definition} 
\label{def-dnn}
Let $F(\bx; \{\bW_l, l=1,\ldots,L \}): \mathbb{R}^{D} \rightarrow \mathbb{R}$ be an L-layer neural network defined as a composite function with L transformation functions:
\begin{equation} 
F(\bx; \{\bW^l; l=1,\ldots,L \})=f^{L}\left(\bW^L f^{L-1}\left(\bW^{L-1} \ldots f^{1}\left(\bW^{1}\bx\right) \right)\right),
\label{eq:chain_nn}
\end{equation}
\end{definition}
Here, $\{\bW_l, l=1,\ldots,L \}$ represents the parameter set, and $f^l(\cdot)$ denotes the transformation function of the $l$-th layer. For simplicity, we have omitted the bias term in the mathematical representation. It is important to note that the expression should be modified if extended to a Transformer model.

To simplify the notation, let us define:
\begin{equation} 
\begin{aligned}
    \by^l &= \bW^{l} \bx^{l-1}, \\
\bx^l &=f^{l}\left(\by^l \right),
\label{eq:dl_definition_with_x_y}
\end{aligned}
\end{equation}
where $\bx^{l-1}$ represents the input features and $\bx^l$ denotes the output activation. $\bW^l$ refers to the parameters in the $l$-th layer, and $f(\cdot)$ represents the transformation function (e.g., ReLU). It is worth noting that this notation can be extended according to Equation~\ref{eq:chain_nn}, yielding $\bx^l =f^{l}\left(\bW^L \bx^{l-1}\right) = f^{l}\left(\bW^{l}  f^{l-1}\left( \bW^{l-1} \bx^{l-2} \right)  \right)$.

In this section, we will define two fundamental principles of optimization in deep learning. Before that, let us further review several definitions and lemmas that will be used in the optimization principles.

Given $\frac{\partial \mathcal{L}}{\partial \bx^l}$, we can calculate $\frac{\partial \mathcal{L}}{\partial \bx^{l-1}}$ and $\frac{\partial \mathcal{L}}{\partial \bW^l}$ as follows:
\begin{equation} 
\begin{aligned}
\frac{\partial \mathcal{L}}{\partial \bx^{l-1}} &= {\bW^l}^{\top} \frac{\partial \bx^l}{\partial \by^l} \frac{\partial \mathcal{L}}{\partial \bx^l}, \\
\frac{\partial \mathcal{L}}{\partial \bW^{l}} &=  \frac{\partial \bx^l}{\partial \by^l} \frac{\partial \mathcal{L}}{\partial \bx^l} {\bx^{l-1}}^{\top}.
\end{aligned}
\label{eq:dl_bp_x_w}
\end{equation}

Let us define $\operatorname{Lip}(f^{l}(\bW^l {\bx}^{l-1}))$ as the Lipschitz constant of the $l$-th layer. We have the following lemma.

\begin{lemma} 
\label{lemma-ldnn}
Given the Lipschitz constant of each transformation function in a network $F$, the following inequality holds:
\begin{equation}
   \operatorname{Lip}(F({\bx}; \{\bW^l, l=1,\ldots,L \})) \leq \prod_{l=1}^{L} \operatorname{Lip}(f^{l}(\bW^l {\bx}^{l-1})). 
\label{eq:model_lipschitz}
\end{equation}
\end{lemma}

From Lemma~\ref{lemma-ldnn}, the Lipschitz constant of a network is upper-bounded by the product of each layer's Lipschitz constant. It should be noted that $\prod_{l=1}^{L} \operatorname{Lip}(f^{l}(\bW^l {\bx}^{l-1}))$ is the upper bound of the Lipschitz constant of the whole network. Considering that the network cannot always reach the upper bound Lipschitz constant in each layer, in practice, the true Lipschitz constant of the whole network will be smaller than the upper bound.

One method to estimate the Lipschitz constant is to simulate the value according to its definition as:
\begin{equation}
    K_s = \operatorname{max}_{\bm{\epsilon}; \bx} \frac{ \|F(\bx+\bm{\epsilon}; \{\bW_l, l=1,\ldots,L \}) - F(\bx; \{\bW_l, l=1,\ldots,L \}) \|} {\|\bx+\bm{\epsilon}-\bx\|}
\end{equation}

In practice, for efficiency, we cannot enumerate all points, so we only sample some points. For instance, we select 100 points $\bx$, and for each point, we randomly select 100 points of different $\bm{\epsilon}$. If we denote the simulated  Lipschitz constant as $K_s$, the true Lipschitz constant as $K_t$, and the theoretical
 upper bound of Lipschitz constant as $K_u$, we have the following relationship among these three values:
\begin{equation}
    K_s \leq K_t \leq K_u.
\end{equation}

$K_u$ can be obtained by theoretical derivation, and it is usually very large. However, obtaining $K_t$ is challenging because it is difficult to enumerate all possible points. We can approximate $K_t$ by simulating as many points as possible, thus we have $K_s \leq K_t$.

\subsubsection{An Example to Understand Lipschitz Constants of Networks}
\label{sec:intro_an_example_for_lips}
\begin{figure}[thb] 
    \centering
    \includegraphics[width=0.45\textwidth]{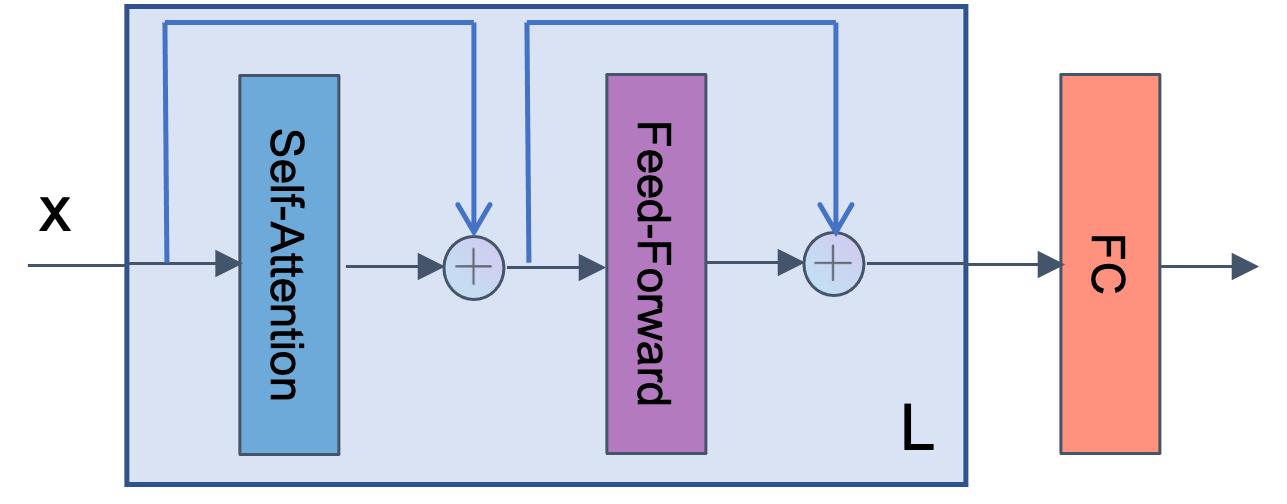}
    \caption{Visualization of a simple Transformer with $L$ layers and a FC layer.} 
    \label{fig:simplenet2}
\end{figure}

To gain a deeper understanding of the computation of the Lipschitz constant, let us consider an example. The network structure is illustrated in Figure~\ref{fig:simplenet2}.

Same as our previous example, $\bx$ represents the input data. The network consists of $L$ layers in the stem, where each layer contains two submodules: self-attention (SA), feed-forward network (FFN),  each module includes a residual shortcut. Finally, we have a classification layer.

Mathematically, we can define the forward process as follows:
\begin{equation}
\begin{aligned}
{\by}^{1} &= \bx + \operatorname{SA}^{1}({\bx}; {\bm{\sW}}^{1}) \\
{\bx}^{1} &= \by^{1} + \operatorname{FFN}^{1}({\by^1}; {\bm{\sV}}^{1}) \\
\vdots \\
{\by}^{l+1} &= \bx^{l} + \operatorname{SA}^{l+1}({\bx^l}; {\bm{\sW}}^{l+1}) \\
{\bx}^{l+1} &= \by^{l+1} + \operatorname{FFN}^{l+1}({\by^{l+1}}; {\bm{\sV}}^{l+1}) \\
\vdots \\
{\by}^{L} &= \bx^{L-1} + \operatorname{SA}^{L}({\bx^{L-1}}; {\bm{\sW}}^{L}) \\
{\bx}^{L} &= \by^{L} + \operatorname{FFN}^{L}({\by^{L}}; {\bm{\sV}}^{L}) \\
\bo &= \bO \bx^{L} \\
\end{aligned}
\end{equation} 
where  ${\bm{\sW}}^{l}$ and ${\bm{\sV}}^{l}$ are learnable parameters of the self-attention and the feed-forward network submodules in the $l$-th layer individually.

We can compute the Lipschitz constant of the whole network as:
\begin{equation}
\begin{aligned}
      \operatorname{Lip}\left(F({\bx}; ({\bm{\sW}}^{l}, {\bm{\sV}}^{l}), l=1,\ldots,L)\right) & \leq \left( \prod_{l=1}^{L} \operatorname{Lip}\left(f^{l}({\bx}^{l-1}; ({\bm{\sW}}^{l}, {\bm{\sV}}^{l})\right) \right)  \operatorname{Lip}(\bO \bx^{L})  \\
      & = \left( \prod_{l=1}^{L} \left(1+\operatorname{Lip}\left(  \operatorname{SA}^{l}({\bx}^{l-1}; {\bm{\sW}}^{l})\right)\right)  \left(1+\operatorname{Lip}\left(  \operatorname{FFN}^{l}({\by}^{l}; {\bm{\sV}}^{l})\right)\right)  \right)  \operatorname{Lip}(\bO \bx^{L}) \\
      & = \left( \prod_{l=1}^{L} \left(1+\operatorname{Lip}\left(  \operatorname{SA}^{l}({\bx}^{l-1}; {\bm{\sW}}^{l})\right)\right)  \left(1+\operatorname{Lip}\left(  \operatorname{FFN}^{l}({\by}^{l}; {\bm{\sV}}^{l})\right) \right)  \right) \cdot \sigma_{max}(\bO), \\
\end{aligned}
\label{eq:model_lipschitz_example}
\end{equation}
where $\sigma_{max}(\bO)$ is the maximum absolute eigenvalue of the matrix $\bO$.

From this example, we can see that the Lipschitz constant of the network is highly related to the submodules of $\operatorname{Lip}\left(  \operatorname{SA}^{l}({\by}^{l}, {\bm{\sW}}^{l})\right)$ and $\operatorname{Lip}\left(  \operatorname{FFN}^{l}({\by}^{l}, {\bm{\sV}}^{l})\right)$. If these submodules are unstable (their Lipschitz constant are very large or unbound.), then the whole network will be unstable.  \textit{The Lipschitz constant of each module affects the training stability of the network, and it can be calculated according to its Jacobian matrix.} Therefore, we should analyze the Jacobian matrix of each module in detail. This observation further explains our understanding overview in Figure~\ref{fig:optimization_overview}.

\subsection{Principles of Optimization}
In this subsection, we will clarify two simple and fundamental optimization principles.

\begin{principle}[Forward Principle of Optimization]
\label{lemma-forward_principle}
To ensure the stability of model training, it is necessary to ensure that the values of activations across all layers in the forward process satisfy the following condition:
\begin{equation}
\bx^l, \by^l < \mathcal{R}, \ \text{for\ } l \in [1, L],
\end{equation}
where $\mathcal{R}$ represents the maximum value range of the float precision used, and $\bx^l, \by^l$ are defined as in Equation~\ref{eq:dl_definition_with_x_y}.
\end{principle}

When using single precision (Float32) training, $\mathcal{R} = 3.4 \times 10^{38}$. In MPT~\citep{mixedprecision_micikevicius2017mixed}, the FP16 range is $[-65,504, 65,504]$, which means $\mathcal{R} = 65,504$. If the value of an activation exceeds $\mathcal{R}$, it will trigger an overflow and result in an Infinity value. Performing an operation on two Infinity values will trigger a NAN (Not a Number) warning or error. As for the underflow problem, the model can still be trained stably, although the precision may be slightly affected due to the decreased precision from underflow.

Normalization techniques such as BN and LN are highly effective in ensuring the validity of the forward principle of optimization. Without normalization, in a network where each layer is an expansion mapping, the activation values may overflow after several layers. However, when a normalization operation is applied after each layer, the feature's norm is consistently normalized to a relatively small value, preventing any overflow issues during the forward process. While normalization plays a crucial role in upholding the forward principle, it should be acknowledged that for certain abnormal inputs, normalization might violate the backward optimization principle.

\begin{principle}[Backward Principle of Optimization]
\label{lemma-backward_principle}
To ensure the convergence of model training, we need to ensure that the gradients of the activations across all layers in the backward process satisfy the condition:
\begin{equation}
\nabla_{\bx^l} \mathcal{L}, \nabla_{\by^l} \mathcal{L} < \mathcal{R}, \ \ \text{for\ \ } l \in [1, L].
\end{equation}
\end{principle}

Based on the backward computation shown in Equation~\ref{eq:dl_bp_x_w}, we can observe that Princple~\ref{lemma-backward_principle} typically implies:
\begin{equation}
\nabla_{\bW^l} \mathcal{L}  < \mathcal{R}, \text{for\ \ } l \in [1, L].
\end{equation}

\textit{Principles 1 and 2 are two fundamental principles for a stable network training.} Forward principle of optimization is usually easy to promise via some normalization skills, but backward principle is harder to satisfy considering that the training process of deep learning is a dynamic process. In each training step, the Jacobian matrices and the Lipschitz constants are evolving.

As depicted in Figure~\ref{fig:optimization_overview}, optimization in deep learning mainly faces two main challenges: gradient vanishing and gradient exploding. Gradient vanishing does not cause the network training to collapse but results in a weak representation. On the other hand, gradient exploding directly leads to failed model training.

Back-propagation involves the chain composition of the Jacobian matrix of each layer. The Lipschitz constant of each layer can be calculated using the Jacobian matrix. Therefore, considering the Jacobian matrix of each transformation function in the network is an effective approach to understanding deep learning optimization.

Table~\ref{tab:bigtable} presents the forward definitions of some common layers, their Jacobian matrices or gradients, and their theoretical Lipschitz constants. A large Lipschitz constant indicates that the layer may often result in an expansion mapping for the gradients in the backward process. Similarly, a small Lipschitz constant implies that the gradient norm may not expand significantly. From the table, we observe that if Sigmoid is placed in the stem, it can lead to gradient vanishing. ReLU, GeLU, and Swish all propagate the gradients effectively, with ReLU being non-smooth while GeLU and Swish being smooth functions. The residual shortcut is an effective way to preserve the gradient flow in the stem, even if the branch experiences gradient vanishing.
Linear and Convolution are two homogeneous operators, and they have similar forms of Lipschitz constants. For most normalization methods, the values of their Jacobian matrices can be very large when abnormal data points are inputted. This indicates a large Lipschitz constant for these layers. Three attention mechanisms are shown in the table, where dot-product attention is not Lipschitz continuous despite its powerful representation ability. $L_2$ distance attention is Lipschitz continuous when $\bW^Q = \bW^K$. Scale cosine similarity attention is Lipschitz continuous without requiring specific conditions on the weight matrices.

\begin{table}   
\centering
\small
\tiny
\scriptsize
{\renewcommand{\arraystretch}{1.9}
\begin{tabular}{llll} 
\hline
\toprule
Layer Type & Definition & Gradient or Jacobian & Lipschitz Constant    \\ \hline
\midrule
\makecell[l]{\textbf{Linear}} & $\by = \bW \bx$ & $\frac{\partial \by}{\partial \bx} = \bW^{\top}$ &  $\sigma_{\max}(\bW)$\\  \hline 
\makecell[l]{\textbf{Convolution} \\ \cite{lenet_lecun1998gradient}} & $\by^O = \bW_{K,K,C}^{O} \bx^{K,K,C}$  & $\frac{\partial \by^{O}}{\partial \bx^{D}} = {\left(\bW_{D}^{O}\right)}^{\top} = {\bW_{O}^{D} }  $  &   $\sigma_{\max}(\bW_{O}^{D})$ \\   \hline 
\midrule 
\makecell[l]{\textbf{Sigmoid}} & $y_i = \frac{1}{1+\operatorname{exp}(-x_i)}$  & $\frac{\partial y_i}{\partial x_i} = \sigma(x_i)(1-\sigma(x_i))$  & $\frac{1}{4}$\\ \hline

\makecell[l]{\textbf{Softmax}} & $y_i = \frac{\operatorname{exp}(x_i)}{\sum_{i=0}^{d-1} \operatorname{exp} (x_i)}$ & $\frac{\partial y_i}{\partial x_i} = \left(y_i\right)(1{\left(i==j\right)} - y_i)$ &  \makecell[l]{$\leq 1$ \\ 
\cite{softmax_lipschitz_gao2017properties}}\\  \hline 

\makecell[l]{\textbf{ReLU} \\ 
\citeauthor{relu_dahl2013improving}} & $y_i = \operatorname{max}(0, x_i)$ & $\frac{\partial y_i}{\partial x_i} =  1{\left(x_i>0\right)}$ & 1.0 \\  \hline 

\makecell[l]{\textbf{GeLU} \\ \citeauthor{gelu_hendrycks2016gaussian}} & $y_i = x_i\operatorname{P}(x < x_i) \approx x_i \sigma(1.702 x_i)$
& \makecell[l]{$\frac{\partial y_i}{\partial x_i} \approx \sigma(1.702 x_i) + 1.702 x_i \cdot$ \\ $\sigma(1.702 x_i) \left(1- \sigma(1.702 x_i) \right)$ }  &  $\approx 1.1$ \\ \hline

\makecell[l]{\textbf{Swish} \\ \citeauthor{swish_ramachandran2017searching} }& $y_i = x_i \sigma(x_i) $ & \makecell[l]{$\frac{\partial y_i}{\partial x_i} = \sigma(x_i) + x_i \cdot$ \\ $\sigma(x_i) \left(1- \sigma(x_i) \right)$ }   & $\approx 1.1$ \\ \hline

\midrule 

\makecell[l]{\tbf{DP Attention} \\ \citeauthor{transformer_vaswani2017attention}} 
& 
\makecell[l]{$\bY =	\boldsymbol{W}^{V} \bX \cdot \mathcal{S} \left(\frac{ {(\boldsymbol{W}^{Q} \bX)}^{\top}  \left( \boldsymbol{W}^{K} \boldsymbol{X} \right)}{\sqrt{D/H}}\right)$}
& 
{\makecell[l]{See Equation 12 in~\cite{l2distance_attention_kim2021lipschitz}}}
& $\infty$ \\ \hline

{\makecell[l]{\tbf{$L_2$ Attention} \\ \citeauthor{l2distance_attention_kim2021lipschitz}}} & 
{\makecell[l]{ $\bY = \bW^V \bX \cdot$ \\
$\mathcal{S} \left(-\frac{\bX^{\top} {\left(\bW^{Q}-\bW^{K}\right)}^{\top} {\left(\bW^{Q}-\bW^{K}\right)}\bX} {\sqrt{D / H}}\right) $}}
&
{\makecell[l]{See Equation 19 and 20 \\ in~\cite{l2distance_attention_kim2021lipschitz}}}
& \makecell[l]{ $
\frac{\sqrt{N}}{\sqrt{D / H}}\left(4 \phi^{-1}(N-1)+1\right)$ \\  $\left(\sqrt{\left\|\bW^{Q}\right\|_{2}^{2}\left\|\bW^{V}\right\|_{2}^{2}}\right)\left\|\bW^{O}\right\|_{2}$ \\ 
$\text{when} \ \  \bW^Q = \bW^K $
} \\ \hline
\makecell[l]{\tbf{
SCS Attention}  \\ \citeauthor{lipsformer_qi2023lipsformer}} & 
\makecell[l]{ $ \bY =  \nu \bV \boldsymbol{P}, $ \\ $
				\text{where } 
				\boldsymbol{P} = \operatorname{softmax}\left(\tau \bQ^{\top} \bK \right)$}  
& 
{\makecell[l]{See Equation 13 and 14 \\ in~\cite{lipsformer_qi2023lipsformer}}}
& \makecell[l]{ $2 N (N-1)  \nu  \tau  \epsilon^{-\frac{1}{2}} {\|{{\bW^K}}\|}_{2} + $ \\ $ 2 (N-1) \nu \tau \epsilon^{-\frac{1}{2}} {\|{{\bW^Q}}\|}_{2}  + $ \\ $  2 N  \nu \epsilon^{-\frac{1}{2}} {\|{{\bW^V}^{\top}}\|}_{2}$} \\ \hline 
\midrule 
\makecell[l]{\tbf{
LayerNorm} \\ \citeauthor{layernorm_ba2016layer}} & \makecell[l]{$\by =  \left(\boldsymbol{I}-\frac{1}{D}  \boldsymbol{1}  \boldsymbol{1}^{\top}\right) \bx$ \\
    $\bz =  \sqrt{D} \frac{\by}{\sqrt{\| \by \|_2^2 +\epsilon }}$  \\
   $\operatorname{LN}(\bx) =  \boldsymbol{\gamma} \odot \bz + \boldsymbol{\beta}
    $}  
&\makecell[l]{$ \frac{\partial \operatorname{LN} (\bx)}{\partial \bx} =  \frac{\sqrt{D}}{\sqrt{\| \by \|_2^2 +\epsilon }}  \left(\boldsymbol{I}-\frac{1}{d} \boldsymbol{1} \boldsymbol{1}^{\top}\right) $ \\ $  \left( \bI - \frac{\by \by^{\top}}{\| \by \|_2^2 +\epsilon}\right)  \operatorname{diag}\left(\bm{\gamma}\right)$ }
& $\frac{\max _{D}\left|\gamma_{D}\right|}{\epsilon^{\frac{1}{2}} } \sqrt{D}
 $ \\ \hline
\makecell[l]{\tbf{
BatchNorm} \\ \citeauthor{batchnorm_ioffe2015batch}} & 
\makecell[l]{$\bm{\mu}  = \frac{1}{N} \sum_{i=1}^{N} \bX_{:,i}$ \\
          $\bm{\sigma}^{2} =  \frac{1}{N} \sum_{i=1}^{N}\left(\bX_{:,i}-\bm{\mu}\right) \odot \left(\bX_{:,i}-\bm{\mu}\right) $\\
          $\widehat{\bX_{:,i}} = \left(\bX_{:,i}-\bm{\mu}\right) \oslash {\sqrt{\bm{\sigma}^{2}+\epsilon}} $\\
          $\mathrm{BN}\left(\bX_{:,i} \right)   = \bm{\gamma} \odot \widehat{\bX_{:,i}} + \bm{\beta}$}
& \makecell[l]{see Equation \ref{eq:bn_deriv} and \ref{eq:bn_deriv2}} & $\approx$ \  $\max _{D} \frac{ \left|\gamma_{D}\right| }{\sqrt{\sigma_{D}^2 + \epsilon}} $ \\  \hline

\makecell[l]{\tbf{
WeightNorm} \\ \citeauthor{weight_normalization_salimans2016weight}} & \makecell[l]{$\bW(i, :)  = \gamma_i \frac{\bv_i}{{\sqrt{{\|\bv_i\|}_2^2 + \epsilon}}} $\\
        $\operatorname{WN}(\bx)  = \bW \bx $} &  \makecell[l]{$\frac{\partial \operatorname{WN}(\bx)}{\partial \bx} = \bW,$ \\
        \text{where}, $\bW(i, :) = \gamma_i \frac{\bv_i}{\sqrt{{\|\bv_i\|}_2^2 + \epsilon}}$} & \makecell[l]{$\sigma_{\max}(\bW) \leq \sqrt{\sum_{i=1}^{O} \gamma_i^2}$}\\ \hline

\makecell[l]{\tbf{
RMSNorm} \\ \citeauthor{rmsnorm_zhang2019root}} 
& \makecell[l]{$\operatorname{RMSN}(\bx)   = \bm{\gamma} \odot \frac{\sqrt{D} \bx}{\sqrt{{\|\bx\|}_2^2 + \epsilon}} + \bm{\beta}$} 
& \makecell[l]{$\frac{\partial \operatorname{RMSN}(\bx)}{\partial \bx} = \frac{\sqrt{D}}{\sqrt{\| \bx \|_2^2 +\epsilon }} $ \\ $ \left( \bI - \frac{\bx \bx^{\top}}{\| \bx \|_2^2 +\epsilon}\right) \operatorname{diag}\left(\bm{\gamma}\right)$}& $\frac{ 
\max _{D}\left|\gamma_{D}\right| }{\epsilon^{\frac{1}{2}}} \sqrt{D} $ \\ \hline

\makecell[l]{\tbf{
CenterNorm} \\  \citeauthor{lipsformer_qi2023lipsformer}} &  \makecell[l]{$\by = \left(\boldsymbol{I}-\frac{1}{D}  \boldsymbol{1} \boldsymbol{1}^{\top}\right) \bx$ \\
     $\operatorname{CN}(\bx) = \frac{D}{D-1} \boldsymbol{\gamma} \odot \by + \boldsymbol{\beta} $}
& \makecell[l]{$\frac{\partial \operatorname{CN}(\bx)}{\partial \bx} =  \frac{D}{D-1}  \left(\boldsymbol{I}-\frac{1}{D}  \boldsymbol{1} \boldsymbol{1}^{\top}\right)$ \\ $\operatorname{diag}\left(\bm{\gamma}\right)  $} & $\frac{D}{D-1}\max _{D}\left|\gamma_{D}\right| $ \\   \hline 
\midrule

\makecell[l]{\tbf{
Residual} \\ \citeauthor{resnet_he2016deep} } & $\by = \bx + f(\bx, \bW)$ & $\frac{\partial \by}{\partial \bx} = \bI + \frac{\partial f(\bx, \bW)}{\partial \bx}$ & $1 + \operatorname{Lip}(f(\bx, \bW))$ \\  \hline

\makecell[l]{\tbf{Weighted Residual} \\ \tbf{Block} \citeauthor{lipsformer_qi2023lipsformer} } & $f(\bx, \bW) = \bx + \bm{\nu}_1 \odot f(\bx, \bW)$ & $\frac{\partial \by}{\partial \bx} = \bI +  \frac{\partial f(\bx, \bW)}{\partial \bx} \operatorname{diag}(\bm{\nu})$ & $1 + \operatorname{max}_D (\left|\nu_D\right|) \operatorname{Lip}(f(\bx, \bW))$ \\  \hline

\midrule


\makecell[l]{\tbf{
MaxPooling} \\ \citeauthor{maxpooling_ranzato2007sparse}} & $y = \operatorname{max}(\bx)$ & $\frac{\partial y}{\partial \bx} = \bm{1}{(\bx == y)}$ & 1 \\  \hline 
\makecell[l]{\tbf{
AveragePooling} \\ \citeauthor{averagepooling_lecun1989handwritten}} & $y = \frac{1}{D} \operatorname{sum}(\bx)$& $\frac{\partial y}{\partial \bx} = \frac{1}{D} \bm{1} $& $\frac{1}{{D}}$ \\

\bottomrule
 
\end{tabular}}
\caption{The definitions, Jocabian matrices and Lipschitz constants of some widely used modules in deep learning.
All Lipschitz constants are calculated under the $L_2$ norm. Here, $D$ represents the dimension of the input feature.
In the definition of Convolution, we use Einstein notation in its Jacobian and Lipschitz constant calculations, where $D=K\times K \times C$. 
$\sigma(\cdot)$ denotes the sigmoid function, and $\mathcal{S}(\cdot)$ represents the softmax function. In the definition of Lipschitz constant of the $L_2$ attention, $\phi(x_i) = x_i \operatorname{exp}(x_i + 1)$. Additional details about DP attention and $L_2$ attention can be found in~\cite{l2distance_attention_kim2021lipschitz}. For all attention mechanisms, we only consider their single-head attention in this context.
Symbolic mathematical tools such as~\cite{matrixcalculus_LaueMG2020} can be employed to obtain simple Jacobian. 
Due to space constraints, we do not present the calculation process for deriving the Jacobian matrices and Lipschitz constants. However, Appendix~\ref{sec:appendix_derivations} provides some details on calculating the Lipschitz constants for certain modules. 
For the DP attention and the $L_2$ attention, please refer to~\cite{l2distance_attention_kim2021lipschitz} for further information. Detailed proofs for SCS attention can be found in~\cite{lipsformer_qi2023lipsformer}. In this paper, we directly use their reported results.
We have made every effort to ensure the accuracy of the derivations. If you find any results that are not rigorous or incorrect, please feel free to provide a correction to us.}
\label{tab:bigtable}
\end{table}

\newpage

\newcolumntype{R}{>{$}r<{$}}
\newcolumntype{L}{>{$}l<{$}}
\newcolumntype{M}{R@{${}={}$}L}

\section{Implicit Optimization in Deep Learning}
\label{sec:implicit_optimization}
\subsection{Normalization}
Normalization is an effective re-parameterization technique~\footnote{\url{https://sassafras13.github.io/ReparamTrick/}} that can significantly improve the training process and performance of deep neural networks. By re-scaling and centering the input features, normalization helps mitigate issues related to the scale and distribution of the data. In this section, we will discuss different types of normalization techniques and their benefits in deep learning.

\begin{theo}[Normalization]{thm:normalization}
\label{remark:normalization}
\begin{enumerate}[leftmargin=*]
\item Normalization is an effective approach to mitigating gradient vanishing. \\
\item Normalization smoothens the landscape. \\
\item Generally, BN is more stable than LN, but LN has a broader range of applications than BN.
\end{enumerate}
\end{theo}

In Section~\ref{sec:intro_normalization}, we have briefly reviewed LayerNorm and BatchNorm. In Table~\ref{tab:bigtable}, we list the definitions of some other normalizations along with their Jacobian or gradients, and Lipschitz constants. Due to space and time limitations, we could not include many other normalization methods such as Group Normalization~\citep{groupnorm_wu2018group} and Instance Normalization~\citep{instancenorm_ulyanov2016instance}, and others.

From the perspective of coordinate centralization, we consider the following ranking:
\begin{equation}
\label{eq:normalization_centralization}
    \text{LayerNorm} > \text{BatchNorm} > \text{CenterNorm} > \text{RMSNorm} > \text{WeightNorm}.
\end{equation}

LayerNorm centralizes and re-scales the activations at each spatial or sequence point, providing a more fine-grained normalization. On the other hand, BatchNorm centralizes and re-scales the activations by computing a moving average mean and standard deviation. CenterNorm only centralizes the features without re-scaling them, while RMSNorm scales the features based on their $L_2$ norm. WeightNorm, on the other hand, normalizes the weights instead of the activations.

From the perspective of Lipschitz stability, we consider the following ranking:
\begin{equation}
\label{eq:normalization_stability}
    \text{CenterNorm} > \text{WeightNorm} > \text{BatchNorm} > \text{RMSNorm} \approx  \text{LayerNorm}.
\end{equation}

Their corresponding Lipschitz constants, according to Table~\ref{tab:bigtable}, are:

\begin{equation}
\label{eq:normalization_lipschitz_constant}
\frac{D}{D-1}\max_{D}\left|\gamma_{D}\right| <   \sqrt{\sum_{i=1}^{O} \gamma_i^2} <
\max_{D} \frac{\left|\gamma_{D}\right| }{\sqrt{\sigma_{D}^2 + \epsilon}} <
\frac{\max _{D}\left|\gamma_{D}\right| }{\epsilon^{\frac{1}{2}}}  \sqrt{D} \approx
\frac{\max _{D}\left|\gamma_{D}\right|}{\epsilon^{\frac{1}{2}}} \sqrt{D}
\end{equation}

From their Jacobian matrix, we can see that when the input features are equal across all dimensions, LayerNorm will have a very large Lipschitz constant when the values in all dimensions are equivalent. RMSNorm will have a large Lipschitz constant when $\bx \approx 0$. Due to the mean and standard value being computed from the entire batch via a moving average, there is a lower probability of centering the features to 0 across all dimensions. CenterNorm and WeightNorm have Lipschitz constants that are close to the norm of $\bm{\gamma}$.

We make several remarks about normalization in Remark~\ref{remark:normalization}. As we have described, the forward process of a typical neural network propagates computation as $\by^{l+1} = {\bW^{l+1}} \bx^l$, where $\bx^l$ and $\bW^{l+1}$ are the input and weight matrix of Layer $l+1$. To back-propagate the network loss $\mathcal{L}$, we have:
\[
\frac{\partial \mathcal{L}}{\partial \bx^{l}} = {\bW^{l+1}}^{\top} \frac{\partial \mathcal{L}}{\partial \by^{l+1}}, \ \ 
\frac{\partial \mathcal{L}}{\partial \bW^{l+1}} = {(\frac{\partial \mathcal{L}}{\partial \by^{l+1}})} {\bx^{l}}^{\top}.
\]

When $\bx^{l}$ is normalized, the gradient of $\frac{\partial \mathcal{L}}{\partial \bW^{l+1}}$ in all channels will be distributed more evenly across all channels. This alleviates the issue of gradient vanishing. As also pointed out in~\cite{how_bn_santurkar2018does}, BN smooths the entire landscape of the network. We will discuss this property further in the experimental section.

As shown in~\cite{postandpre_ln_xiong2020layer}, the main difference between pre-LN and post-LN is that post-LN is used in the stem, while pre-LN is used in the branch. We have discussed earlier that LayerNorm is important in smoothing the landscape, but we also find out that it has a high probability of creating abnormal gradients for some abnormal input points, which leads to unstable training. Since the abnormal gradients occur in the stem, they affect layers from the current layer to the input, and thus lead to unstable training.

\subsection{Self-attention}
\label{sec:opt_selfattention}
In Section~\ref{sec:intro_selfattention}, we reviewed the basic Dot-product (DP) attention. Here, we will further review some improvements over DP attention.

In~\cite{l2distance_attention_kim2021lipschitz}, Kim et al. prove that the standard dot-product attention is \emph{not} Lipschitz continuous and introduced an alternative L2 attention which is Lipschitz continuous. Their $L_2$ distance attention (referred to as "L2 attention" below) can be defined as:
\begin{equation}
\label{eq:l2_attention}
\operatorname{Attn\_L_2}(\boldsymbol{X}; \boldsymbol{W}^{Q}, \boldsymbol{W}^{K}, \boldsymbol{W}^{V}) = \bW^{V} \bX \cdot \mathcal{S} {\left(-\frac{{\left(\bW^{Q} \bX -\bW^{K} \bX \right)}^{\top} {\left(\bW^{Q}\bX -\bW^{K} \bX \right)}}{\sqrt{D / H}}\right)},
\end{equation}

where $\mathcal{S}(\cdot)$ denotes the softmax operation, $D$ is the hidden dimension and $H$ is the number of heads. 

Qi et al.~\citep{lipsformer_qi2023lipsformer} introduce Scaled Cosine Similarity Attention (referred to as "SCS attention" or "SCSA"), which is defined as:
\begin{equation}
\label{eq:scsa_attention}
\begin{gathered}
\operatorname{Attn\_SCS}(\boldsymbol{X}; \boldsymbol{W}^{Q}, \boldsymbol{W}^{K}, \boldsymbol{W}^{V}, \nu, \tau) = \nu \bV \boldsymbol{P}, 
\text{where } 
\boldsymbol{P} = \operatorname{softmax}\left(\tau \bQ^{\top} \bK \right).
\end{gathered}
\end{equation}
where,
\begin{equation*}
\bQ =\left[\bq_{1}, \cdots,  \bq_{N}\right], \ \ \ 
\bK =\left[\bk_{1}, \cdots,  \bk_{N}\right], \ \ \ 
\bV =\left[\bv_{1}, \cdots,  \bv_{N}\right].
\end{equation*}
where $\nu$ and $\tau$ are predefined or learnable scalars. $\bQ, \bK, \bV$ are $\ell^2$ column-normalized: \\ $\bq_i, \bk_i, \bv_i = \frac{{{\bW^{Q} \bx_i}}}{\sqrt{{\|{{\bW^{Q} \bx_i}} \|}^2 + \epsilon}}, \frac{{{\bW^{K} \bx_i}}}{\sqrt{{\|{{\bW^{K} \bx_i}} \|}^2 + \epsilon}}, \frac{{{\bW^{V} \bx_i}}}{\sqrt{{\|{{\bW^{V} \bx_i}} \|}^2 + \epsilon}}$.
Here, $\epsilon$ is a smoothing factor that guarantees the validity of cosine similarity computation even when ${\|{\bW^Q  \bx_i} \|} = 0$. 
For arbitrary pairs of rows of $\boldsymbol{Q}$ and $\boldsymbol{K}$ denoted as $\bq_i$ and $\bk_j$, the cosine similarity on their $\ell^2$-normalized vectors is proportional to their dot product. The upper bound of SCS Attention's Lipschitz constant with respect to $\|\cdot\|_2$ is shown in Table~\ref{tab:bigtable}.

For easy understanding, in the following, we abbreviate SCS attention as SCSA, $L_2\ \text{attention}$ as $L_2$A, and \text{dot-product attention} as DPA.

\begin{theo}[Self-Attention]{thm:attention}
\label{remark:attention}
\begin{enumerate}[leftmargin=*]
    \item Self-attention is a high-order nonlinear operator with strong representation ability. \\
    \item DP attention is not Lipschitz continuous, which can result in training instability if warmup, weight decay, and learning rate are not properly set. \\
    \item DP attention and LN are two modules that often trigger training instability due to their unbounded or large Lipschitz constants. \\
    \item Considering the Lipschitz constants of different attention mechanisms, SCS attention is a more stable version of attention.
\end{enumerate}
\end{theo}

According to the Lipschitz stability of all attention mechanisms, we reckon that:
\begin{equation*}
    \text{SCSA} > L_2\text{A} > \text{DPA}.
\end{equation*}

In Remark~\ref{remark:attention}, we have provided several remarks about self-attention. Self-attention is a higher-order nonlinear operator that differs from linear layers and convolutions. From the Jacobian and gradient derivations in Table~\ref{tab:bigtable}, we can see that self-attention and LN are two modules that can result in large gradients, leading to unstable training.

In Table~\ref{tab:bigtable}, we have listed the Lipschitz constants for DPA, $L_2$A, and SCSA. More detailed derivations can be found in~\cite{l2distance_attention_kim2021lipschitz, lipsformer_qi2023lipsformer}. $L_2$ attention is Lipschitz continuous under the assumption that $\bW^{Q} = \bW^{K}$.

\begin{theo}[Residual Shortcut]{thm:residualshortcut_2}
\label{remark:residual}
\begin{enumerate}[leftmargin=*]
    \item Residual shortcut is an effective approach to mitigating the gradient vanishing problem. \\ 
    \item Residual shortcut helps smooth the landscape of a network. \\
    \item However, residual shortcut may also increase the Lipschitz constant of the network, which can potentially exacerbate the issue of gradient exploding.
\end{enumerate}
\end{theo}

\subsection{Residual Shortcut}
Residual shortcuts, introduced in ResNet architectures~\citep{resnet_he2016deep,identity_mapping_he2016identity}, are an effective way to alleviating the vanishing gradient problem that often affects deep neural network training. By incorporating skip connections, residual shortcuts enable gradients to flow more easily through the network, resulting in improved training and performance. Since the introduction of residual shortcuts, several enhancements have been made in this area.

ReZero, introduced by \citeauthor{rezero_bachlechner2021rezero}, is one such enhancement applied to residual networks. ReZero is defined as:
\begin{equation}
    \bx^{l+1} = \bx^{l} + \bm{\nu}_1 \odot f(\bx^{l}; \bW),
\label{eq-initialization_define}
\end{equation}

where $ \bm{\nu}_1$ is a learned parameter initially set to $\bm{0}$. ReZero serves as an initialization method, ensuring that the module after ReZero has a Lipschitz constant of 1.0 under the initial condition. This allows network training even without warmup.

In contrast, \citet{lipsformer_qi2023lipsformer} introduce a Weighted Residual Shortcut (WRS) block instead of initializing $\bm{\nu}$ to 0. WRS initializes $\bm{\nu}$ to $\frac{1}{L}$, where $L$ represents the number of layers. In their study, after WRS initialization, the Lipschitz constant of the network becomes a value related to Euler's number $e$.

A potential issue is that the value of $\bm{\nu}$ may increase rapidly, leading to an increased Lipschitz constant for the network. A simple solution is to constrain the values such that $\operatorname{abs}(\bm{\nu}) < \omega$, where $\omega$ can be set, for example, to 2.0. This helps maintain the Lipschitz stability of the network.

We have made several remarks about residual shortcut in Remark~\ref{remark:residual}. $\bx^{l+1} = \bx^{l} + f(\bx^{l}; \bW^{l+1})$, since the Jacobian matrix $\frac{\partial \bx^{l+1}}{\partial \bx^{l}} = \bI + \frac{\partial f(\bx^l; \bW^l)}{\partial \bx^l}$, even when $\frac{\partial f(\bx^l; \bW^l)}{\partial \bx^l} \approx \bm{0}$, the gradient can still be propagated to lower layer because $\frac{\partial \mathcal{L}}{\partial \bx^{l}} = \frac{\partial \mathcal{L}}{\partial \bx^{l+1}}$ when $\frac{\partial f(\bx^{l}; \bW^{l+1})}{\partial \bx^{l}} = \bm{0}$.

\begin{theo}[Activation]{remark:activation}
\label{remark:activation}
\begin{enumerate}[leftmargin=*]
\item Activation functions introduce non-linearity into the network. \\
\item The sigmoid function is prone to the problem of gradient vanishing, while ReLU function disables half of the gradient back-propagation. On the other hand, GELU and Swish functions do not suffer from these issues. \\
\item ReLU is a non-smooth function. From the perspective of classical numerical optimization, non-smooth functions tend to have slower convergence rates during training and may exhibit generalization problems.
\end{enumerate}
\end{theo}

\subsection{Activation}
\label{sec:activation_improvement}
Activation functions play a crucial role in deep neural networks by introducing non-linearity, allowing the network to learn complex, non-linear relationships between input and output. Without activation functions, neural networks (classical multi-layer perceptron (MLP) and convolutional neural network (CNN) without normalization) would be limited to learning only linear transformations, greatly reducing their capacity to model real-world problems. In this section, we discuss the role of activation functions in deep learning and their impact on network performance.

In Table~\ref{tab:bigtable}, we provide the definitions of several activation functions along with their gradients and Lipschitz constants. All the mentioned activation methods do not have Lipschitz stability issues. However, Sigmoid is prone to gradient saturation, which can hinder the flow of gradients. As a result, they are not suitable for the stem of the network but can be used in the branch part. We have provided further remarks in Remark~\ref{remark:activation}.

In recent large language models (LLM)~\citep{palm_chowdhery2022palm, llama_touvron2023llama}, Gated Linear Units (GLU)~\citep{glu_shazeer2020glu} have been utilized. GLU naturally induces more non-linearity into the network.

We have built several remarks about activations in Remark~\ref{remark:activation}.

\subsection{Initialization}
\label{sec:opt_initialization}
Weight initialization plays a critical role in the training process of deep neural networks. Proper initialization can lead to faster convergence and improved model performance.

\begin{table}[h]
\centering
\small 
\begin{tabular}{lc}
\hline
Method Name &  Method      \\ \hline
\toprule
\makecell{\tbf{Xavier Initialization}\\ \cite{xavier_init_glorot2010understanding}} & \makecell{$W_{i, j} \sim \operatorname{U}\left(-\sqrt{\frac{6}{n_{in}+n_{out}}}, \sqrt{\frac{6}{n_{in}+n_{out}}}\right)$ \\ or,
$W_{i, j} \sim \operatorname{N}\left(0, \frac{2}{n_{in}+n_{out}}\right)$
}
\\ \hline
\makecell{\tbf{Kaiming Initialization} \\ \cite{prelu_he2015delving}} &  \makecell{$W_{i, j} \sim \operatorname{N}\left(0, \frac{2}{(1+a^{2}) \times {n_{in}}}\right)$}  \\ \hline
\makecell{\tbf{Orthogonal Initialization} \\ \cite{orthogonal_init_saxe2013exact}} & \makecell{Initialize $\bW_1$ with Xavier initialization, \\
$\bU, \bS, \bV = \operatorname{SVD}(\bW_1)$, \\ 
$\bI = \operatorname{eye}(\bW_1)$, \\
$\bW = \bU\bI\bV^{\top}$ } \\ \hline
\makecell{\tbf{Spectral Initialization} \\  \cite{lipsformer_qi2023lipsformer}} & \makecell{Initialize $\bW_1$ with Xavier initialization, \\
$\bU, \bS, \bV = \operatorname{SVD}(\bW_1)$, \\ 
$\bW = \frac{\bW_1}{S[0]}$} \\ \hline 
\makecell{\tbf{Depth-aware Initialization} \\ \cite{depth_scale_zhang2019improving}} & \makecell{Initialize $\bW_1$ with Xavier initialization, \\ $\bW = f(\bW_1, L)$, e.g., $\bW = \frac{\bW_1}{\sqrt{L}}$} \\ 
\bottomrule
\end{tabular}
\caption{Initialization methods. In Kaiming initialization, $a$ is the slope of the non-linearity function. $\operatorname{eye}(\bW_1)$ is to create an indentity matrix with the same shape of $\bW_1$.}
\label{tab:init_examples}
\end{table}

\begin{theo}[Initialization]{remark_init}
\begin{enumerate}[leftmargin=*]
    \item For large models, the number of layers $L$ should be taken into consideration during initialization because traditional Xavier initialization does not consider $L$, leading to a very large Lipschitz constant. A large Lipschitz constant can trigger training instability. \\
    \item Generally, deeper networks should use a smaller initialization variance. \\
    \item Many previous works, including Admin~\citep{admin_liu2020understanding}, Fixup~\citep{fixup_zhang2019fixup}, DS-Init~\citep{depth_scale_zhang2019improving}, Deepnet~\citep{deepnet_wang2022deepnet}, ReZero~\citep{rezero_bachlechner2021rezero}, and more, focus on constraining the Lipschitz constant of the network in the initial stage, although they may not explicitly mention it.
\end{enumerate}
\end{theo}

In Table~\ref{tab:init_examples}, we have listed several initialization methods. Here, we would like to suggest a general initialization method as follows:
\begin{equation}
    \bx^{l+1} = \bx^{l} +  f(\bx^{l}; {\nu}_2 \odot \bW).
\label{eq-rescut2222}
\end{equation}

In this method, $\bW$ is initialized using Xavier initialization, and ${\nu}_2$ is a fixed parameter that is pre-set and used only once during the initialization stage of the network. Two suggested choices for $\nu_2$ are $\frac{1}{\sqrt{L}}$ or $\frac{1}{L}$. These choices correspond to different Lipschitz constants. When $\nu_2 = \frac{1}{\sqrt{L}}$, the depth-aware initialization~\cite{depth_scale_zhang2019improving} follows the distribution:
\begin{equation*}
\small
\begin{aligned}
    W_{i,j} & \sim \operatorname{U}\left(-\sqrt{\frac{6}{n_{in}+n_{out}}} \frac{1}{\sqrt{L}}, \sqrt{\frac{6}{n_{in}+n_{out}} } \frac{1}{\sqrt{L}}\right) \quad \text{or}, \\
    W_{i, j} & \sim \operatorname{N}\left(0, \frac{2}{n_{in}+n_{out}} \frac{1}{L} \right)
\end{aligned}
\end{equation*}

For smaller models, weight initialization is not sensitive for network training. However, for larger models (e.g., 175 billion parameters or larger), weight initialization becomes more important.

In Remark~\ref{remark_init}, we have presented several remarks about initialization.

Here, we will not discuss the Lipschitz constants of Fixup~\citep{fixup_zhang2019fixup}, DS-Init~\citep{depth_scale_zhang2019improving}, Admin~\citep{admin_liu2020understanding}, Deepnet~\citep{deepnet_wang2022deepnet}, and ReZero~\citep{rezero_bachlechner2021rezero} operators. However, it is worth noting that these works on network initialization can be reconsidered from the perspective of constraining the Lipschitz constant. Interested readers can calculate the corresponding values in their initializations.

\begin{theo}[DropPath]{remark:droppath}
\begin{enumerate}[leftmargin=*]
    \item DropPath is an effective method to mitigate overfitting. \\
    \item In the training stage, DropPath is also an effective way to stabilize  training by constraining the Lipschitz constant of a network.
\end{enumerate}
\end{theo}

\subsection{DropPath}
DropPath~\citep{droppath_huang2017densely} is another effective technique for training deep transformers. It can be defined as follows:
\begin{equation}
    \by =\left\{
\begin{array}{ll}
\bx, & \text { if the residual path is dropped} \\
\bx + {\rho} \cdot f(\bx), & \text { otherwise }
\end{array}
\right.
\end{equation}

When using DropPath with a drop probability $p$ within each residual block, the Lipschitz constant of LipsFormer is refined as:
\begin{equation*}
\operatorname{Lip}({F}) \leq \prod_{s=1}^{S} \prod_{m=1}^{M_s} (1 + \operatorname{DropPath}(\rho_{s,m} \operatorname{Lip}(f_{s,m}), p)),
\label{def-droppathLips}
\end{equation*}

where
\begin{equation*}
   \operatorname{DropPath}(\alpha_{s,m},p) = \{
\begin{array}{ll}
0, & \text { with probability $p$} \\
\alpha_{s,m} \operatorname{Lip}(f_{s,m}), & \text { with probability $1-p$}.
\end{array} 
\end{equation*}

DropPath effectively decreases the upper bound of a network's Lipschitz constant by randomly dropping the contributions of residual paths.

While DropPath is widely used in Vision Transformers (ViT), it is not often used in language Transformers. One possible reason is that for vision problems like ImageNet, overfitting is more common, whereas for large language models, overfitting is not a concern due to the availability of rich training data~\citep{pile_gao2020pile, roots_laurenccon2022bigscience}.
Analyzing the variation of the Lipschitz constant after applying Dropout~\citep{dropout_srivastava2014dropout} is not an easy task and requires further research.

\section{Explicit Optimization in Deep Learning}
\label{sec:explicit_optimization}
According to the definition in Table~\ref{table:explicit_optimization}, we define explicit optimization as including operations on weight $\bW$, gradient $\displaystyle \nabla_{\bW} \mathcal{L}$, learning rate $\mathbf{\alpha}$, and weight decay factor $\lambda$. In this section, we discuss each factor and its impact on optimization. Additionally, we provide several remarks about each factor.

\subsection{On Choice of Optimizer}
Before delving into each factor, let us discuss general guidelines for selecting a suitable optimizer. We have analyzed that ResNet is a homogeneous network, while Transformer is a heterogeneous network. In ResNet, since each sub-module is homogeneous, there is no quantitative difference in the Lipschitz constants of each sub-module, allowing us to choose an optional learning rate. However, in Transformer, the Lipschitz constant of each sub-module varies significantly. As a result, we can only select a minimal learning rate to ensure stable training, but this may degrade the network's performance.

Classical optimization primarily focuses on SGD and its variants, such as mSGD~\citep{msgd_nesterov1983method} and SVRG~\citep{svrg_johnson2013accelerating}. However, SGD methods have disadvantages in deep learning, especially in heterogeneous networks. We expect that more researchers will focus on adaptive learning rate methods. Overall, AdamW is one of the best-performing methods in almost all types of networks. There have also been several analyses~\citep{adaminstability_molybog2023theory, on_adam_convergence_reddi2019convergence, adam_convergence_chen2018convergence} on the convergence rate of Adam.

\begin{theo}[On Choice of Optimizer]{remark:Optimizer}
\begin{enumerate}[leftmargin=*]
\item Adaptive learning rate methods (e.g., Adam, RMSProp, AdaGrad) perform much better than SGD on heterogeneous networks (e.g., Transformer and RNN), making them a better choice for Transformer and RNN over SGD. \\
\item The learning rate in SGD is sensitive to the Lipschitz constant of the network, while Adam is more robust to the Lipschitz constant due to its use of a normalized update value (the element-wise division between the first-order momentum and the square root of the second-order momentum). \\
\item Both SGD and Adam are suitable for convolutional networks (homogeneous networks), especially for shallow networks. In some shallow convolutional networks, SGD may outperform Adam. However, as the network depth increases, Adam becomes more competitive and outperforms SGD. \\
\item The learning rate in SGD is sensitive to the Lipschitz constants of all layers, which is closely related to the Jacobian matrix. On the other hand, Adam leverages a normalized operator, making its learning rate less sensitive to the Jacobian matrix compared to SGD. \\
\item Weight decay is an effective way to stabilize network training by constraining the Lipschitz constant of the network. It acts as a contraction mapping and consistently improves performance. \\
\item AdamW improves Adam by rectifying the weight decay term. The original Adam uses a wrong weight decay scheduler. \\
\item The default parameters ($\beta_1=0.9$, $\beta_2=0.999$) are not optimal. When the input data has a large noise, the loss may not be stable. A suitable choice is to use ($\beta_1=0.9$, $\beta_2=0.98$) or ($\beta_1=0.9, \beta_2=0.95$).
\end{enumerate}
\end{theo}

In Remark~\ref{remark:Optimizer}, we have provided several remarks on the choice of optimizer. In the experiments, we will observe that the Jacobian matrix of a heterogeneous network varies significantly across all layers, indicating that the gradients in different layers vary significantly. This necessitates the selection of a very small learning rate to prevent exploding gradients. However, this approach compromises the network's representation ability. Adaptive learning rate methods can effectively mitigate this issue. For instance, Adam normalizes the gradient by dividing the first-order momentum by the square root of the second-order momentum.

\subsection{On Weight}
\label{sec:on_weight}
For the optimizer in deep learning, most works focus on the gradient, such as first-order and second-order momentum, and variance reduction in multiple steps of gradients. There are few works that focus on the weight operator. Initialization methods are one example of focusing on the weight, but they are only applied once at the beginning of training.

\begin{theo}[On Weight]{remark:weight}
\begin{enumerate}[leftmargin=*]
    \item Initialization of the weight matrix is important as it significantly affects the training stability and the final representation ability of the network. \\
    \item The eigenvalues of the weight matrix determine the Lipschitz constant of each sub-module. Unstable training often occurs when the eigenvalues of the weight matrix increase rapidly. An possible choice is to clamp the maximum absolute eigenvalue in the training process to constraint the Lipschitz constant of the network as in BigGAN~\citep{big_gan_brock2018large}. \\
    \item Re-parameterization is an effective approach to mitigating the negative effects of fast-growing Lipschitz constants, which can cause instability in network training. Examples of re-parameterization include weight normalization and scaled cosine similarity attention. \\
    \item Exponential Moving Average (EMA) is a useful technique for improving the generalization ability of the model.
\end{enumerate}
\end{theo}

We have presented several remarks about the weight operator in Remark~\ref{remark:weight}. The eigenvalues of the weight matrix (e.g., FC or Convolution) or the norm of the vector ($\gamma$ in LN or BN) reflect the properties of the network. In the experiments, we will investigate how the weight matrix varies along with the training process. 

Re-parameterization, which involves changing the parameters or variables of a model, is an effective way to facilitate learning, improve numerical stability, or enable certain types of computation. It is widely used in deep learning, such as in BN~\citep{batchnorm_ioffe2015batch}, WeightNorm~\citep{weight_normalization_salimans2016weight}, and Scaled cosine similarity attention~\citep{lipsformer_qi2023lipsformer}. 

Exponential Moving Average (EMA) is a technique used to improve the generalization ability of a model. It is commonly used in small and medium-sized models, but it requires storing a copy of the weights in memory. It should be noted that EMA is sensitive to FP16 precision.

\subsection{On Gradient}
As shown in Algorithm~\ref{alg:sgd}, in SGD, the update value is $\color{blue}{\bv_{t}}=
				\beta_{t} \bv_{t-1} + (1-\beta_{t}) \bg_{t}$. For the $l$-th layer, its $\bg^l = \frac{\partial \mathcal{L}}{\partial \bW^l} = \frac{\partial \mathcal{L}}{\partial \bx^{l+1}} {\bx^{l}}^{\top}$. 
If $\frac{\partial \mathcal{L}}{\partial \bx^{l+1}}$ is unbounded, resulting in unbounded gradient values $\bg^l$. Deeper models tend to have larger ranges of gradient values with high probability. Additionally, the ranges of gradient values across different layers can vary significantly. Therefore, using a single learning rate for all layers may not be suitable. However, for simplicity, most SGD-based methods employ this strategy.

As shown in Algorithm~\ref{alg:adam}, in Adam, the updated value is ${\color{green}{\boldsymbol{\mu}_{t}}} = \frac{{\boldsymbol{m}}_{t}}{\boldsymbol{v}_t} = (\beta_{1} \boldsymbol{m}_{t-1}+\left(1-\beta_{1}\right) \boldsymbol{g}_{t})  \oslash {\sqrt{\beta_{2} \boldsymbol{v}_{t-1}+\left(1-\beta_{2}\right) \boldsymbol{g}_{t}^{2}}}.$ 
When $\operatorname{abs}(\bg_t) \gg \operatorname{abs}(\bmm_{t-1})$, the absolute value of ${\color{green}{\boldsymbol{\mu}_{t}}}$ will approximately be $\frac{1-\beta_1}{\sqrt{1-\beta_2}}$. For example, when using the default parameters $(\beta_1=0.9, \beta_2=0.999)$ in Adam, $\frac{1-\beta_1}{\sqrt{1-\beta_2}} = \sqrt{10}$. Thus, the range of the updated value ${\boldsymbol{\mu}_{t}}$ is approximately $\left[-\sqrt{10}, \sqrt{10}\right]$. If we use $(\beta_1=0.9, \beta_2=0.99)$, then the range of the updated value becomes $\left[-1.0, 1.0\right]$. When $(\beta_1=0.0, \beta_2=0.0)$, Adam is equivalent to signSGD~\citep{signsgd_bernstein2018signsgd}.

Compared to SGD, Adam provides a bounded update value to the weights, making it more stable during network training. This partly explains why a learning rate of 5e-4 is often effective for small and shallow networks. However, even for shallow networks, tuning the learning rate multiple times may still be necessary. In contrast, Adam allows each layer to actively learn since the ranges of values in different layers are comparable. In SGD, due to issues like vanishing or exploding gradients, only certain layers (usually higher or shallower layers) receive significant updates while others are not strongly updated. This can lead to weaker representation ability compared to Adam.

\begin{theo}[On Gradient]{remark:gradient}
\begin{enumerate}[leftmargin=*]
    \item In Adam, the absolute value of the update value $\left|\boldsymbol{\mu}_{t}\right|$ is bounded by $\frac{1-\beta_1}{\sqrt{1-\beta_2}}$. In SGD, the update value $\bv_{t}$ is not bounded and is influenced by the Jacobian matrix and the input activation. \\
    \item NAN and INF values are often encountered in LayerNorm and Self-Attention due to their unbounded or very large Lipschitz constants. \\
    \item The variance of Lipschitz constants in Transformer is larger than that in ResNet. As a result, the gradients in different layers exhibit larger variations in Transformer compared to ResNet.
\end{enumerate}
\end{theo}

Gradient clipping is a common technique used in deep learning. It is important to note that gradient clipping is typically applied after the entire back-propagation process is completed. Therefore, gradient clipping cannot solve the NAN or INF problems that may occur within the current batch, but it can influence the weights in the next batch. One suitable approach is to apply gradient clipping on-the-fly during training.

\subsection{On Learning Rate}
\label{sec:learning_rate}
In classical numerical optimization literature \citep{book_nesterov_nesterov2003introductory, numerical_optimization_nocedal1999numerical, book_high_dimensional_data_wright2022high}, the optimal learning rate is typically chosen as $\frac{1}{K_1}$, assuming that the function is $K_1$-smooth. If the learning rate exceeds $\frac{2}{K_1}$, the training process is likely to result in an explosion.

\begin{theo}[on Learning Rate]{remark:learningrate}
\begin{enumerate}[leftmargin=*]
    \item The choice of learning rate in SGD is strongly influenced by the network structure, including its depth and width (see Remark 1 on Gradient).\\
    \item Larger models require smaller learning rates because their Lipschitz constant $K_0$ tends to be larger than that of smaller models.\\
    \item Warmup duration is closely related to the Lipschitz constant $K_0$. Generally, larger models require longer warmup periods.\\
    \item The learning rate should decrease during the training process because the Lipschitz constant of the network usually increases as training progress.
\end{enumerate}
\end{theo}

In general, the choice of learning rate should take into account the constants $K_0$ and $K_1$. However, in practice, estimating the Lipschitz constant and Lipschitz gradient constant for each layer, let alone the entire network, is challenging. This makes it difficult to determine the optimal learning rate. Using an optimal learning rate would ensure faster convergence.

Given that the Lipschitz constant and Lipschitz gradient constant of each module in a Transformer vary more significantly compared to those in a ResNet, classical SGD is not well-suited for Transformer models. Adaptive learning rate methods like Adam are more suitable for Transformers.

\subsection{On Weight Decay}
\label{sec:wd_understanding}
In mathematics, the weight decay operator is represented as follows:
\begin{equation}
    \bW_{new} = (1-\alpha \lambda) \bW,
\end{equation}
where $\lambda$ is a preset weight decay parameter and $\alpha$ is the learning rate. For example, we can set $\lambda = 0.1$ and $\alpha = 5e-4$.

Applying the weight decay operator to the parameters $\bW$ will result in a decrease in the Lipschitz constant of the module. Let us consider a Feed-Forward Network (FFN) as defined in Equation~\ref{eq:ffn_equation} as an example. The original Lipschitz constant of an FFN module is given by $\sigma_{max}(\bW_1) \cdot \sigma_{max}(\bW_2)$, and after applying weight decay, the Lipschitz constant of the FFN becomes:
\begin{equation}
    \operatorname{Lip}(\operatorname{FFN}) = (1-\alpha \lambda)^2 \cdot \sigma_{max}(\bW_1) \cdot \sigma_{max}(\bW_2).
\end{equation}

\begin{theo}[on Weight Decay]{remark:weightdecay}
\begin{enumerate}[leftmargin=*]
    \item The assumption for weight decay is that $W_{ij}$ follows a $\mathcal{N}(\mu = 0,\,\sigma^{2})$ distribution. Under this prior assumption, $\gamma$ in BatchNorm, LayerNorm, and the scale factor $\nu$ in ReZero~\citep{rezero_bachlechner2021rezero} and WRS~\citep{lipsformer_qi2023lipsformer} should not use weight decay. \\ 
    \item Weight decay can accelerate training convergence. The choice of weight decay depends on the training epochs, with longer training requiring smaller decay values. \\
    \item Weight decay can decrease the Lipschitz constant of a network, acting as a contraction mapping.
\end{enumerate}
\end{theo}

From the above equation, we observe that each weight decay operator reduces the Lipschitz constant of the network. This reduction is particularly important for large models because, after gradient updates, the Lipschitz constant of the network tends to increase. If we do not counteract this trend with weight decay, the network training can become more unstable.

We have made several remarks in Remark~\ref{remark:weightdecay}. A potential assumption is that weight value admits a Gaussian distribution $\mathcal{N}(\mu = 0,\,\sigma^{2})$. If the prior value of the weight does not admit Gaussian distribution, one should not use weight decay, or, it will degrade the performance. For instance, the $\gamma$ in LN and BN has a assumptive
 value 1.0. Thus, the weight decay should not be applied on the $\gamma$ term. A general strategy is to enforce larger weight decay for bigger
models. Weight decay is a contraction mapping.

\section{A Guideline for Deep Learning Optimization}
\label{sec:guideline}
In this section, we will compile some guidelines for deep learning optimization based on our previous analysis and discussion. 

\subsection{Guideline for Exploding Gradient}
\label{sec:subsectionofgradientexploding}
\begin{figure}[thb] \centering
    \includegraphics[width=0.8\textwidth]{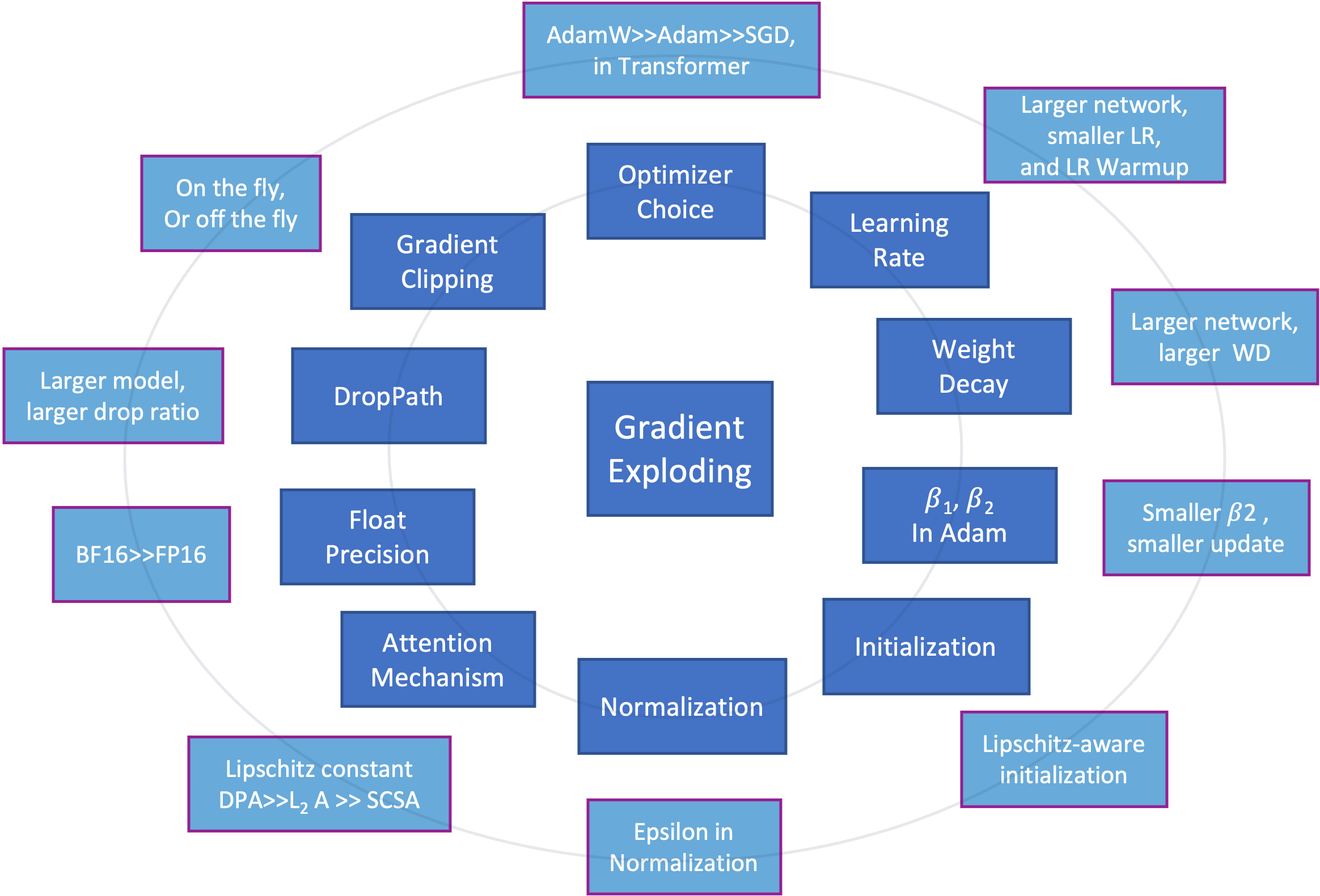}
    \caption{A general guideline for solving exploding gradient problem.} 
    \label{fig:guideline_exploding}
\end{figure}
For the problem of exploding gradients, we have compiled ten guidelines that need to be carefully considered:

\textbf{1. Optimizer choice}: The choice of optimizer is crucial for training a neural network. As discussed in Section~\ref{sec:intro_resnetandtransformer}, ResNet has homogeneous blocks with comparable Lipschitz constants in each block, while Transformer has heterogeneous blocks (Self-attention and FFN) with extremely diverse Lipschitz constants. SGD is sensitive to the Lipschitz constant of the network, while Adam is more robust to variations in the Lipschitz constant due to its element-wise division between first-order momentum and the square root of second-order momentum in the weight update term. In conclusion, Adam (along with other adaptive learning rate optimizers) is a better option for Transformer than SGD. AdamW further improves upon Adam.

After choosing an optimizer like AdamW, we need to consider the learning rate, weight decay, and the updated weight. These can be understood from the following equation:
\begin{equation}
    \boldsymbol{w}_{t} = \boldsymbol{w}_{t-1} - \alpha_{t}{\color{green}{\boldsymbol{\mu}_{t}}} - \alpha_{t} \lambda_{t} \boldsymbol{w}_{t-1},
\end{equation}
which is shown in Algorithm~\ref{alg:adam}. Let us consider each factor one-by-one.

\textbf{2. Learning rate}: The learning rate is the parameter that is most frequently adjusted. It is typically inversely related to the Lipschitz constant of the network. Larger networks generally have larger Lipschitz constants. A general guideline for setting the learning rate is to use a smaller learning rate for larger networks (deeper and wider). In the initial state, Transformer has a very sharp landscape, so warmup is always necessary to smooth the landscape. Deeper networks will require a longer warmup period.

\textbf{3. Weight decay}. As we discussed in Section~\ref{sec:wd_understanding}, weight decay is an effective way to constrain the growing trend of the Lipschitz constant of the network during training. Weight decay acts as a contraction mapping in nature. Generally, larger networks should use a larger weight decay. It should be noted that the parameters $w$ that use weight decay should have a prior that $w \in \mathcal{N}(\mu = 0,\,\sigma^{2})$. Some terms, such as $\gamma$ in LayerNorm and BatchNorm, has a prior value 1.0. These terms should not use weight decay, or it will decrease the performance.

\textbf{4. Adam parameters}. We have presented the AdamW algorithm in Algorithm~\ref{alg:adam} and discussed the influence of parameters $\beta_1$ and $\beta_2$. As mentioned earlier, the range of the updated value is determined by these parameters. In PyTorch~\citep{pytorch_paszke2017automatic}~\footnote{\url{https://pytorch.org/docs/stable/generated/torch.optim.AdamW.html}}, the default values are $\beta_1=0.9$ and $\beta_2=0.999$. With these default parameters, $\frac{1-\beta_1}{\sqrt{1-\beta_2}} = \sqrt{10}$. Consequently, the range of the updated value ${\color{green}{\boldsymbol{\mu}_{t}}}$ becomes $\left[-\sqrt{10}, \sqrt{10}\right]$. Such a large range can lead to unstable training, especially when data have noise or incorrect labels, causing $\operatorname{abs}(\bg_t) \gg \operatorname{abs}(\bmm_{t-1})$ (see Algorithm~\ref{alg:adam}). By using $(\beta_1=0.9, \beta_2=0.99)$, the range of the updated value becomes $\left[-1.0, 1.0\right]$. Similarly, when using $(\beta_1=0.9, \beta_2=0.95)$, the range of the updated value is $\left[-0.447, 0.447\right]$.

\textbf{5. Initialization}. Initialization is crucial for training a neural network, especially when the network is deep. Some classical initialization methods were proposed without considering very deep networks. In deep networks, the Lipschitz constant of the network can become very large. If we continue to use classical initialization methods, the training process becomes prone to instability. One possible solution is to employ Lipschitz-aware initialization, which is equivalent to depth-aware initialization~\citep{depth_scale_zhang2019improving} in specific implementations. Lipschitz-aware initialization allows us to theoretically constraint the Lipschitz constant of the network to a known value in its initial state. 

\textbf{6. Normalization}. Normalization is an extremely effective module for smoothing the landscape of the network. This will be further demonstrated in the experimental section. As discussed in Section~\ref{sec:implicit_optimization}, normalization enforces network activations to satisfy forward principles. However, from a backward perspective, it's important to note that LN (LayerNorm) has potential problems. Non-smoothing LN is not Lipschitz continuous, while smoothing LN is Lipschitz continuous but with a very large Lipschitz constant. As shown in Table~\ref{tab:bigtable}, the Lipschitz constant of LN is very large when considering very small $\epsilon$, such as $\epsilon = 10^{-8}$. Equation~\ref{eq:normalization_stability}, we have shown the stability of different normalizations.

\textbf{7. Self-attention mechanism}. Self-attention is a high-order nonlinear operator with powerful representation abilities. However, we have observed that the Lipschitz constant of dot-product attention (DPA), as shown in Table~\ref{tab:bigtable}, is unbounded, which can lead to overflow problems. As alternatives, we can consider $L_2$ attention ($L_2$A) and scaled cosine similarity attention (SCSA). Based on the calculated Lipschitz constants of these three attention mechanisms, we can expect that SCSA and $L_2$A will exhibit more stable properties than DPA.
Experiments will further validate that SCSA will smooth the landscape of the Transformer better than DPA.

\textbf{8. Floating-point precision}. In most current training models, mixed precision training is used, where the forward and backward computations are performed with FP16 precision, and the weight updates are done in FP32. However, this introduces a problem: in mixed precision training, we are more prone to encountering overflow issues compared to using FP32 throughout. According to Table~\ref{tab:bigtable}, normalization and self-attention modules are particularly prone to precision problems (overflow) due to their higher Lipschitz constants compared to other modules such as Convolution and FFN. One possible alternative to FP16 is to use BF16 or even FP32, which has a larger integer range. Higher float precision can only be only applied to some unstable modules instead of the whole network.
While this can partially address the problem, there is still a significant possibility of encountering overflow. A better strategy is to use powerful and stable normalization and self-attention modules.

\textbf{9. DropPath}. DropPath is an effective method for mitigating overfitting in network training. In computer vision, it is common to train models on the training data for hundreds of epochs. Without using DropPath, it is easy to overfit the training data. In large language models (LLMs)~\citep{llama_touvron2023llama, palm_chowdhery2022palm, chinchilla_hoffmann2022training, opt_zhang2022opt, openai_gpt4}, overfitting is generally less of a concern due to the abundance of data (e.g., trillions of tokens)~\citep{pile_gao2020pile,roots_laurenccon2022bigscience}. Another benefit of DropPath is that it can reduce the Lipschitz constant of a network at runtime by dropping several layers. In conclusion, in ViT models, DropPath is effective in mitigating overfitting and reducing the Lipschitz constant of the network during runtime.

\textbf{10. Gradient clipping}. Gradient clipping is a widely used technique to prevent the exploding gradient problem. Typically, it is applied as a post-processing step after back-propagation is completed. Alternatively, an on-the-fly approach can be used, where gradient clipping is performed after each layer during back-propagation. Generally, larger models require smaller clipping thresholds.

\subsection{Guideline for Vanishing Gradient}
\begin{figure}[thb] \centering
    \includegraphics[width=0.45\textwidth]{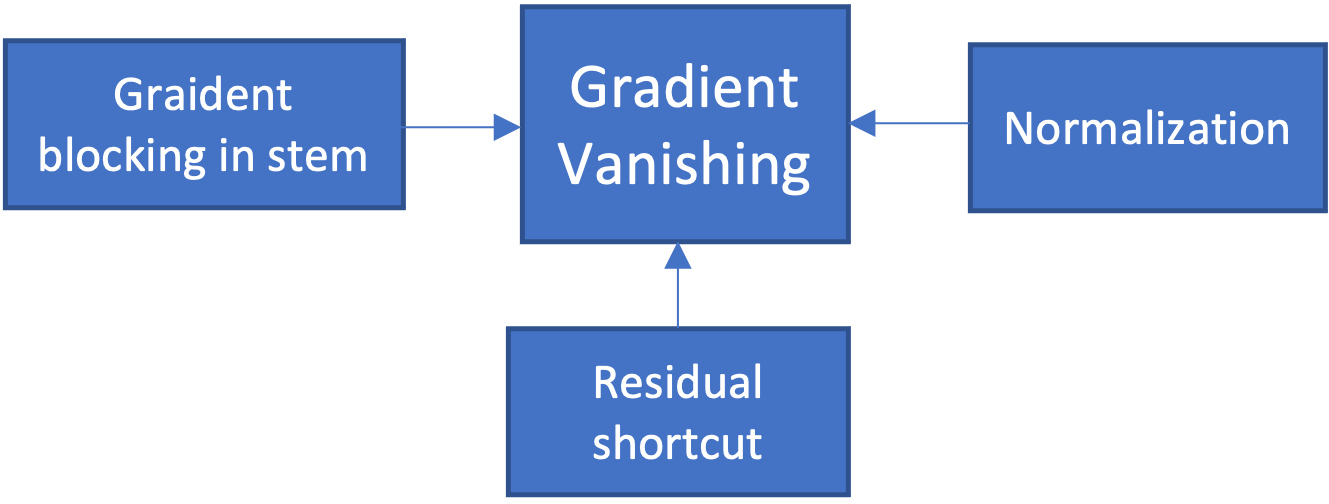}
    \caption{A general guideline for solving vanishing gradient problem.} 
    \label{fig:guideline_exploding}
\end{figure}

As discussed previously, the vanishing gradient problem does not disable the training process but rather leads to a weaker representation in the network. With the introduction of residual shortcuts~\citep{resnet_he2016deep, identity_mapping_he2016identity}, non-gradient saturation activation functions~\citep{relu_dahl2013improving, swish_ramachandran2017searching, gelu_hendrycks2016gaussian}, and various effective normalizations~\citep{batchnorm_ioffe2015batch, layernorm_ba2016layer, weight_normalization_salimans2016weight, rmsnorm_zhang2019root, lipsformer_qi2023lipsformer}, the vanishing gradient problem is no longer a major obstacle for a successful network training. In this subsection, we present three guidelines to debug vanishing gradient problems.

\textbf{1. Residual shortcut}. Residual shortcut, as discussed in Section~\ref{sec:intro_residual}, is a breakthrough technique that effectively addresses the vanishing gradient problem. The residual shortcut is defined as follows:
\begin{equation*}
    \by = \bx + f(\bx; \bW).
\end{equation*}
The Jacobian matrix of $\by$ with respect to $\bx$ is:
\begin{equation*}
\begin{aligned} 
\boldsymbol{J}_{\by}(\bx) = \frac{\partial \by}{\partial \bx} = \frac{\partial \bx}{\partial \bx} + \frac{\partial f(\bx; \bW)}{\partial \bx} = \bI + \frac{\partial f(\bx; \bW)}{\partial \bx}.
\end{aligned}
\end{equation*}
Even if we have gradients close to zero in the branch $\frac{\partial f(\bx, \bW)}{\partial \bx}$, the gradient can still be propagated to lower layers. If  a vanishing gradient problem is encountered, it is important to check whether the residual shortcut is properly utilized in the stem.

\textbf{2. Normalization}. Normalization typically normalizes the activation to a comparable level with a target mean value and standard variance. Let us see how it mitigates the vanishing gradient problem. Suppose $\by = \operatorname{Normalization}(\bx)$ and $\bz = \bW \by$. In the backward process, we have $\frac{\partial \mathcal{L}}{\partial \bz}$. Since $\by$ is a normalized value, the gradients of each point contributing to $\bW$ are evenly distributed. This prevents the situation where the gradient from point $\bx$ to $\bW$ is small due to small activation values in $\bx$. As $\bW$ can obtain correct and valid gradients and $\frac{\partial \mathcal{L}}{\partial \by} = \bW^{\top} \frac{\partial \mathcal{L}}{\partial \bz}$, $\frac{\partial \mathcal{L}}{\partial \by}$ remains valid.

\textbf{3. Gradient blocking in stem}. An often observed phenomenon in vanishing gradient problems is continuous oscillation of the loss around a relatively large value. For example, when training a model on ImageNet, the loss decreases from approximately 10.0 to 7.0 and then oscillates around 7.0. In such cases, it is important to check whether gradients are blocked in one layer in the stem.

\section{Experimental Analysis}
\label{sec:experiments}
In this section, we will analyze the properties of different networks and explore the underlying reasons through experiments. Let us first describe our experimental settings.

Our analysis is based on the simulated Lipschitz constant of various networks. The computational equation is as follows:
\begin{equation}
    K = \operatorname{max}_{\bx, \epsilon, \bz} \frac{\left\|f(\bx+\epsilon \bz; \bW) - f(\bx; \bW) \right\|_p}{\left\| \bx + \epsilon \bz - \bx \right\|_p}    
\end{equation}
Here, $f(\cdot)$ represents a network, $\bx$ denotes a randomly selected point, $\bz$ is a random Gaussian noise, and $\epsilon$ is a small scaling factor.  $\|\|_p$ denotes the $L_p$ norm, and by default, we will use the $L_2$ norm. We will also compare different norms in the experiments. The default settings include $\epsilon = 1e\-7$ and $\bz$ admits a Gaussian distribution with $\mathcal{N}(\mu = 0, \sigma^{2}=1.0)$. Since it is impractical to enumerate all possible values for $\epsilon$ and $\bz$, we will use 10 input points $\bx$, and for each $\bx$, we will select 10 points of $\bz$ to obtain the simulated value of $K$. In practice, we find that the variance is small between several random seeds.

\textbf{What does the value $K$ mean?} The value of $K$ reflects the landscape of a network. If $K$ equals 0, it means that the output does not change with respect to any variation in $\bx$. On the other hand, if $K$ is very large, it indicates that the gradient changes rapidly and the curvature is substantial around the point $\bx$. In conclusion, the value of $K$ describes the landscape of the network.

Let us discuss the settings for the network $f(\cdot)$. By default, both ResNet and Transformer have 12 layers, with a hidden dimension of 1024. For Transformer, we use 8 heads for queries, keys, and values, and the expansion scale in the feed-forward network (FFN) is 4. The input $\bx$ is a randomly created data with a shape of $\text{Width}\times \text{Height}\times \text{Hidden\_dimension}$ by default. We set the default values for Width and Height as 32. For ResNet, we input the tensor directly, while for Transformer, we reshape it into a sequence tensor with a shape of $\text{Length} \times \text{Hidden\_dimension}$, where $\text{Length} = \text{Width}\times \text{Height}$. The final output will have the same shape as the input.

In ResNet, each layer can be calculated as follows:
\begin{equation}
    \bx^{l+1} = f(x^l; \bW^{l+1}) = \bx^l + \operatorname{BN}\left(\operatorname{Conv}\left(\operatorname{ReLU}\left(\operatorname{BN}\left(\operatorname{Conv}\left(\bx^l\right)\right)\right)\right)\right).
\end{equation}

For Transformer, we will evaluate two types of attention mechanisms: dot-product attention and scale cosine similarity attention. We will refer to them as DPA Transformer and SCSA Transformer, respectively. We do not include the $L_2$ distance attention~\citep{l2distance_attention_kim2021lipschitz} because we do not have the implementation of their code. The difference between our implementation and the  authors' code may lead to unobjective assessment.

When evaluating a network without residual shortcuts, we remove all residual shortcuts. For example, in  Transformer, we  remove the residual shortcuts in both the attention and feed-forward network (FFN) modules. Similarly, when evaluating a network without normalization, we remove all normalization in the network. We use BatchNorm for ResNEt and use LayerNorm for Transformer. When using LayerNorm, we use post-norm. 

By default, we use Xavier initialization~\footnote{\url{https://pytorch.org/docs/stable/_modules/torch/nn/init.html\#xavier_normal_}} to initialize the network. To simulate the training process where the eigenvalues of the weight matrix consistently grow, we will use a gain of 2.0 in Xavier initialization in default. This means that after initialization, each weight matrix $\bW$ will be multiplied by 2.0.

We can express the operation as follows:
\begin{equation*}
\small
\begin{aligned}
    \text{First} \ W_{i,j} & \sim \operatorname{N}\left(0, \frac{2}{n_{in}+n_{out}} \right), \ \text{then} \ \ W_{i,j} = 2.0 \times W_{i,j}.
\end{aligned}
\end{equation*}

It is important to note that in this article, our aim is not to achieve state-of-the-art (SOTA) performance. Instead, our goal is to analyze the properties of a network. We set the gain to 2.0 to match the observation that the eigenvalues increase significantly after some training steps. This allows us to uncover potential problems that some modules may have. It should be emphasized again that our goal is not to achieve SOTA or introduce a new method.

In the following experiments, we will vary different parameters to observe how $K$ changes. For example, we will vary the number of layers $L$ across ten different settings: [1, 2, 4, 8, 12, 16, 24, 32, 48, 64], and analyze how $K$ varies accordingly.

\subsection{Why is Transformer Harder to Optimize than ResNet?}
To compare Transformer with ResNet, we vary the number of layers and observe how the $K$ values change across different layers. We also compare them with and without residual shortcuts (shown as solid and dashed lines, respectively).

\begin{figure}[h]
    \centering
    \includegraphics[width=0.55\textwidth]{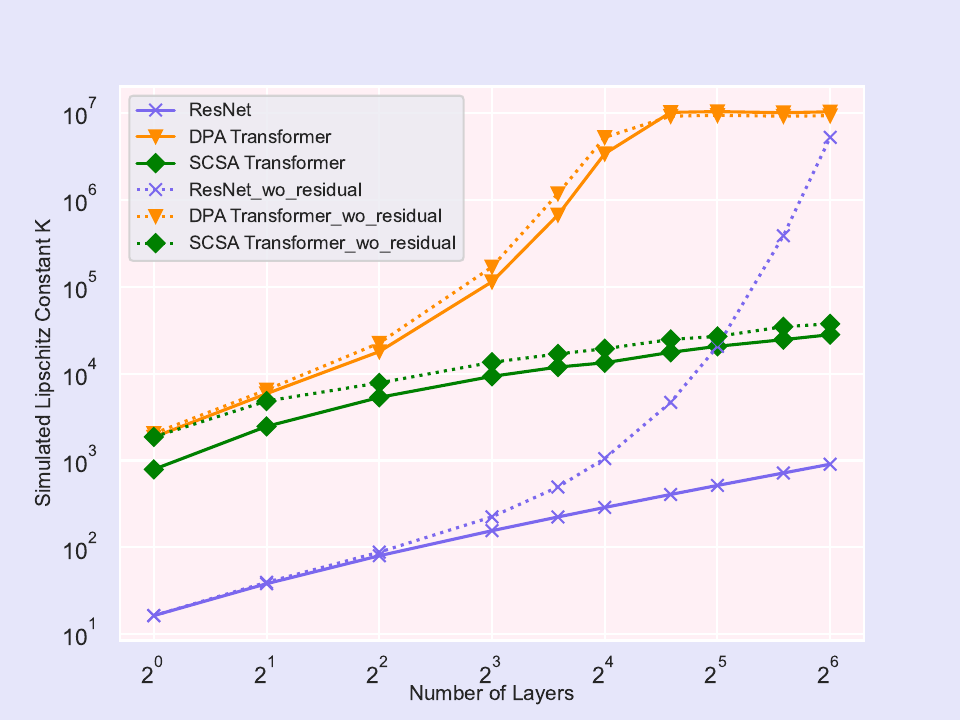}
    \caption{Simulated Lipschitz constants of ResNet, DPA Transformer, SCSA Transformer with or without residual shortcuts for different numbers of layers. The horizontal axis is scaled by $\log_2$, and the vertical axis is scaled by $\log_{10}$ for better visualization.}
    \label{fig:resnet_transfomer_with_without_residual}
\end{figure}

We simulate the Lipschitz constants of ResNet (\textcolor{blue}{blue} color), DPA Transformer (\textcolor{orange}{orange} color), SCSA Transformer (\textcolor{green}{green} color)
 with or without residual shortcuts for different numbers of layers. The results are shown in Figure~\ref{fig:resnet_transfomer_with_without_residual}.

From Figure~\ref{fig:resnet_transfomer_with_without_residual}, we make the following three observations:

\begin{itemize}[leftmargin=*]
    \item \textbf{DPA Transformer becomes harder to optimize as the network becomes deeper.} With an increase in the number of layers, the simulated $K$ value for DPA Transformer increases quickly. For example, when $L=64$, the $K$ value exceeds the maximum range of FP16, leading to an ``INF'' value during the training process.
    \item \textbf{Residual shortcuts effectively smooth the landscape.} ResNet with residual shortcuts (solid blue line) exhibits a slower increase in the $K$ value compared to a very fast increase of ResNet without residual shortcuts (dashed blue line). A smaller $K$ value indicates a smoother landscape. For instance, a 64-layer ResNet with residual shortcuts has a simulated $K$ value of 1e3, which is still smaller than 65504. This means that the learning process will not explode under the current conditions. In our implementation, we use post-layernorm. In this way, residual shortcut do not smooth the landscape as it in ResNet. ResNet uses a pre-norm  way.
    \item \textbf{SCSA attention mechanism exhibits a smoother landscape compared to DPA attention.} We observe that SCSA Transformer with residual shortcuts shows a very slow increase in the $K$ value. This indicates that SCSA attention effectively smooths the landscape. The introduced re-parametrization technique in SCSA works well.
\end{itemize}

These observations highlight the challenges faced in optimizing Transformer models compared to ResNet models. Meanwhile, these observations also verify that 1) residual shortcut can smooth the landscape of a network effectively; 2) Transformer is harder to optimize than ResNet.

\begin{figure}[h]
    \centering
    \includegraphics[width=0.55\textwidth]{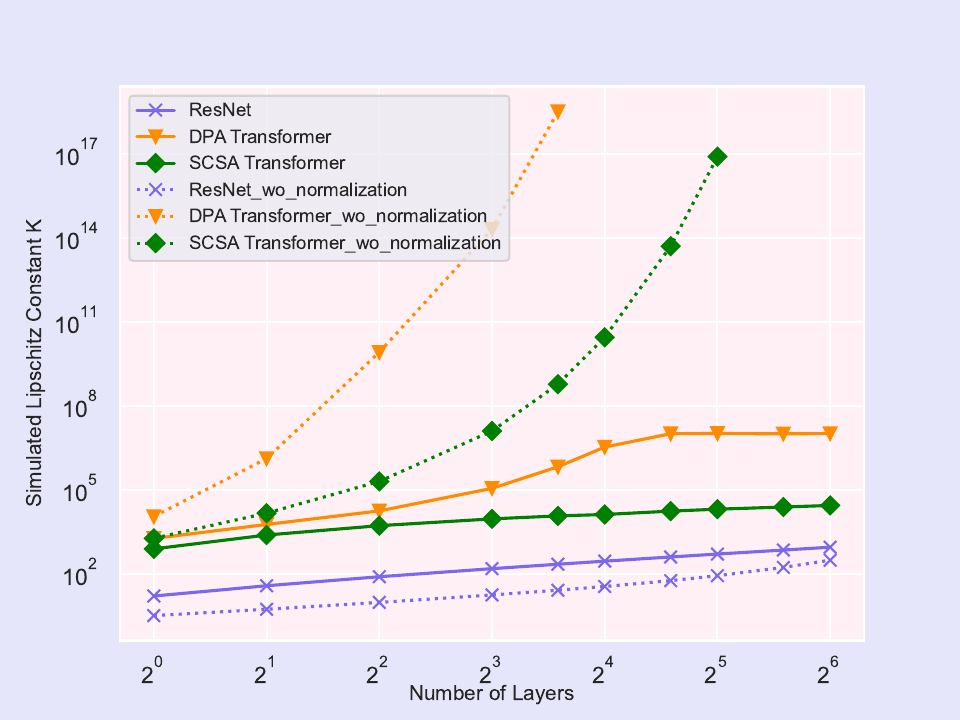}
    \caption{Simulated Lipschitz constants of ResNet, DPA Transformer, SCSA Transformer with or without normalization (BN for ResNet and LN for Transformer) across different numbers of layers. The horizontal axis is scaled by $\log_{2}$, and the vertical axis is scaled by $\log_{10}$ for better visualization. $K$ values of DPA Transformer without normalization larger than 16 layers become INF and thus have not been plotted in the figure. The same issue applies to SCSA Transformer with 64 layers.}
    \label{fig:normalization}
\end{figure}

\subsection{On Normalization}
\label{sec:experiment_norm}
We further compare ResNet, DPA Transformer, and SCSA Transformer with or without normalization across different numbers of layers. The result is shown in Figure~\ref{fig:normalization}.

From Figure~\ref{fig:normalization}, we observe the following:
\begin{itemize}[leftmargin=*]
    \item \textbf{Normalization is extremely effective in smoothing the landscape.} Transformer without LN will quickly experience an explosion when the layer number exceeds 16. Adding LN noticeably smooths the landscape for Transformer. ResNets with or without BN both have a smoothed landscape, as mentioned before, BN can mitigate the gradient vanishing problem.
\end{itemize}

The observation highlight that normalizations can smooth the landscape of the networks effectively. It is an extremely useful skill in deep learning.

\subsection{Simulated Lipschitz Constants across Different Layers}
Here give a point $\bx$, we denote $f^l(\bx, \bW)$ as the output of the $l$-th layer.  We define:
\begin{equation*}
    K_{l0} = \operatorname{max}_{\epsilon, \bz} \frac{\left\|f^l(\bx+\epsilon \bz; \bW) - f^l(\bx; \bW) \right\|_p}{\left\| \bx + \epsilon \bz - \bx \right\|_p}, \ \ \text{for}\ l \ \text{in [0, L]}.    
\end{equation*}
And we define:
\begin{equation*}
    K_{Ll} = \operatorname{max}_{\epsilon, \bz} \frac{\left\|f^L(\bx+\epsilon \bz; \bW) - f^L(\bx; \bW) \right\|_p}{\left\| f^l(\bx+\epsilon \bz; \bW) - f^l(\bx; \bW)  \right\|_p}, \ \    \text{for}\ l \ \text{in [0, L]}. 
\end{equation*}

\begin{figure}[h]
\centering
\begin{subfigure}{.45\textwidth}
  \centering
  \includegraphics[width=.95\linewidth]{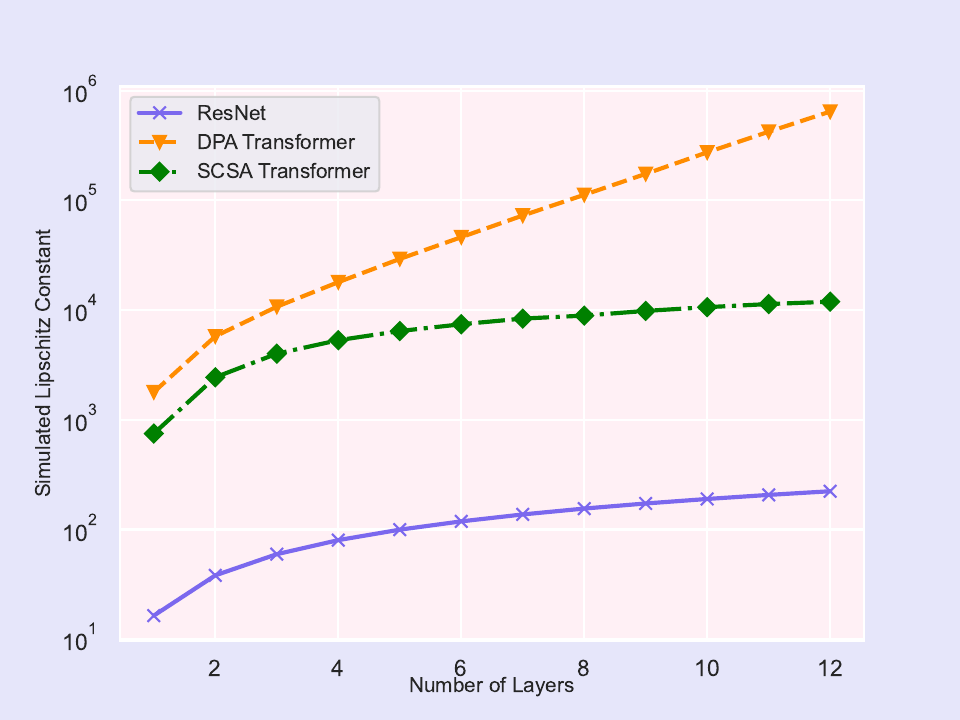}
  \caption{Simulated Lipschitz constant of \\ $K_{l0} = \operatorname{max}_{\epsilon, \bz} \frac{\left\|f^l(\bx+\epsilon \bz; \bW) - f^l(\bx; \bW) \right\|_p}{\left\| \bx + \epsilon \bz - \bx \right\|_p}, \ \ \text{for}\ l \in [0, L]$.}
  \label{fig:sfig1}
\end{subfigure}%
\begin{subfigure}{.45\textwidth}
  \centering
  \includegraphics[width=.95\linewidth]{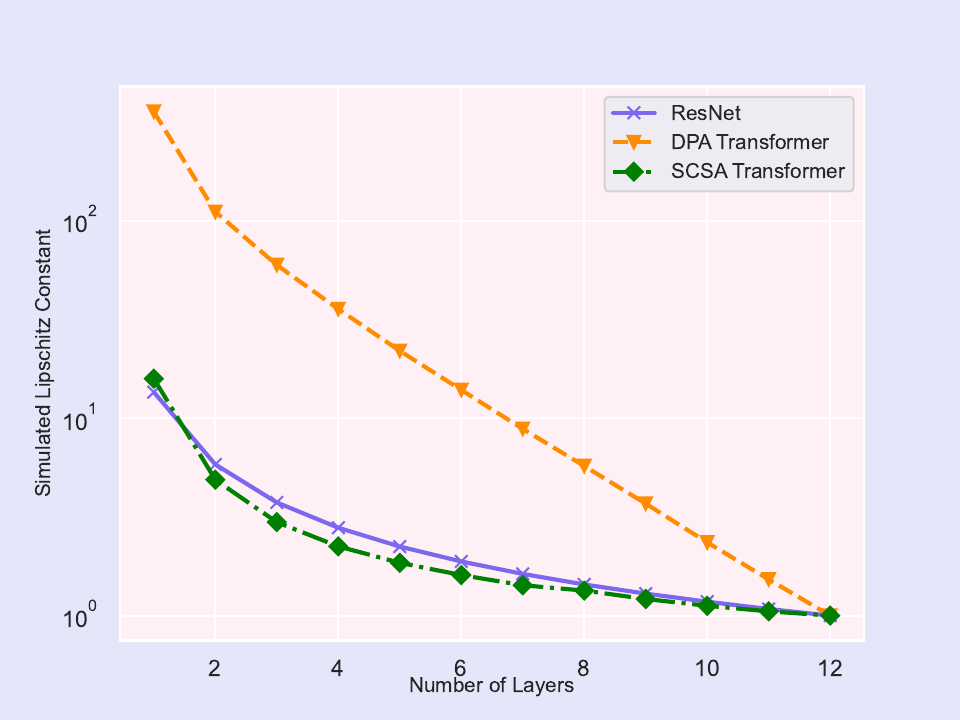}
  \caption{Simulated Lipschitz constant of \\ $K_{Ll} = \operatorname{max}_{\epsilon, \bz} \frac{\left\|f^L(\bx+\epsilon \bz; \bW) - f^L(\bx; \bW) \right\|_p}{\left\| f^l(\bx+\epsilon \bz; \bW) - f^l(\bx; \bW)  \right\|_p}, \ \    \text{for}\ l \in [0, L]$.}
  \label{fig:sfig2}
\end{subfigure}
\caption{Simulated Lipschitz constants $K_{l0}$ and $K_{Ll}$ across different layers}
\label{fig:k1_and_k2}
\end{figure}

In this subsection, we will evaluate the $K_{l0}$ and $K_{Ll}$ values. The results are shown in Figure~\ref{fig:k1_and_k2}. We can observe that the $K_{l0}$ value for the DPA Transformer increases rapidly as the number of layers increases. ResNet exhibits a slower increase, while SCSA Transformer shows the slowest growth rate.
The trend on the right side of Figure~\ref{fig:k1_and_k2} aligns with our expectations.

\subsection{Sensitivity Analysis of Output with Respect to Different Parameters}
\label{sec:experiment_different_weights}
\begin{figure}[h]
\centering
\begin{subfigure}{.42\textwidth}
  \centering
  \includegraphics[width=.95\linewidth]{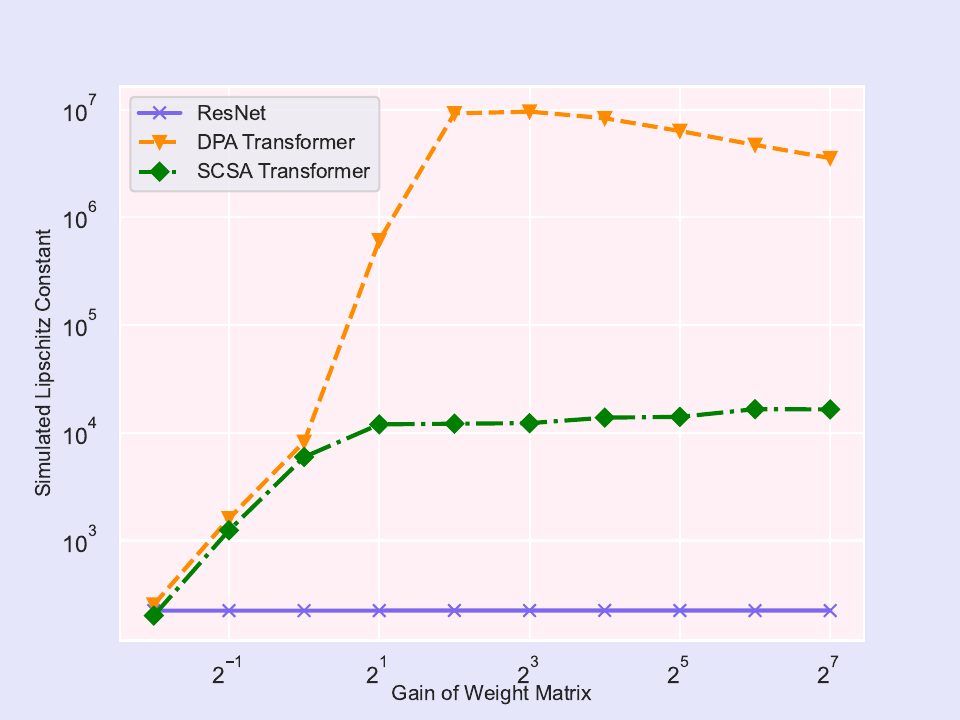}
  \caption{on scale factor of weight matrix}
  \label{fig:sfig2}
\end{subfigure}
\begin{subfigure}{.42\textwidth}
  \centering
  \includegraphics[width=.95\linewidth]{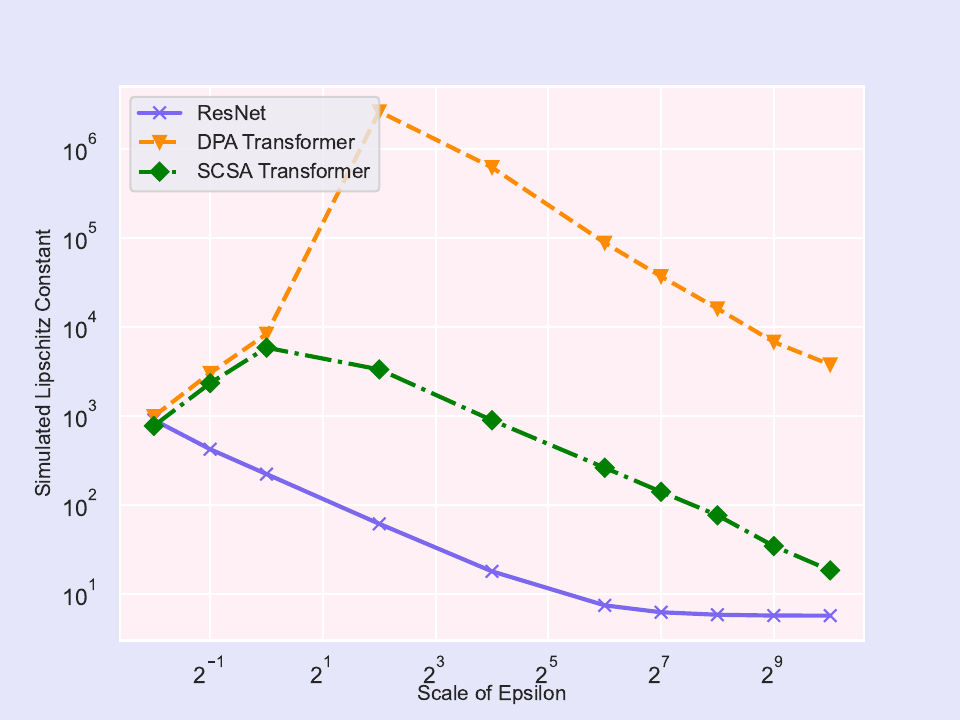}
  \caption{on $\epsilon$.}
  \label{fig:sfig2}
\end{subfigure}
\begin{subfigure}{.42\textwidth}
  \centering
  \includegraphics[width=.95\linewidth]{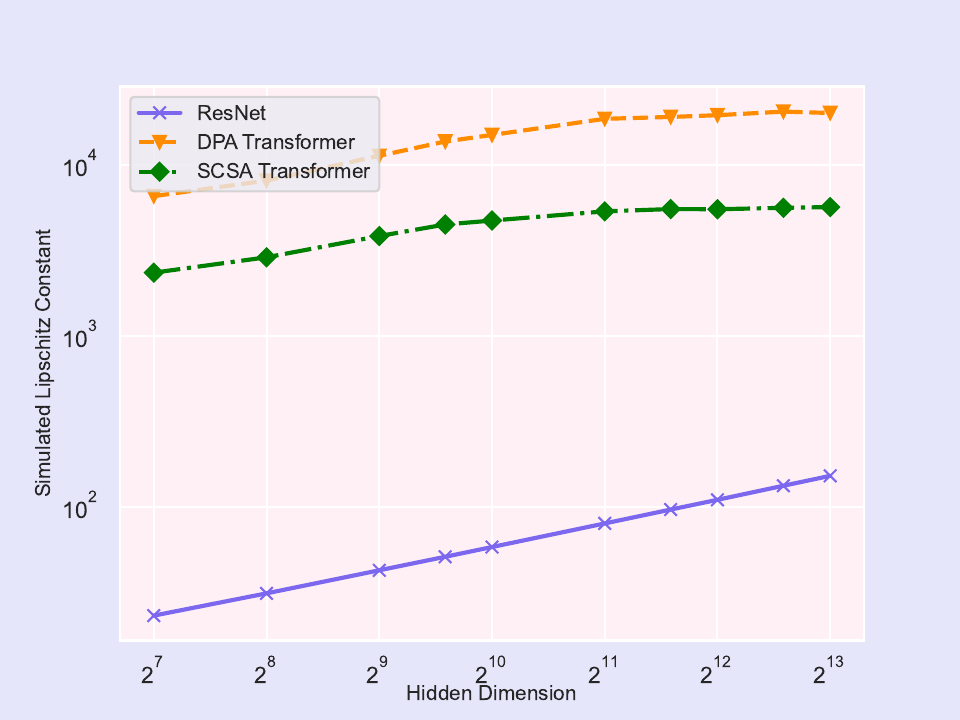}
  \caption{on hidden dimension}
  \label{fig:sfig2}
\end{subfigure}
\begin{subfigure}{.42\textwidth}
\centering
  \includegraphics[width=.95\linewidth]{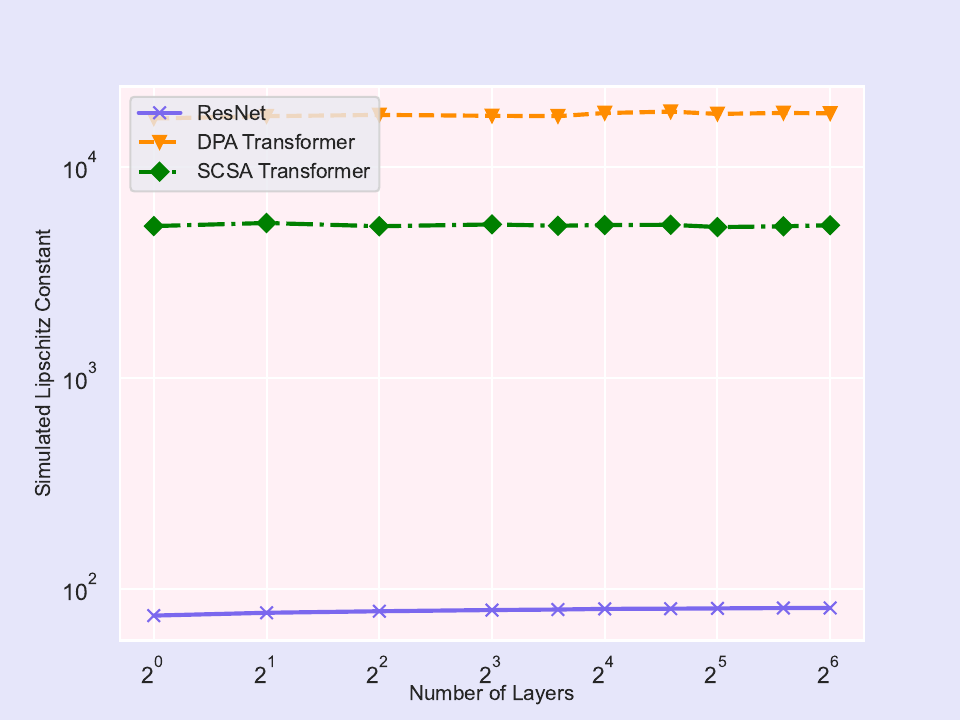}
  \caption{on input size (sequence length)}
  \label{fig:sfig2}
\end{subfigure}
\caption{Evaluation of influence of different parameters, including the scale factor of the gain in the weight matrix, $\epsilon$, hidden dimension and the input size, to the simulated Lipschitz constant value $K$ using ResNet, DPA Transformer and SCSA Transformer.}
\label{fig:four_different_params}
\end{figure}

In this subsection, we further evaluate four different parameters: $\epsilon$, hidden dimension, the scale factor of the gain in the weight matrix, and the input size. $\epsilon$ represents the perturbation distance from $\bx$ to $\bx + \epsilon \bz$. We select ten settings for $\epsilon$: [0.25, 0.5, 1.0, 4.0, 16.0, 64.0, 128.0, 256.0, 512.0, 1024.0]. For the hidden dimension, we choose ten settings: [128, 256, 512, 768, 1024, 2048, 3072, 4096, 6144, 8192]. The weight scale factor (also known as the gain in PyTorch) is varied in ten settings: [0.25, 0.5, 1.0, 2.0, 4.0, 8.0, 16.0, 32.0, 64.0, 128.0]. The input size is selected from the following list: [32, 64, 128, 256, 384, 512, 768, 1024, 1532, 2048]. When evaluating hidden dimension and  input size, we use networks with only 4 layers instead of 12 layers due to memory limitation.
The results are shown in Figure~\ref{fig:four_different_params}.

We have the following observations from Figure~\ref{fig:four_different_params}:
\begin{itemize}[leftmargin=*]
    \item The gain of the weight matrix significantly affects the landscape of the DPA Transformer, while it does not have a significant impact on ResNet.
    \item The $K$ value of the DPA Transformer is greatly affected by the choice of $\epsilon$. When $\epsilon = 4.0$, the DPA Transformer has the highest $K$ value. In contrast, the $K$ value decreases with increasing $\epsilon$ for ResNet and SCSA Transformer.
    \item The $K$ values for all three networks increase along with the hidden dimension. It partly explains that larger models (with wider hidden dimension) are harder to optimize.
    \item As the image size or sequence length increases, the $K$ value does not change too much for all three networks. 
\end{itemize}


\subsection{Sensitivity Analysis of Different Norms}
\label{seq:experiment_different_norms}
\begin{figure}[h]
\centering
\begin{subfigure}{.46\textwidth}
  \centering
  \includegraphics[width=.95\linewidth]{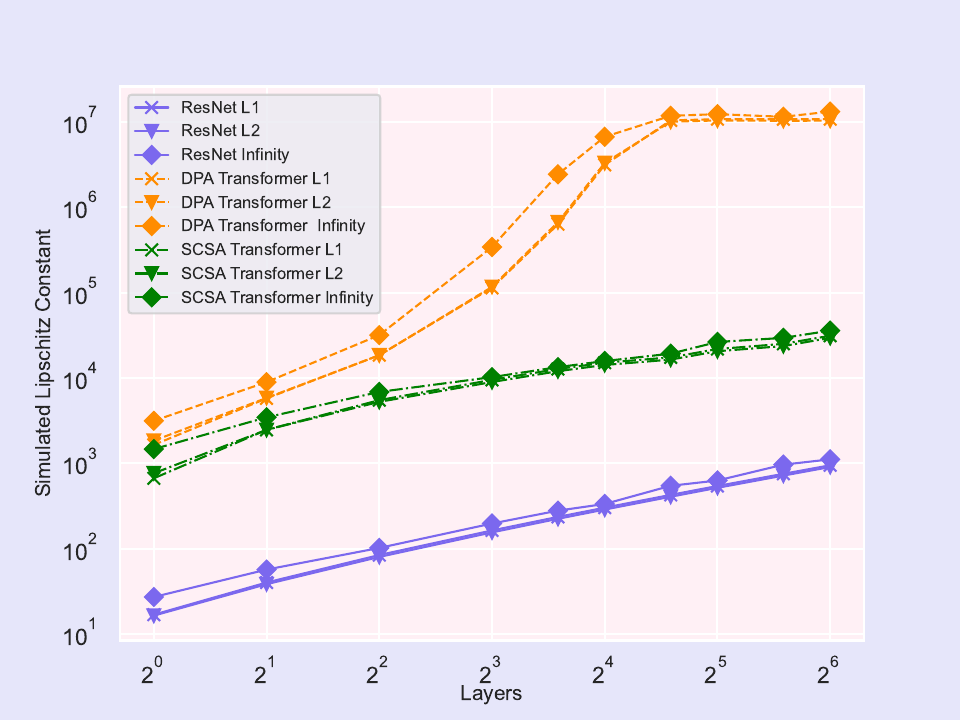}
  \caption{$L_1$, $L_2$, and $L_\infty$}
  \label{fig:differentnorms_1}
\end{subfigure}%
\begin{subfigure}{.46\textwidth}
  \centering
  \includegraphics[width=.95\linewidth]{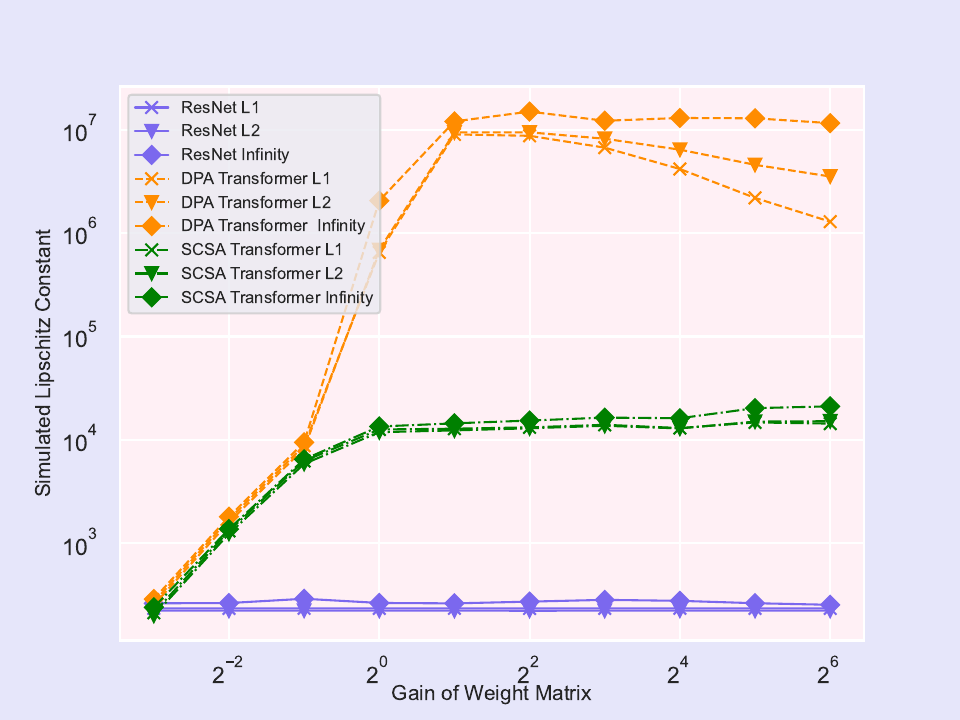}
  \caption{$L_1$, $L_2$, and $L_\infty$}
  \label{fig:differentnorms_2}
\end{subfigure}
\caption{Simulated Lipschitz constant value $K$ of ResNet, DPA Transformer and SCSA Transformer under the metric of $L_1$, $L_2$ and $L_\infty$.}
\label{fig:differentnorms}
\end{figure}

We have calculated the $K$ value under different norms, including $L_1$, $L_2$ and $L_\infty$. The result is shown in Figure~\ref{fig:differentnorms}. 
It shows that the results under different norm metric are consistent.


\section{Discussion}
\label{sec:discussion}
\subsection{The Difficulty of Training Large Models}
For the training of large models, including ViT~\citep{vit_dosovitskiy2020image, swintransformer_liu2021swin, swin_v2_liu2021swin2, wu2021cvt} or Large Language Models~\citep{gpt_radford2018improving, gpt2_radford2019language, gpt3_brown2020language, palm_chowdhery2022palm, openai_gpt4, opt_zhang2022opt, llama_touvron2023llama}, we encounter two types of problems: system-level optimization problems and numerical optimization problems. Regarding system-level optimization, very large models can be trained using 3D parallelism (data, model, and pipeline parallelism) if sufficient hardware resources are available. While this article primarily focuses on numerical optimization, it is worth acknowledging that many current successes of large language models owe credit to system-level optimization techniques, including Zero~\citep{zero_rajbhandari2020zero}, DeepSpeed~\citep{deepspeed_rasley2020deepspeed}, and Megatron~\citep{megatron_smith2022using}.

When referring to large models, we typically mean network with a larger depth $L$ and wider hidden dimension $D$ in each $\bW$. The Lipschitz constant of a network can be calculated by the following equation:
\begin{equation*}
   \operatorname{Lip}(F_{\bx}(\{\bW^l, l=1,\ldots,L \})) \leq \prod_{l=1}^{L} \operatorname{Lip}(f^{l}({\bx}^{l-1}; \bW^l)).
\label{eq:model_lipschitz}
\end{equation*}
From the above equation, we can see that the Lipschitz constant of the network is a multiplier of the Lipschitz constant of each layer. Deeper networks entail more terms in the multiplication equation. Typically, the Lipschitz constant in each layer is greater than 1.0, meaning that deeper networks tend to have larger Lipschitz constants. Consequently, deeper networks, with their often large Lipschitz constants, are more likely to violate the principles of forward and backward optimization.

Let us consider the influence of network width $D$. In each training step, the update equation is given by
\begin{equation*}
\bW_{new}^l = \bW^l - \mathbf{\alpha} \displaystyle \nabla_{\bW^l} \mathcal{L} - \alpha \lambda  \bW^l.
\label{eq:weight_update_mapping}
\end{equation*}
It is unclear how the dimension $D$ affects the training process, but we observe that larger $D$ usually has larger probability to have a larger absolute eigenvalue. In this way, it will enlarge the Lipschitz constant of a network, and it will make the training harder.

For the dynamics of the training process, it is challenging to determine whether $\bW_{new}^l = \bW^l - \mathbf{\alpha} \displaystyle \nabla_{\bW^l} \mathcal{L}$ is a contraction mapping or an expansion mapping. If it is an expansion mapping, what is the expansion factor after this operation? Based on experimental observations, we know that in an unstable training process, the weight update corresponds to an expansion mapping, causing the eigenvalues of the weight matrix to increase rapidly. Currently, our understanding of the properties of weight updates remains limited. We believe that Random Matrix Theory (RMT)~\citep{rmt_edelman2005random} might shed some light on this problem. But currently, it remains unclear to us.

\subsection{Open Questions}

\textbf{The properties of weight update in optimizers.} Is the weight update function shown in Equation~\ref{eq:weight_update_mapping} a contraction mapping or an expansion mapping? If it is an expansion mapping, what is the expansion factor? This understanding is crucial for deep learning optimization. Currently, there is limited research on this topic, and we hope to see future works addressing this problem.

\textbf{Automatic setup and adjustment of learning rate and weight decay.} Given the weight update equation mentioned earlier, how can algorithms select optimal values for $\alpha$ and $\lambda$ to ensure stable weight updates? Currently, the choices of $\alpha$ and $\lambda$ are mainly based on researchers' empirical experience.

\textbf{The relationship between representation ability and training stability.} As discussed in the paper, we can constrain the Lipschitz constant of the network during initialization or training. For example, we can use a Lipschitz-aware initialization as follows:
\begin{equation*}
    \bx^{l+1} = \bx^{l} +  f(\bx^{l}; {\nu}_2 \odot \bW),
\end{equation*}
Here, $\nu_2$ can be set to $\frac{1}{L}$ or $\frac{1}{\sqrt{L}}$ in the initialization stage, where $L$ is the number of layers. Initializing it to $\frac{1}{L}$ leads to more stable training compared to $\frac{1}{\sqrt{L}}$, but how does it affect the learning representation ability? This topic requires further in-depth study.

\textbf{The value and necessity of warmup should be investigated.} The theoretical and practical necessity of warmup in Transformers is still not fully understood, despite some existing studies. We observe that even Transformers with 12 layers have a large Lipschitz constant. Training with a large learning rate in the initial stage usually leads to instability. In contrast, ResNet50 can often be trained successfully without warmup. Exploring the inner workings and effects of warmup would be interesting.

\textbf{More attention to the backward process.} Back-propagation, which calculates chain gradients, is crucial in deep learning. In widely used deep learning tools, gradients are returned by auto-differentiation, providing average gradients instead of gradients for each input. However, exploring gradient flow or even obtaining gradients for each sample may inspire new optimization methods. Better consideration of the backward process can lead to novel insights.

There are many other open problems that warrant further exploration, such as better second-order optimization techniques, incorporating Lipschitz smoothness (widely considered in classical numerical optimization) into deep learning, and comparing the generalization ability between smooth and non-smooth functions.

\textbf{Second-order optimization is a promising direction}, although the widely used methods such as SGD momentum and Adam are first-order optimization methods. Obtaining the second-order Hessian matrix is not straightforward in PyTorch and TensorFlow due to cumbersome operations. As a result, many second-order methods like Adahessian rely on approximation methods to estimate the Hessian matrix, which may not be well bounded. We need better tools to calculate higher-order information. We have observed that the JAX toolbox allows for easier acquisition of higher-order information. Future methods may consider utilizing it.

\textbf{Optimization under constrained conditions}, such as on Riemannian manifolds, is an interesting topic. Most deep learning approaches in computer vision and natural language processing are formulated as unconstrained optimization problems. Many constrained optimization problems can be transformed into unconstrained problems by adding regularization terms. However, it would be interesting to directly study constrained optimization and observe its performance.

\section{Conclusion}
\label{sec:Conclusion}
This article has provided a comprehensive analysis of optimization methods in deep learning, with a particular focus on addressing the challenges of gradient vanishing and gradient exploding. We delved into various strategies to tackle these issues, including improving gradient flow and constraining the Lipschitz constant of a network. Our analysis covers both explicit optimization methods, which directly act on optimizer parameters, and implicit optimization methods, which aim to improve the landscape of a network by enhancing its modules. Throughout the article, we have provided an in-depth analysis of these optimization classes and examined the gradients or Jacobians of widely used deep learning modules. We identified existing issues and discussed current and potential enhancements. Moreover, we empirically verified our theoretical analysis. Our goal is to provide readers with a deeper understanding of deep learning optimization and thus inspire the development of more robust, efficient, and high-performing models. The field of deep learning optimization is continuously evolving, presenting both challenges and opportunities. We anticipate exciting developments on the horizon.

\bibliography{iclr2021_conference}
\bibliographystyle{iclr2021_conference}

\appendix
\section{Appendix}
\label{sec:appendix}
\subsection{List of Notations}
\label{sec:appendix_notation}
We use the same notation system as the famous deep learning book~\cite{book_goodfellow2016deep}.

\begin{minipage}{\textwidth}
\centerline{\bf Numbers and Arrays}
\bgroup
\def\arraystretch{1.5}
\begin{tabular}{cp{3.25in}}
$\displaystyle a$ & A scalar (integer or real)\\
$\displaystyle \va$ & A vector\\
$\displaystyle \mA$ & A matrix\\
$\displaystyle \mI_n$ & Identity matrix with $n$ rows and $n$ columns\\
$\displaystyle \mI$ & Identity matrix with dimensionality implied by context\\
$\displaystyle \text{diag}(\va)$ & A square, diagonal matrix with diagonal entries given by $\va$\\
\end{tabular}
\egroup
\index{Scalar}
\index{Vector}
\index{Matrix}
\index{Tensor}
\end{minipage}

\begin{minipage}{\textwidth}
\centerline{\bf Sets and Graphs}
\bgroup
\def\arraystretch{1.5}
\begin{tabular}{cp{3.25in}}
$\displaystyle \sA$ & A set\\
$\displaystyle \R$ & The set of real numbers \\
$\displaystyle \{0, 1\}$ & The set containing 0 and 1 \\
$\displaystyle \{0, 1, \dots, n \}$ & The set of all integers between $0$ and $n$\\
\end{tabular}
\egroup
\index{Scalar}
\index{Vector}
\index{Matrix}
\index{Tensor}
\index{Graph}
\index{Set}
\end{minipage}

\begin{minipage}{\textwidth}
\centerline{\bf Indexing}
\bgroup
\def\arraystretch{1.5}
\begin{tabular}{cp{3.25in}}
$\displaystyle \eva_i$ & Element $i$ of vector $\va$, with indexing starting at 1 \\
$\displaystyle \emA_{i,j}$ & Element $i, j$ of matrix $\mA$ \\
$\displaystyle \mA_{i, :}$ & Row $i$ of matrix $\mA$ \\
$\displaystyle \mA_{:, i}$ & Column $i$ of matrix $\mA$ \\
\end{tabular}
\egroup
\end{minipage}

\begin{minipage}{\textwidth}
\centerline{\bf Linear Algebra Operations}
\bgroup
\def\arraystretch{1.5}
\begin{tabular}{cp{3.25in}}
$\displaystyle \mA^\top$ & Transpose of matrix $\mA$ \\
$\displaystyle \mA \odot \mB $ & Element-wise (Hadamard) product of $\mA$ and $\mB$ \\
$\displaystyle \mathrm{det}(\mA)$ & Determinant of $\mA$ \\
\end{tabular}
\egroup
\index{Transpose}
\index{Element-wise product|see {Hadamard product}}
\index{Hadamard product}
\index{Determinant}
\end{minipage}

\begin{minipage}{\textwidth}
\centerline{\bf Calculus}
\bgroup
\def\arraystretch{1.5}
\begin{tabular}{cp{3.25in}}
$\displaystyle\frac{d y} {d x}$ & Derivative of $y$ with respect to $x$\\ [2ex]
$\displaystyle \frac{\partial y} {\partial x} $ & Partial derivative of $y$ with respect to $x$ \\
$\displaystyle \nabla_\vx y $ & Gradient of $y$ with respect to $\vx$ \\
$\displaystyle \nabla_\mX y $ & Matrix derivatives of $y$ with respect to $\mX$ \\
$\displaystyle \nabla_\tX y $ & Tensor containing derivatives of $y$ with respect to $\tX$ \\
$\displaystyle \frac{\partial f}{\partial \vx} $ & Jacobian matrix $\mJ \in \R^{m\times n}$ of $f: \R^n \rightarrow \R^m$\\
$\displaystyle \nabla_\vx^2 f(\vx)$ & The Hessian matrix of $f$ at input point $\vx$\\
\end{tabular}
\egroup
\index{Derivative}
\index{Integral}
\index{Jacobian matrix}
\index{Hessian matrix}
\end{minipage}


\begin{minipage}{\textwidth}
\centerline{\bf Functions}
\bgroup
\def\arraystretch{1.5}
\begin{tabular}{cp{3.25in}}
$\displaystyle f: \sA \rightarrow \sB$ & The function $f$ with domain $\sA$ and range $\sB$\\
$\displaystyle f \circ g $ & Composition of the functions $f$ and $g$ \\
  $\displaystyle f(\vx ; \vtheta) $ & A function of $\vx$ parametrized by $\vtheta$.
  (Sometimes we write $f(\vx)$ and omit the argument $\vtheta$ to lighten notation) \\
$\displaystyle \log x$ & Natural logarithm of $x$ \\
$\displaystyle \sigma(x)$ & Logistic sigmoid, $\displaystyle \frac{1} {1 + \exp(-x)}$ \\
$\displaystyle || \vx ||_p $ & $\normlp$ norm of $\vx$ \\
$\displaystyle || \vx || $ & $\normltwo$ norm of $\vx$ \\
$\displaystyle x^+$ & Positive part of $x$, i.e., $\max(0,x)$\\
\end{tabular}
\egroup
\index{Sigmoid}
\index{Softplus}
\index{Norm}
\end{minipage}

Sometimes we use a function $f$ whose argument is a scalar but apply
it to a vector, matrix, or tensor: $f(\vx)$, $f(\mX)$, or $f(\tX)$.
This denotes the application of $f$ to the
array element-wise. For example, if $\tC = \sigma(\tX)$, then $\etC_{i,j,k} = \sigma(\etX_{i,j,k})$
for all valid values of $i$, $j$ and $k$.




\typeout{END_CHAPTER "notation" \theabspage}

\subsection{Lipschitz Constants of Some Modules}
\label{sec:appendix_derivations}
\textbf{LayerNorm}.
The Jacobian matrix of LayerNorm is,
\begin{equation*}
\centering
 \frac{\partial \operatorname{LN} (\bx)}{\partial \bx} =  \frac{\sqrt{D}}{\sqrt{\| \by \|_2^2 +\epsilon }}   \left(\boldsymbol{I}-\frac{1}{D} \boldsymbol{1} \boldsymbol{1}^{\top}\right)   \left( \bI - \frac{\by \by^{\top}}{\| \by \|_2^2 +\epsilon}\right)   \operatorname{diag}\left(\bm{\gamma}\right)  
\end{equation*}

Let us look at these terms under $L_2$ norm. 
We have the following equations:
\begin{equation*}
\begin{aligned}
    \frac{\sqrt{D}}{\sqrt{\| \by \|_2^2 +\epsilon }} & \leq \frac{\sqrt{D}}{\sqrt{\epsilon}}, \\
    \sigma_{max}\left(\operatorname{diag}\left(\bm{\gamma}\right)\right)  & = \max _{D}\left|\gamma_{D}\right| , \\
    \sigma_{max} \left(\boldsymbol{I}-\frac{1}{d} \boldsymbol{1} \boldsymbol{1}^{\top}\right) & \leq 1.0, \\
    \sigma_{max}\left(\left( \bI - \frac{\by \by^{\top}}{\| \by \|_2^2 +\epsilon}\right) \right)& \leq 1.0.
\end{aligned}
\end{equation*}
Thus, the final Lipschitz constant for LayerNorm under $L_2$ norm is $ \frac{\sqrt{D}}{\sqrt{\epsilon}} \max _{D}\left|\gamma_{D}\right| $. We would highlight the difference between our derivation and \citeauthor{l2distance_attention_kim2021lipschitz} when deriving the Lipschitz constant of smoothed LayerNorm, we use $L_2$ norm but they use $L_\infty$. In this article, all of our Lipschitz constants are derived based on $L_2$ norm.

\

\textbf{RMSNorm}. The Jacobian matrix of RMSNorm is, 
\begin{equation*}
   \frac{\partial \operatorname{RMSN}(\bx)}{\partial \bx} = \frac{\sqrt{D}}{\sqrt{\| \bx \|_2^2 +\epsilon }}  \left( \bI - \frac{\bx \bx^{\top}}{\| \bx \|_2^2 +\epsilon}\right) \operatorname{diag}\left(\bm{\gamma}\right)
\end{equation*}
Similar to the LayerNorm, we can compute its Lipschitz constant as $ \frac{\sqrt{D}}{\sqrt{\epsilon}} \max _{D}\left|\gamma_{D}\right| $.

A key difference between LayerNorm and RMSNorm  is that the  term $\frac{\sqrt{D}}{\sqrt{\| \by \|_2^2 +\epsilon }}$ in LayerNorm comes from $\by =  \left(\boldsymbol{I}-\frac{1}{D} \boldsymbol{1} \boldsymbol{1}^{\top}\right) \bx$, thus $\| \by \|_2^2 \leq \| \bx \|_2^2$.

\ 

\textbf{WeightNorm}.
The Jacobian matrix of WeightNorm is,
\begin{equation*}
    \frac{\partial \operatorname{WN}(\bx)}{\partial \bx} = \bW,
        \text{where}, \bW(i, :)  = \gamma_i \frac{ \bv_i}{\sqrt{{\|\bv_i\|}_2^2 + \epsilon}}.
\end{equation*}

Under $L_2$ norm, the Lipschitz constant of WeightNorm is $\sigma_{max} (\bW)$.  $\bW$ have dimensions $\mathbb{R}^{O \times D}$. We have the following inequality, 
\begin{equation*}
    \sigma_{\max }(\bW) \leq\|\bW\|_{\mathrm{F}}=\left(\sum_{i=1}^{O} \sum_{j=1}^{D}\left|W_{i j}\right|^{2}\right)^{\frac{1}{2}}= \sqrt{\sum_i^{O} \gamma_i^2}
\end{equation*}

where 
$\|\bW\|_{\mathrm{F}}$ is the Frobenius norm. Equality holds if and only if the matrix 
$\bW$ is a rank-one matrix or a zero matrix. It means all $\bv_i = \bv_j$ for each $i$ and $j$. 

\ 

\textbf{BatchNorm}. The definition of BatchNorm is,
\begin{equation*}
\centering
\begin{aligned}
\bm{\mu}  & = \frac{1}{N} \sum_{i=1}^{N} \bX_{:,i} \\
          \bm{\sigma}^{2} & =  \frac{1}{N} \sum_{i=1}^{N}\left(\bX_{:,i}-\bm{\mu}\right) \odot \left(\bX_{:,i}-\bm{\mu}\right) \\
          \widehat{\bX_{:,i}} &= \left(\bX_{:,i}-\bm{\mu}\right) \oslash {\sqrt{\bm{\sigma}^{2}+\epsilon}} \\
          \mathrm{BN}\left(\bX_{:,i} \right)  &= \bm{\gamma} \odot \widehat{\bX_{:,i}} + \bm{\beta}
\end{aligned}
\end{equation*}

Let us consider the batch size $N$ is very large. Since $\mu$ and $\sigma$ is updated using a moving average, and $N$ is the large. We can simply approximate the Jacobian of $\frac{\partial  \widehat{\bX_{:,i}}}{\partial \bX_{:,i}} \approx \operatorname{diag}(\bm{\gamma})  \operatorname{diag}(\frac{1}{{\sqrt{\bm{\sigma}^{2}+\epsilon}}}) $. Since these two terms are diagonal matrix, it is easy to obtain the approximate Lipschitz constant is $\max_{D} \frac{ \left|\gamma_{D}\right|}{\sqrt{\sigma_{D}^2 + \epsilon}}$.

\end{document}